\documentclass[journal, letterpaper]{IEEEtran}

\usepackage{graphicx}
\usepackage{url}        
\usepackage{amsmath}    
\usepackage{textgreek}
\usepackage{listings}
\usepackage{csvsimple}
\usepackage{longtable}

\usepackage[table]{xcolor}
\usepackage[table,dvipsnames]{xcolor} 
\usepackage[most]{tcolorbox}
\usepackage{tabularx} 
\usepackage{colortbl}     
\definecolor{lightgray}{gray}{0.96} 
\usepackage[normalem]{ulem} 
\usepackage{booktabs}
\usepackage[ruled,vlined,linesnumbered]{algorithm2e}
\usepackage{adjustbox}
\usepackage[table]{xcolor}
\usepackage{multirow}
\usepackage{pifont}
\definecolor{darkred}{RGB}{220,20,60}
\usepackage{amssymb}

\usepackage[title]{appendix}
\usepackage{hyperref}

\usepackage[utf8]{inputenc}
\usepackage{authblk}
\usepackage{bbding}
\newcommand{\equalcontrib}{\textsuperscript{\dag}}

\begin{document}

\title{Beyond Surface Artifacts: Capturing Shared Latent Forgery Knowledge Across Modalities}

\author[1]{Jingtong Dou\equalcontrib}
\author[1]{Chuancheng Shi\equalcontrib}
\author[2]{Jian Wang}
\author[3]{Fei Shen\textsuperscript{\Envelope}}
\author[1]{Zhiyong Wang}
\author[3]{Tat-Seng Chua}

\affil[1]{The University of Sydney}
\affil[2]{Nanjing University of Posts and Telecommunications} 
\affil[3]{National University of Singapore}

\affil[ ]{\dag ~ Equal Contribution}
\affil[ ]{\Envelope ~ Corresponding Author}

\maketitle

\begin{abstract}
	As generative artificial intelligence evolves, deepfake attacks have escalated from single-modality manipulations to complex, multimodal threats. Existing forensic techniques face a severe generalization bottleneck: by relying excessively on superficial, modality-specific artifacts, they neglect the shared latent forgery knowledge hidden beneath variable physical appearances. Consequently, these models suffer catastrophic performance degradation when confronted with unseen "dark modalities." To break this limitation, this paper introduces a paradigm shift that redefines multimodal forensics from conventional "feature fusion" to "modality generalization." We propose the first modality-agnostic forgery (MAF) detection framework. By explicitly decoupling modality-specific styles, MAF precisely extracts the essential, cross-modal latent forgery knowledge. Furthermore, we define two progressive dimensions to quantify model generalization: transferability toward semantically correlated modalities (Weak MAF), and robustness against completely isolated signals of "dark modality" (Strong MAF). 
    To rigorously assess these generalization limits, we introduce the DeepModal-Bench benchmark, which integrates diverse multimodal forgery detection algorithms and adapts state-of-the-art generalized learning methods. This study not only empirically proves the existence of universal forgery traces but also achieves significant performance breakthroughs on unknown modalities via the MAF framework, offering a pioneering technical pathway for universal multimodal defense.
\end{abstract}

\section{Introduction}
	
\begin{figure}[t]
    \centering
    \includegraphics[width=1\linewidth]{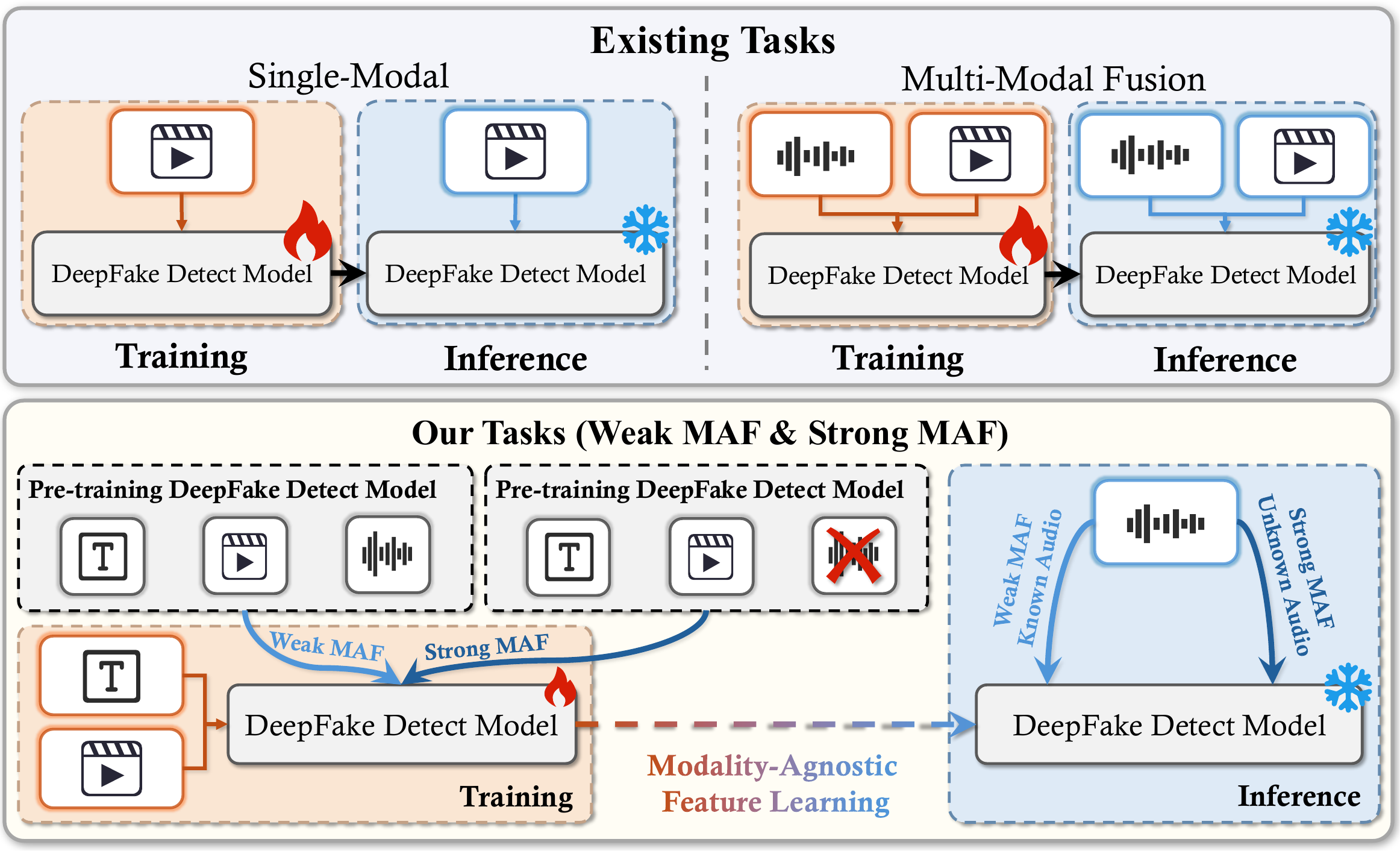} 
\caption{\textbf{From modality-binding to modality-agnostic forgery (MAF).} Unlike existing tasks that overfit to specific physical representations seen during training, our proposed MAF task focuses on extracting shared latent forgery knowledge. This modality-agnostic feature learning enables robust source-free domain generalization against unknown modalities (e.g., unseen audio), systematically tackling both Weak and Strong MAF challenges.}
    \label{fig:intro}
\end{figure}

\PARstart{W}{ith} the explosive evolution of generative AI, deepfake attacks have transitioned from single-modality~\cite{dou2026dnauncoveringuniversallatent,qin2025scalingaigeneratedimagedetection, wang-etal-2018-glue, martinek-bartuzi-trokielewicz-2025-detecting} visual manipulation to highly coordinated, multimodal deceptions~\cite{xu2026maremultimodalalignmentreinforcement, klein2025pindropitaudiovisual, kukanov2025klassifyverifyaudiovisualdeepfake, guo2025rethinkingvisionlanguagemodelface}. Modern forgery techniques can now synthesize highly realistic video, audio, and associated text simultaneously. This cross-modal consistency not only enhances deceptiveness but also poses a severe challenge to traditional forensic technologies. To counter this, mainstream multimodal defenses have relied heavily on feature fusion strategies. However, as illustrated in Fig.~\ref{fig:intro}, these traditional paradigms operate under a strict "closed-modality" mindset, assuming test modalities must be seen during the training phase. Consequently, they suffer from a severe "modality-binding" bottleneck. When confronted with the continuous emergence of novel media formats in the real world, such as infrared signals, depth streams, or heterogeneous unstructured data, these models exhibit catastrophic performance degradation because they cannot detach from the specific physical representations they have memorized.

This vulnerability exposes a fundamental limitation: existing methods~\cite{kundu2025universalsyntheticvideodetector,Zhou_2021_ICCV,yu2025unlockingcapabilitieslargevisionlanguage} merely overfit to the surface-level artifacts of specific known modalities, failing entirely to capture the underlying essence of generative algorithms. To break this impasse, we argue that the strategic focus of multimodal forgery detection must undergo a paradigm shift, from modality-binding feature fusion to modality generalization. We propose a bold hypothesis: although the physical manifestations of forgery vary drastically across different media, there exists a shared latent forgery knowledge region that transcends these modal boundaries. Capturing this inherent statistical bias, which is left behind by generative models regardless of the output medium, is the key to achieving a universal, "train once, generalize across all modalities" defense mechanism.

Based on this hypothesis, we formally introduce the modality-agnostic forgery (MAF) task and detection framework. As depicted in Fig.~\ref{fig:intro}, MAF is designed to decouple modality-specific style noise, precisely extracting the essential, cross-modal latent forgery features required to maintain robust discriminative power on entirely unseen test modalities. To comprehensively assess this capability under varying real-world perceptual constraints, we establish two progressive evaluation paradigms. Weak MAF evaluates generalization toward unseen modalities that can still be mapped into a unified semantic space via existing pre-trained perceptors (e.g., ImageBind~\cite{girdhar2023imagebind} or LanguageBind~\cite{zhu2023languagebind}), verifying whether a model can achieve "forensic fingerprint" alignment beneath macro-semantic alignment. In contrast, Strong MAF tackles the rigorous "perception island" challenge: defending against completely isolated "dark modalities" that are neither seen during training nor semantically alignable by existing models. This strict setting forces the framework to capture the inherent statistical flaws of generative AI purely from the underlying logic level, achieving higher-dimensional robustness without the prerequisite of shared representations.

To systematically evaluate and drive forward this modality-agnostic research trajectory, we construct DeepModal-Bench, the first comprehensive benchmark specifically tailored for modality generalization in  forgery detection. Integrating diverse state-of-the-art multimodal algorithms and adapting advanced domain generalization (DG) strategies, our extensive experiments across multiple large-scale public datasets provide compelling empirical evidence of universal forgery traces that transcend physical representations. 

In summary, the main contributions of this work are threefold:
\begin{itemize}
    \item We redefine the objective of multimodal forensics, shifting the paradigm from modality-binding feature fusion to modality generalization by proposing the concept of shared latent forgery knowledge.
    \item We introduce the MAF framework and formulate two rigorous evaluation settings (Weak MAF and Strong MAF) to systematically tackle both semantically correlated unseen modalities and physically isolated "dark modalities."
    \item We release DeepModal-Bench, the first comprehensive benchmark for this task. Our extensive evaluations not only validate the objective existence of universal forgery traces but also illuminate the critical pathway from semantic alignment to true forensic alignment, offering a pioneering technical route for universal multimodal defense.
\end{itemize}

\section{Related Work}

\noindent\textbf{Modality Binding.}
Multimodal learning has rapidly evolved from simple pairwise alignment toward unified modality binding. While CLIP~\cite{radford2021learning} laid the foundational paradigm, advanced models such as ImageBind~\cite{girdhar2023imagebind} and its successors~\cite{zhu2023languagebind, srivastava2024omnivec2, wang2023one, guo2023point, lyu2024unibind} have successfully anchored diverse modalities within a shared semantic embedding space. To further enhance binding quality and system robustness, researchers have actively explored optimization pathways ranging from feature decoupling and geometric constraints~\cite{song2025bridge, yinmultimodal, li2023blip, qian2025decalign, huang2025revisiting, du2026inference, liu2025continual, wang2024open} to dynamic modality selection. However, despite these architectural leaps, existing binding mechanisms remain fundamentally tethered to the physical representations seen during training. When encountering entirely unseen or isolated "dark modalities," their performance degrades catastrophically. This persistent "modality-binding" bottleneck serves as the theoretical motivation for our modality-agnostic forgery (MAF) framework and the DeepModal-Bench benchmark, driving the shift toward truly modality-agnostic detection.

\noindent\textbf{Forgery Detection}
Driven by the surge of multimodal generative technologies, forgery detection is undergoing a crucial paradigm shift from shallow "perceptual classification" to "deep semantic understanding." This evolution is heavily supported by diverse large-scale datasets~\cite{zhao2025deepfakebenchmmcomprehensivebenchmarkmultimodal, shao2023detectinggroundingmultimodalmedia, chen2024demambaaigeneratedvideodetection, jung2026a, cai2025avdeepfake1mlargescaleaudiovisualdeepfake, kuckreja2025tellhabibirealfake} and advanced frameworks like LAV-DF~\cite{cai2023reallymeanthatcontent} and HAMMER~\cite{shao2023detectinggroundingmultimodalmedia}, which utilize multi-task prompt learning for precise manipulation localization. Furthermore, leveraging large multimodal models (LMMs), next-generation architectures construct deep forgery knowledge spaces to defend against unknown diffusion models~\cite{song2025learningmultimodalforgeryrepresentation, jung2026a}, while others integrate LLMs~\cite{xu2026maremultimodalalignmentreinforcement, wen2025spotfakelargemultimodal, guo2025rethinkingvisionlanguagemodelface} for interpretable reasoning. However, these advanced methodologies rely excessively on explicit "modality alignment", inadvertently creating a severe modality-binding problem. Confronted with unseen "dark modalities" (e.g., infrared or depth streams), these models fail to decouple the essence of forgery from physical representations. To break this impasse, we introduce the MAF framework, shifting the forensic focus to modality generalization for unseen test modalities.

\noindent\textbf{Domain Generalization.}
Domain generalization (DG) aims to learn models from multiple source domains that can seamlessly adapt to unknown target distributions without requiring target-specific retraining~\cite{wang2022generalizing}. Early DG research~\cite{ganin2016domainadversarialtrainingneuralnetworks, wei2025indirect, pan2010domain, li2018deep} primarily focused on eliminating inter-domain discrepancies through spatial mapping and adversarial learning, while subsequent data-centric strategies like Mixup~\cite{zhang2017mixup} and DDAIG~\cite{zhou2020deep} significantly enhanced out-of-distribution robustness. More recently, the focus has shifted toward multimodal contexts, where methods like SimMMDG~\cite{dong2023simmmdg} and MBCD~\cite{wang2026modality} emphasize multimodal feature decoupling, and CLIP-powered technologies~\cite{li2023blip, li2026clip} leverage massive pre-training priors for zero-shot generalization. While DG has matured in standard vision and language tasks, its potential in cross-modal forgery detection remains largely underexplored. To bridge this gap, our MAF framework adapts DG strategies, treating physical modalities as distinct domains, to extract invariant latent forgery knowledge.

\begin{figure*}[t]
    \centering
    \includegraphics[width=1\linewidth]{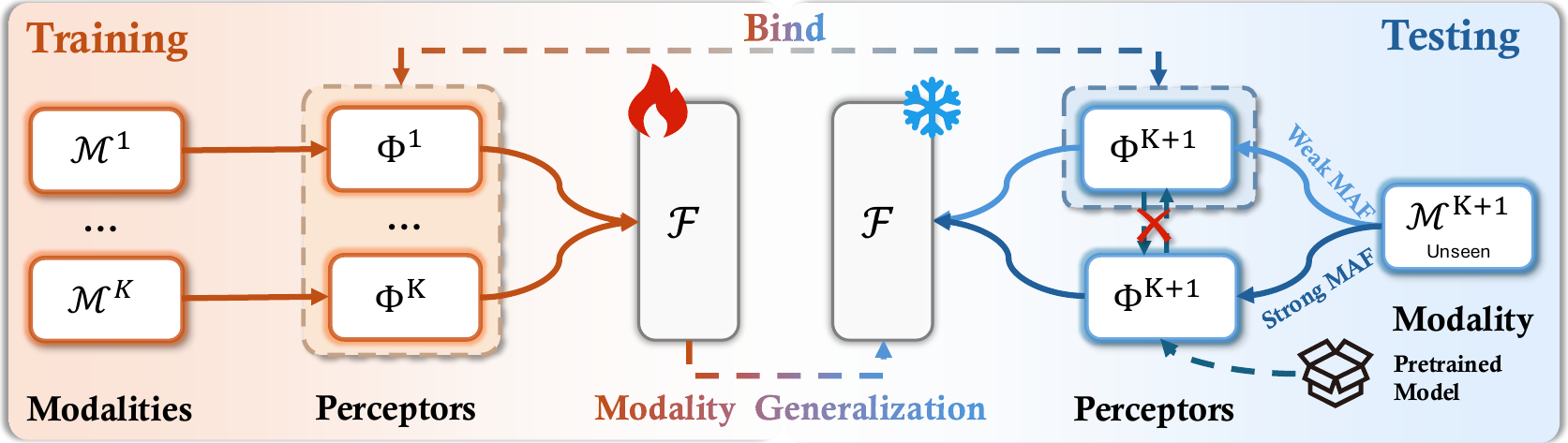}
\caption{Architecture of the MAF framework. Following training on known modalities, the universal detector $\mathcal{F}$ evaluates an unseen modality $\mathcal{M}^{K+1}$ under two settings: Weak MAF utilizes a semantically aligned perceptor, while Strong MAF employs an independent perceptor to assess forensic generalization under strict physical and semantic isolation.}
    \label{fig:pipeline}
\end{figure*}

\section{Formulating Modality-Agnostic Forgery}
\subsection{The "Modality-Binding" Bottleneck}
Most existing multimodal forgery detection frameworks operate~\cite{xu2026maremultimodalalignmentreinforcement, klein2025pindropitaudiovisual} under a strict closed-modality assumption. They rely heavily on extracting modality-specific physical fingerprints (e.g., spatial blending artifacts in images or frequency anomalies in audio). Consequently, these models tend to overfit to the surface-level features of seen modalities, leading to severe performance degradation when encountering novel media formats absent from the training phase. We define this limitation as the "modality-binding" bottleneck. 
To overcome this bottleneck, we redefine the task from modality-specific detection to modality-agnostic generalization. Formally, let $\mathcal{D}_{seen} = \{ \mathcal{M}^1, \mathcal{M}^2, \dots, \mathcal{M}^K \}$ denote a source training dataset consisting of $K$ known modalities, as illustrated in Fig.~\ref{fig:pipeline}. Each modality $\mathcal{M}^k = \{(x_i^k, y_i^k)\}_{i=1}^{N_k}$ contains $N_k$ instances, where $x_i^k \in \mathcal{X}$ represents the raw input signal and $y_i^k \in \mathcal{Y}$ represents the corresponding binary forgery label.

We hypothesize that any forged signal $x$ can be conceptually decomposed into two distinct components: a modality-specific style component ($\mathcal{S}$) representing the physical medium, and a modality-invariant essence component ($\mathcal{E}$) capturing the underlying statistical anomalies left by the generative algorithm. Unlike traditional paradigms that memorize physical representations, our objective is to learn a universal forgery detector $\mathcal{F}$ that explicitly strips away the superficial style $\mathcal{S}$ and exclusively utilizes the forgery essence $\mathcal{E}$ for decision-making. 
To systematically operationalize this disentanglement, we introduce a set of perceptors $\Phi$ to map raw inputs into feature spaces, and evaluate the detector $\mathcal{F}$ under two progressive scenarios: Weak MAF and Strong MAF. The ultimate challenge in both settings is to extrapolate the "local" forgery rules learned from $\mathcal{D}_{seen}$ to capture the "global" forgery essence within a completely unseen test modality $\mathcal{M}^{K+1}$.

\begin{figure*}[t]
    \centering
    \includegraphics[width=0.95\linewidth]{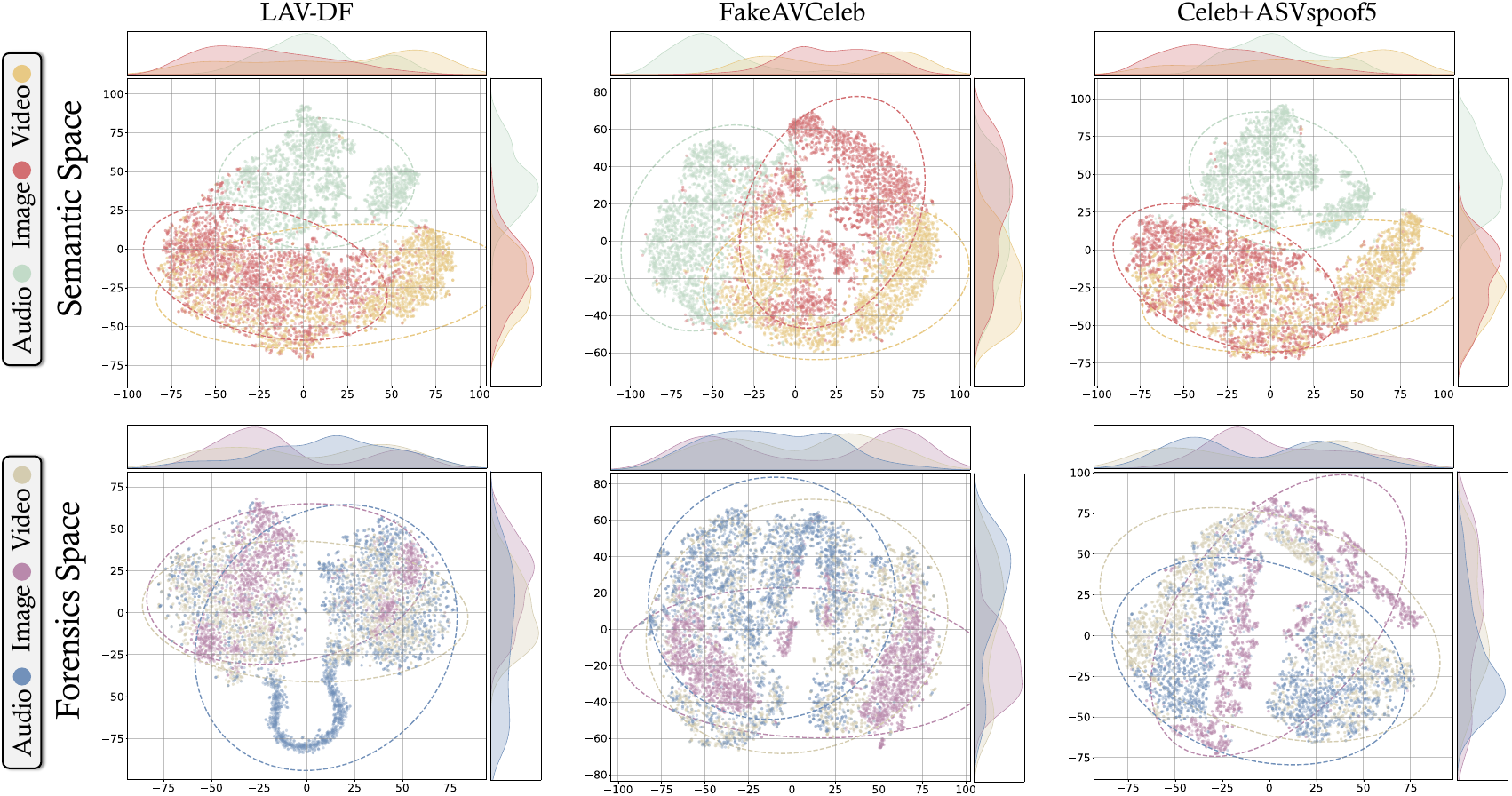}
\caption{Comparison of feature distributions between semantic and forensics spaces. \textit{Top (Semantic Space):} Forgery features exhibit strong modality dependence and lack shared fingerprints, resulting in isolated distributions across modalities. \textit{Bottom (Forensics Space):} Through feature decoupling, the distributions of both known and completely unseen modalities converge, successfully achieving cross-modal universality of latent forgery knowledge.}
    \label{fig:tsne*}
\end{figure*}

\subsection{Weak MAF: Semantic Generalization}
In the context of weak modality-agnostic forgery (Weak MAF), we assume the existence of a unified semantic alignment space $\mathcal{V}$. This space is governed by a set of pre-trained perceptors based on large-scale multimodal foundational models (e.g., ImageBind~\cite{girdhar2023imagebind} or LanguageBind~\cite{zhu2023languagebind}). 
During the training phase, each perceptor $\Phi^k$ serves as a dedicated feature extractor that maps the input data of known modalities into aligned feature vectors $z_i^k \in \mathcal{V}$:
\begin{equation}
    z_i^k = \Phi^k(x_i^k).
\end{equation}
The universal forgery detector $\mathcal{F}$ is then trained on these mapped features from the $K$ known modalities. During the testing phase, a novel modality $\mathcal{M}^{K+1}$ is introduced. By directly applying the detector $\mathcal{F}$ to the extracted features $z^{K+1} = \Phi^{K+1}(x^{K+1})$, we assess the model's ability to capture cross-modal forgery fingerprints within this shared semantic space.

The core challenge of Weak MAF lies in bridging the "forgery gap." Although the test modality is semantically aligned with the training modalities, the distribution of their forgery features often exhibits significant discrepancies due to varying generative mechanisms. Consequently, the detector $\mathcal{F}$ must act as a feature disentanglement operator. It must transcend macro-semantic alignment to effectively filter out the modality style $\mathcal{S}$ from the shared semantic features $z$, thereby distilling the pure forgery essence $\mathcal{E}$ for accurate classification.

\subsection{Strong MAF: The "Dark Modality" Challenge}
In contrast, strong modality-agnostic forgery (Strong MAF) addresses the formidable challenge of "dual isolation in physics and semantics." In this scenario, the newly emerging test modality $\mathcal{M}^{K+1}$ acts as a "dark modality." Due to the lack of paired training data or prior knowledge, it cannot be mapped into the unified semantic space $\mathcal{V}$ using any existing modality-bound perceptors.
To conduct forgery detection under such extreme constraints, we adopt a decoupled technical approach. An independent modality perceptor $\Phi^{K+1}$ is constructed and trained using only the data from the test modality $\mathcal{M}^{K+1}$ (e.g., via self-supervised learning). This isolated perceptor extracts the feature vectors $z_i^{K+1}$, which are then fed directly into the universal detector $\mathcal{F}$, which was pre-trained exclusively on the known modalities for inference. For the $i$-th test sample, the predicted label is formulated as:
\begin{equation}
    \hat{y}_i^{K+1} = \mathcal{F}(\Phi^{K+1}(x_i^{K+1})).
\end{equation}

The algorithmic significance of Strong MAF lies in validating the existence of universal forgery fingerprints. By forcing the detector to classify signals from a completely decoupled, mutually isolated representation space, we demonstrate that the MAF framework bypasses surface-level variations in embeddings. Instead, it successfully locks onto the underlying statistical biases inherent to the logic of generative AI, achieving higher-dimensional robustness against entirely unknown heterogeneous threats.

\section{Proposed MAF Framework}

\subsection{From Semantic to Forensic Alignment}
Although existing multi-modal alignment models achieve significant semantic consistency through joint embedding spaces, in forensic tasks, mere semantic alignment does not equate to forensic alignment. From Fig.~\ref{fig:tsne*}, when mapping modalities into a shared semantic space, the feature distributions of different modalities remain highly isolated from one another. More critically, the distribution of an unseen test modality shows almost no overlap with the training set.
This phenomenon exposes a fundamental limitation: pre-trained multimodal models prioritize capturing macro-level semantic concepts (e.g., visual objects or spoken content). Consequently, the subtle, low-level statistical biases left by generative algorithms are heavily dominated by these powerful semantic signals. Rather than being discarded, they become deeply entangled and sub-optimally weighted beneath physical representations. Simply stacking classifiers on top of these semantic spaces forces the model to overfit to surface-level artifacts, failing to capture the cross-modal essence of forgery.

\vspace{-0.05cm}
To overcome this, the proposed MAF framework introduces a paradigm shift from semantic alignment to forensic alignment. By enforcing a feature decoupling mechanism, MAF strips away modality-specific styles. As shown in the forensic space of Fig.~\ref{fig:tsne*}, this decoupling causes the previously isolated features to converge significantly, allowing unseen modalities to form a substantial distributional overlap with the training data. This proves that capturing the inherent algorithmic biases across digital signals is the key to robust generalization.

\subsection{Framework Architecture Overview}
As outlined in Algorithm~\ref{alg:maf}, the modality-agnostic forgery (MAF) detection framework consists of a unified training phase and a dual-scenario inference phase. The architecture primarily comprises a set of modality-specific perceptors $\Phi$ and a shared, lightweight universal forgery detector $\mathcal{F}$.
During the training phase, inputs $x_i^k$ from each known modality $\mathcal{M}^k$ are encoded into representation vectors $z_i^k$ by their corresponding perceptor $\Phi^k$. These representations are then fed into the universal detector $\mathcal{F}$. The entire framework is optimized jointly using a standard classification loss $\mathcal{L}_{\text{cls}}$ alongside a cross-modal regularization term ($\mathcal{L}_{\text{DG}}$ or $\mathcal{L}_{\text{MML}}$) designed to enforce forensic alignment.
During the inference phase, the model confronts an unseen test modality $\mathcal{M}^{K+1}$. For the Weak MAF scenario, the framework directly invokes a structurally identical or semantically aligned pre-trained perceptor $\Phi^{K+1}$ to extract features. Conversely, for the Strong MAF scenario where the modality is completely isolated, the framework first initializes an independent perceptor $\Phi^{K+1}$ via self-supervised training. In both settings, the extracted latent features $z_i^{K+1}$ are fed directly into the frozen detector $\mathcal{F}$ to produce the final binary forgery prediction $\hat{y}_i^{K+1}$.

\begin{algorithm}[t]
\caption{Modality-Agnostic Forgery (MAF) Detection Framework}
\label{alg:maf}
\LinesNumbered
\KwIn{$\mathcal{D}_{\text{seen}} = \{\mathcal{M}^1, \ldots, \mathcal{M}^K\}$,\quad $\mathcal{D}_{\text{unseen}} = \mathcal{M}^{K+1}$}
\KwOut{$\hat{y}^{K+1}$}
\BlankLine
Initialize detector $\mathcal{F}$ as a lightweight multi-layer perceptron\;
\BlankLine
\For{each training step}{
    \ForEach{$k \in \{1, \ldots, K\}$}{
        $z_i^k = \Phi^k(x_i^k)$;\quad $\mathcal{L}_{\text{cls}} \mathrel{+}= \text{CrossEntropy}(\mathcal{F}(z_i^k),\ y_i^k)$\;
    }
    \eIf{Algorithm $==$ \texttt{MML}}{
        $\mathcal{L}_{\text{total}} = \mathcal{L}_{\text{cls}} + \lambda \cdot \mathcal{L}_{\text{MML}}(\{z_i^1, \ldots, z_i^K\})$\;
    }{
        $\mathcal{L}_{\text{total}} = \mathcal{L}_{\text{cls}} + \lambda \cdot \mathcal{L}_{\text{DG}}(\{z_i^1, \ldots, z_i^K\})$\;
    }
    Update\_Weights$(\mathcal{F},\ \mathcal{L}_{\text{total}})$\;
}
\BlankLine
\eIf{Scenario $==$ \texttt{Weak MAF}}{
    $z_i^{K+1} = \Phi^{K+1}(x_i^{K+1})$\;
}{
    $\Phi^{K+1} \leftarrow \text{SelfSupervisedTrain}(\mathcal{M}^{K+1})$\;
    $z_i^{K+1} = \Phi^{K+1}(x_i^{K+1})$\;
}
\BlankLine
\KwRet{$\hat{y}_i^{K+1} = \mathcal{F}(z_i^{K+1})$}\;
\end{algorithm}

\subsection{Optimizing Forensic Alignment}
The core objective of the MAF framework is to implement the theoretical disentanglement of the modality style component ($\mathcal{S}$) from the shared forgery essence ($\mathcal{E}$) formulated in Section 3. To achieve this, we shift the optimization objective away from merely fitting the training data and instead actively penalize the learning of modality-specific physical priors.
Within the training loop (Algorithm~\ref{alg:maf}), this decoupling operator is mathematically instantiated through the regularization term $\mathcal{L}_{\text{DG}}$. By conceptualizing each different physical modality as a distinct "domain," we adapt advanced domain generalization (DG) strategies (such as invariant risk minimization or information bottleneck) to modulate the gradient updates of the detector $\mathcal{F}$. 
Unlike multi-modal learning ($\mathcal{L}_{\text{MML}}$), which seeks to maximize complementary information across modalities, the $\mathcal{L}_{\text{DG}}$ constraint acts as a severe information filter. It maximizes the similarity of consistent statistical features shared across $\mathcal{D}_{\text{seen}}$ while actively suppressing hardware-related or semantic-related style interference. This dynamic gradient adjustment forces the isolated feature distributions to map into a unified, low-dimensional "decoupled space." Consequently, the framework effectively mitigates the "modality-binding" phenomenon, ensuring that the detector strictly locks onto the fundamental logical biases of AIGC, enabling robust detection on novel heterogeneous attacks via a source-free domain generalization paradigm.

\begin{table}[t]
\centering
\caption{Overview of the experimental datasets in DeepModal-Bench. We use ``$\checkmark$'' and ``\textcolor{gray}{\ding{55}}'' to indicate the presence or absence of observable modalities, respectively. The ``Instances'' reports the combined total of genuine and forged samples used for benchmarking across different semantic categories.}
\label{tab:dataset_stats}
\resizebox{\linewidth}{!}{
\begin{tabular}{l|ccc|c|c}
\toprule
Dataset & Video & Audio  & Image  & Instances & Category \\ \midrule
LAV-DF~\cite{cai2023reallymeanthatcontent} & $\checkmark$ & $\checkmark$ & $\checkmark$  & 101,675 & Aligned \\
FakeAVCeleb~\cite{khalid2022fakeavcelebnovelaudiovideomultimodal} & $\checkmark$ & $\checkmark$ & $\checkmark$  & 20,000 & Aligned \\ \midrule
ASVspoof5 ~\cite{wang2024asvspoof5crowdsourcedspeech} & \textcolor{gray}{\ding{55}} & $\checkmark$ & \textcolor{gray}{\ding{55}}  & 1,002,889 & Unaligned \\
Celeb-DF++~\cite{li2025celeb} & $\checkmark$ & \textcolor{gray}{\ding{55}} & $\checkmark$  & 53,786 & Unaligned \\ 
\bottomrule
\end{tabular}
}
\end{table}

\begin{figure*}[t]
    \centering
    \includegraphics[width=0.95\linewidth]{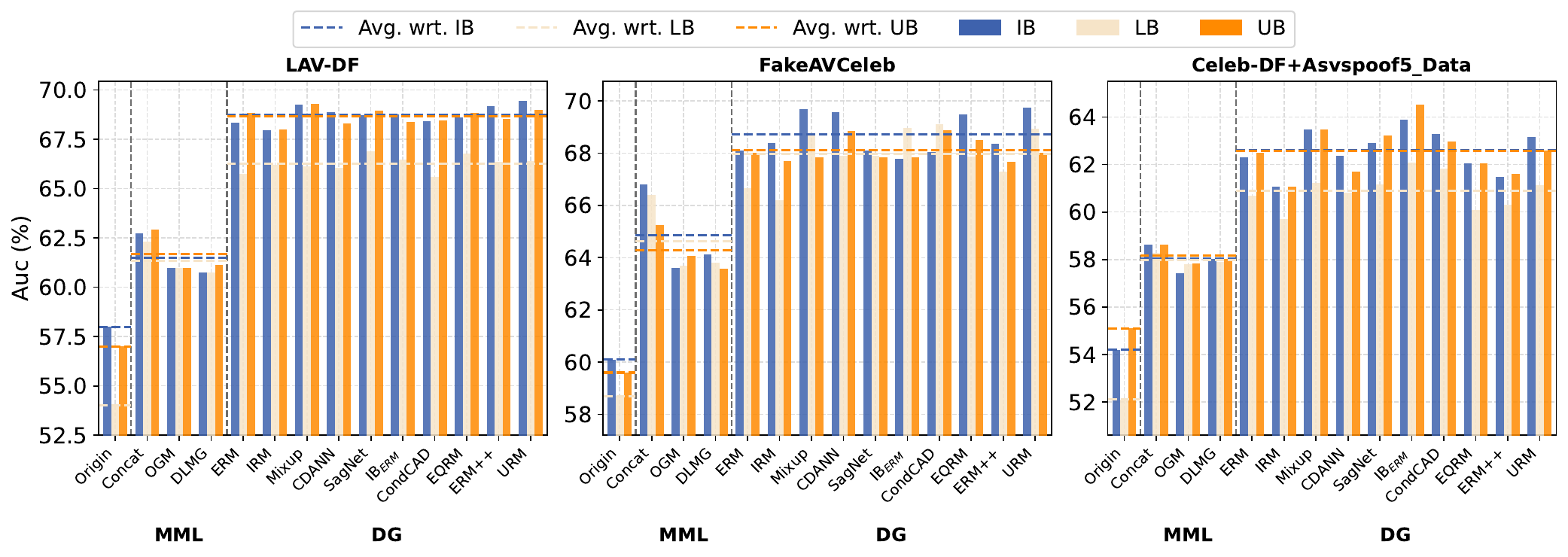}
\caption{Performance evaluation under the Weak MAF setting across three pre-trained perceptors (ImageBind, LanguageBind, and UniBind). Each bar reports the average AUC across the video, audio, and image modalities, utilizing the Oracle validation strategy for model selection. Dashed horizontal lines denote the mean performance of multi-modal learning (MML) and domain generalization (DG) methods for each corresponding perceiver.}
    \label{fig:weakmaf}
\end{figure*}

\begin{figure*}[t]
    \centering
    \vspace{-0.2cm}
    \begin{minipage}[b]{0.34\textwidth}
        \centering
        \includegraphics[width=0.98\linewidth]{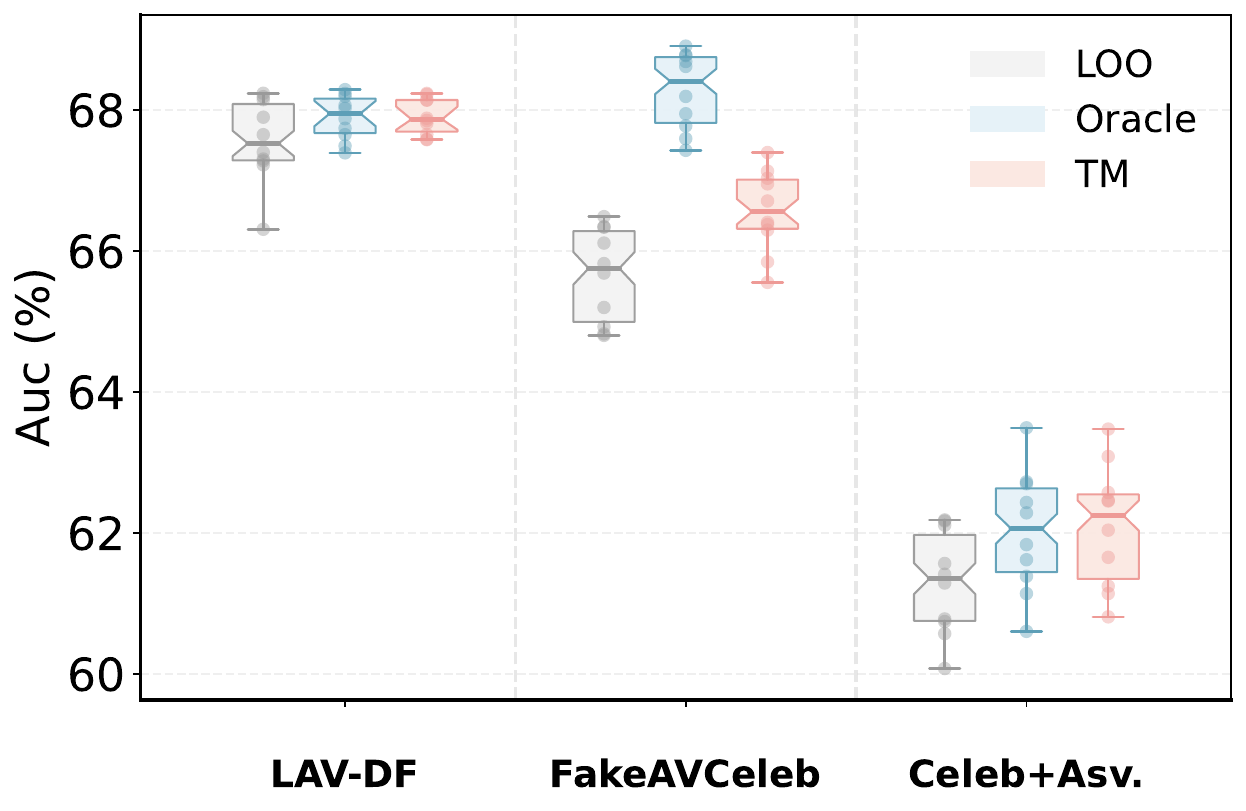} 
    \vspace{-0.2cm}\caption{Averaged performance comparison for different model selection methods across different benchmarks (Weak MAF).}
        \label{fig:weak_model_selection}
    \end{minipage}
    \hfill 
    \begin{minipage}[b]{0.64\textwidth}
        \centering
        \includegraphics[width=0.98\linewidth]{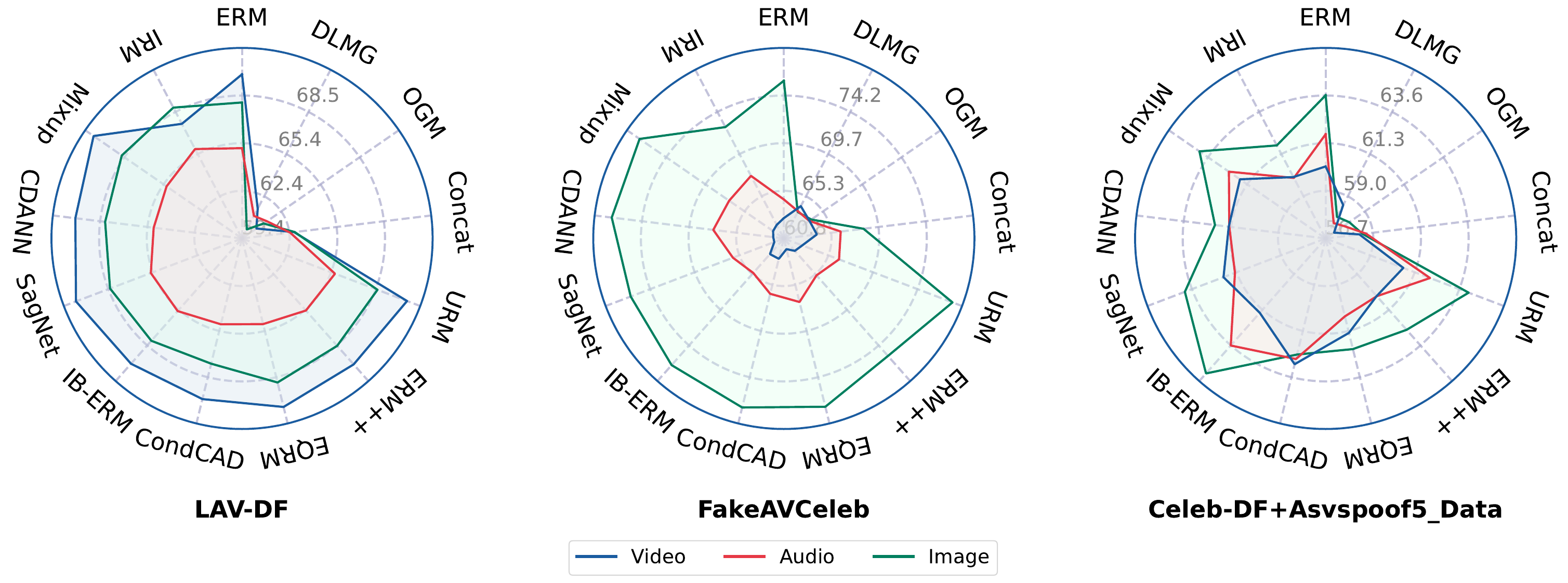} 
        \caption{Weak MAF Performance comparison for specific testing modalities across different algorithms.}
        \label{fig:weak_radar_charts}
    \end{minipage}
        \vspace{-0.3cm}
\end{figure*}

\section{Experimental Setups}

\subsection{DeepModal-Bench Configuration}
We introduce DeepModal-Bench, the first comprehensive benchmark dedicated to modality generalization in multimodal forgery detection. Built upon the ModalBed platform, it integrates diverse large-scale datasets and state-of-the-art baselines to establish a standardized and rigorous environment for evaluating cross-modal forensic capabilities.

\noindent\textbf{Datasets.} The core testing protocol of DeepModal-Bench is driven by a carefully designed dataset taxonomy. To systematically probe model robustness under varying semantic conditions, the benchmark constructs three distinct dataset groups, functioning under two broad categories (Table~\ref{tab:dataset_stats}). The first two groups, LAV-DF~\cite{cai2023reallymeanthatcontent} and FakeAVCeleb~\cite{khalid2022fakeavcelebnovelaudiovideomultimodal}, contain naturally spatiotemporally aligned modalities sharing a common semantic source; these constitute the \textit{modality-aligned} category. The third group, a combination of ASVspoof5~\cite{wang2024asvspoof5crowdsourcedspeech} and Celeb-DF++~\cite{li2025celeb}, is drawn from completely independent sources. Its modalities are fully decoupled physically and semantically, serving as the \textit{modality-unaligned} category. This deliberate contrast within DeepModal-Bench forces the models to prove whether they capture universal forgery traces purely through underlying generative biases, completely independent of semantic correspondence.

\noindent\textbf{Evaluation Protocols.} The benchmark incorporates the progressive Weak MAF and Strong MAF scenarios. For model selection, we adopt three distinct validation strategies to comprehensively assess generalization bounds: (1) \textit{Training-modality (TM) validation} reserves a split from the source modality to assess basic convergence; (2) \textit{Leave-one-out (LOO) validation} designates each known modality in turn as a pseudo-unknown domain, directly encouraging cross-modal invariance learning during training; (3) \textit{Oracle validation} introduces a minimal amount of labeled data from the target test modality for hyperparameter selection, serving as a performance upper-bound reference. Given the inherent class imbalance in cross-modal settings and the varying discrimination difficulty across modalities, we adopt the AUC as the unified primary metric.

\begin{figure*}[t]
    \centering
    \includegraphics[width=0.95\linewidth]{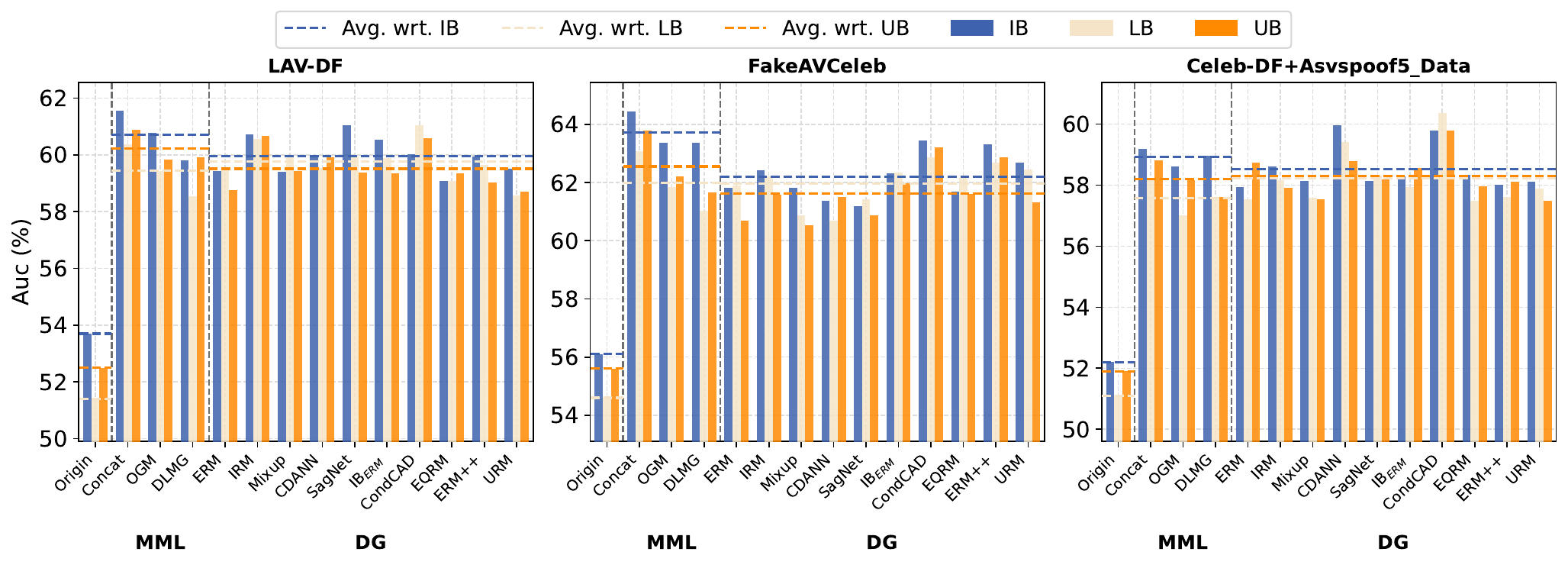}
    \caption{Performance evaluation under Strong MAF across three pre-trained perceptors (ImageBind, LanguageBind, and UniBind). The bar chart illustrates the average accuracy of various detection algorithms across different modalities. The dashed lines represent the overall performance benchmarks for multi-modal learning (MML) and domain generalization (DG) methods, respectively, under each corresponding perceiver. }
\label{fig:strong_mg_loo1}
\end{figure*}

\begin{figure*}[t]
    \centering
    \vspace{-0.2cm}
    \begin{minipage}[b]{0.35\textwidth}
        \centering
        \includegraphics[width=0.98\linewidth]{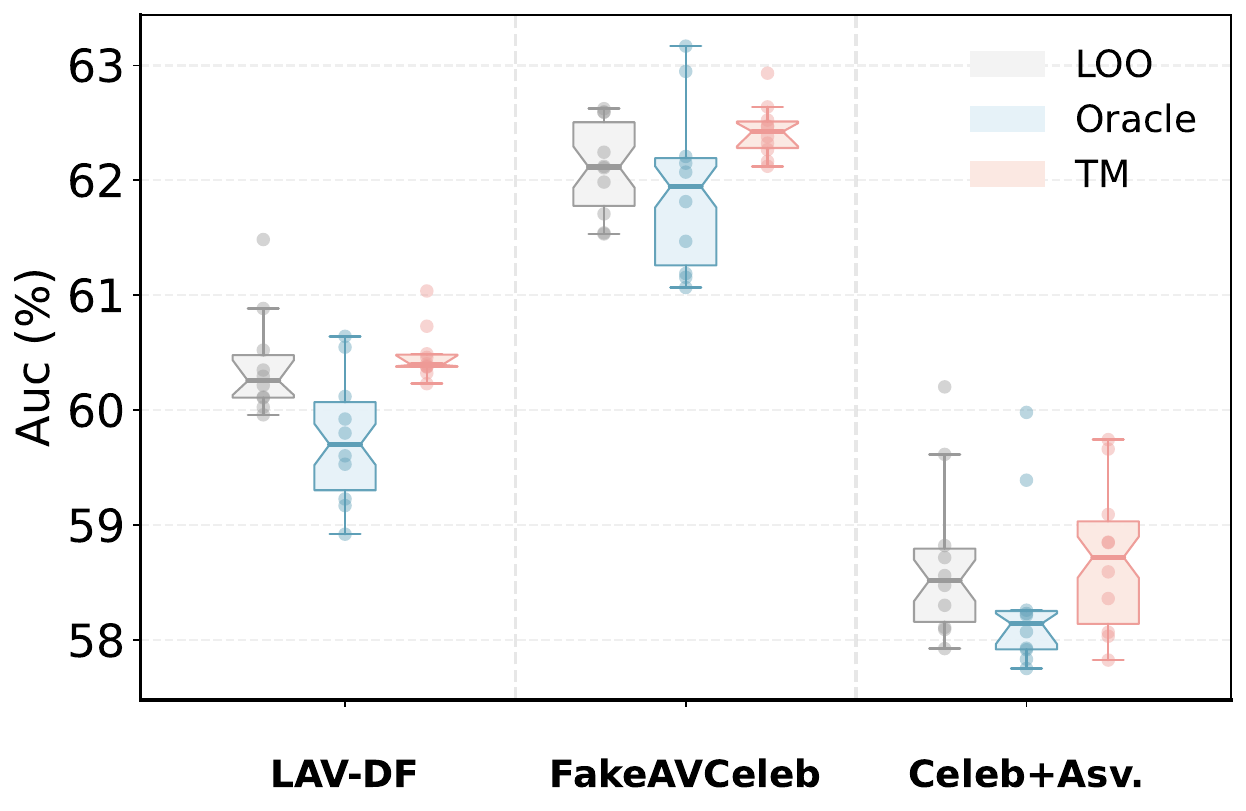} 
        \caption{Averaged performance comparison for different model selection methods across different benchmarks (Strong MAF).}
        \label{fig:strong_model_selection}
    \end{minipage}
    \hfill 
    \begin{minipage}[b]{0.62\textwidth}
        \centering
        \includegraphics[width=0.98\linewidth]{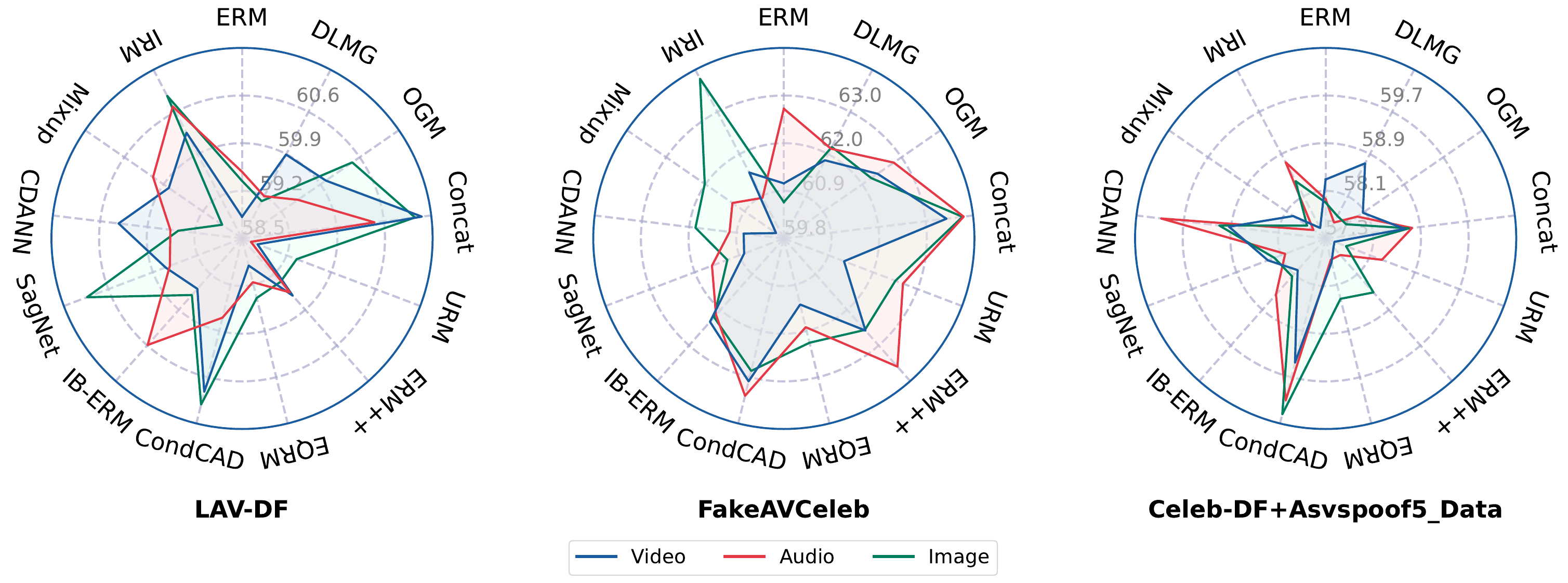} 
        \caption{Strong MAF Performance comparison for specific testing modalities across different algorithms.}
        \label{fig:strong_radar_charts}
    \end{minipage}
        \vspace{-0.3cm}
\end{figure*}

\subsection{Implementation Details}

\noindent\textbf{Feature Extractors (Perceptors).} 
In the Weak MAF setting, we employ three frozen pretrained models, ImageBind~\cite{girdhar2023imagebind}, LanguageBind~\cite{zhu2023languagebind}, and UniBind~\cite{lyu2024unibind}, to map heterogeneous inputs into a unified 1024-dimensional space. By leveraging vision-anchored, language-anchored, and modality-agnostic backends, we mitigate bias from any single alignment paradigm and objectively evaluate universal forgery artifacts relying solely on their semantic associations.
In the Strong MAF setting, we discard pretrained semantic priors entirely. Instead, we construct independent ViT~\cite{dosovitskiy2021imageworth16x16words} encoders for visual and spectral signals, and a T5~\cite{raffel2023exploringlimitstransferlearning} encoder for text. To capture low-level statistical biases without semantic bridges, these architectures are initialized via self-supervised contrastive learning. During fine-tuning, the backbones remain frozen; we introduce only LoRA~\cite{hu2021loralowrankadaptationlarge} modules for lightweight, task-specific forensic adaptation without compromising general representation capacity.

\noindent\textbf{Baseline Algorithms.} 
We integrate two major categories of methods for systematic comparative evaluation. The multi-modal learning (MML) category includes Concat, OGM~\cite{peng2022balancedmultimodallearningonthefly}, and DLMG~\cite{NEURIPS2024_71b17f00}, which focus on exploiting complementary inter-modal information through gradient modulation. The domain generalization (DG) category includes ERM, IRM~\cite{arjovsky2020invariantriskminimization}, Mixup~\cite{yan2020improveunsuperviseddomainadaptation}, SagNet~\cite{nam2021reducingdomaingapreducing}, IB-ERM~\cite{ahuja2022invarianceprinciplemeetsinformation}, CDANN~\cite{Li_2018_ECCV}, CondCAD~\cite{ruan2022optimalrepresentationscovariateshift}, EQRM~\cite{eastwood2023probabledomaingeneralizationquantile}, ERM++~\cite{teterwak2024ermimprovedbaselinedomain}, and URM~\cite{krishnamachari2024uniformly}. These DG methods prioritize extracting universal forgery features independent of physical properties through cross-domain invariance optimization. To ensure a fair and rigorous comparison, all evaluated algorithms are implemented within a unified framework and executed on identical computational hardware.

\noindent\textbf{Training Configuration.} 
All evaluated algorithms are uniformly deployed on top of a shared, lightweight four-layer multi-layer perceptron (MLP) detector, utilizing ReLU activations and a linear classification head to map features to a probability distribution. To ensure rigorous and fair evaluations, critical hyperparameters (including learning rate, weight decay, and MLP depth constraints) are subjected to a random search protocol spanning 3 random seeds and 9 independent trials per configuration. For detailed information, please refer to the appendix.

\section{Results \& Analysis}
\subsection{Results of Weak MAF}
To validate the existence of cross-modal invariant forgery fingerprints and their transferability to unseen modalities, we compare the generalization performance of MML-based methods (focusing on modality fusion) against DG-based methods (prioritizing modality-agnostic feature extraction).

\noindent\textbullet~\textbf{MML vs. DG Comparison.} From Fig.~\ref{fig:weakmaf}, DG-based methods consistently and substantially outperform MML-based methods across all three dataset groups. This indicates that actively extracting modality-agnostic forgery fingerprints through cross-domain invariance optimization is crucial for robust generalization. Conversely, fusion strategies relying solely on inter-modal complementary information are insufficient to overcome the "modality-binding" bottleneck. Ultimately, this validates our core premise: isolating the shared latent generative logic, rather than simply aggregating surface-level artifacts, is the definitive path toward universal multimodal defense. By breaking free from modality-specific physical priors, this approach establishes a proactive defense paradigm capable of neutralizing zero-day generative threats.

\noindent\textbullet~\textbf{The Semantic Masking Effect.} To confirm that the DG advantage does not stem from incidental semantic alignment gains, we compare performance on modality-aligned datasets (LAV-DF~\cite{cai2023reallymeanthatcontent}, FakeAVCeleb~\cite{khalid2022fakeavcelebnovelaudiovideomultimodal}) versus a semantically mismatched combination (Celeb-DF++~\cite{li2020celeb,li2025celeb} and ASVspoof5~\cite{wang2024asvspoof5crowdsourcedspeech}). Although overall accuracy decreases under semantic mismatch, the DG methods' substantial lead remains consistent, empirically confirming that universal forgery traces exist independently of semantic bridges. At the perceptor level, LanguageBind~\cite{zhu2023languagebind} yields lower forensic accuracy than ImageBind~\cite{girdhar2023imagebind} across all settings. This performance degradation stems from LanguageBind's more aggressive extraction of high-level semantic concepts, which systematically suppresses fine-grained forgery traces. We term this phenomenon \textit{semantic masking}, demonstrating that the semantic space is fundamentally distinct from the forensic space.

\noindent\textbullet~\textbf{Validation Strategies and Modality Variance.} We further assess the impact of validation mechanisms (TM, LOO, Oracle). As shown in Fig.~\ref{fig:weak_model_selection}, the LOO strategy effectively promotes cross-modal invariance learning during training, achieving stable performance closely trailing the Oracle upper bound. Furthermore, fine-grained analysis (Fig.~\ref{fig:weak_radar_charts}) reveals that the video modality, carrying highly complex spatiotemporal semantics, exhibits the strongest semantic masking effect, presenting the greatest generalization challenge. 
Nevertheless, the feature-decoupling-based MAF framework successfully transcends these physical dimensions, identifying universal fingerprints independent of macro-semantics.

\begin{table}[t]
\caption{Ablation study on the source of cross-modal forensic knowledge under the Strong MAF setting. We compare our complete framework against two degraded baselines to isolate the contribution of joint multimodal learning.}
\label{tab:ablation}
    \centering
    \small
    \begin{tabular}{l|ccc} 
        \toprule
        Method & LAV-DF & FakeAVCeleb & Celeb+ASV. \\
        \midrule
        Single-modality only & 51.7 \scriptsize{\textcolor{red}{(-8.0)}} & 52.4 \scriptsize{\textcolor{red}{(-8.5)}} & 49.9 \scriptsize{\textcolor{red}{(-7.9)}} \\
        Random initialization & 53.4 \scriptsize{\textcolor{red}{(-6.3)}} & 55.9 \scriptsize{\textcolor{red}{(-5.0)}} & 52.1 \scriptsize{\textcolor{red}{(-5.7)}} \\
        \midrule
        \rowcolor[HTML]{E5F0F7}
        Strong MAF (Ours) & 59.7 & 60.9 & 57.8 \\
        \bottomrule
    \end{tabular}
\end{table}

\subsection{Results of Strong MAF}
To verify universal forgery traces, we evaluate Strong MAF, an extreme 'perceptual island' where the test modality's encoder is physically and semantically isolated from the training phase.

\noindent\textbullet~\textbf{Proving Underlying Logic Consistency.} From Fig.~\ref{fig:strong_mg_loo1}, while the overall AUC range naturally declines compared to Weak MAF, all methods still consistently outperform random prediction. This strongly demonstrates that the statistical biases left by generative algorithms across disparate digital signals share a consistent underlying logic. It marks a critical shift from merely fitting "surface-level artifacts" toward identifying the fundamental "generative logic."

\noindent\textbullet~\textbf{Algorithmic Reversal under Isolation.} When investigating behavioral differences under extreme generalization, we observe a fascinating reversal: the substantial advantage held by DG methods in Weak MAF narrows considerably. On certain datasets, MML variants (e.g., Concat, OGM) match or marginally surpass DG methods. This occurs because complete encoder isolation makes modality-invariant features exceptionally difficult to extract, allowing MML methods to fall back on their robustness derived from known modality complementarities. Furthermore, the fully modality-agnostic UniBind achieves more stable performance than ImageBind in several settings. This exposes a technical limitation of conventional binding-centric paradigms and motivates a shift toward alignment-free, modality-agnostic strategies.

\noindent\textbullet~\textbf{The Open Challenge.} Fig.~\ref{fig:strong_model_selection} reveals that the Oracle upper-bound advantage narrows, and the TM strategy yields more stable gains under Strong MAF constraints. This suggests that capturing universal traces purely from low-level logic remains bounded under physical isolation. As reflected in the contracted radar charts (Fig.~\ref{fig:strong_radar_charts}), achieving robust forensic generalization across completely isolated "dark modalities" remains a highly non-trivial challenge, calling for future exploration into stronger self-supervised encoders or meta-learning mechanisms.

\begin{figure}[t]
    \centering
\includegraphics[width=1\linewidth]{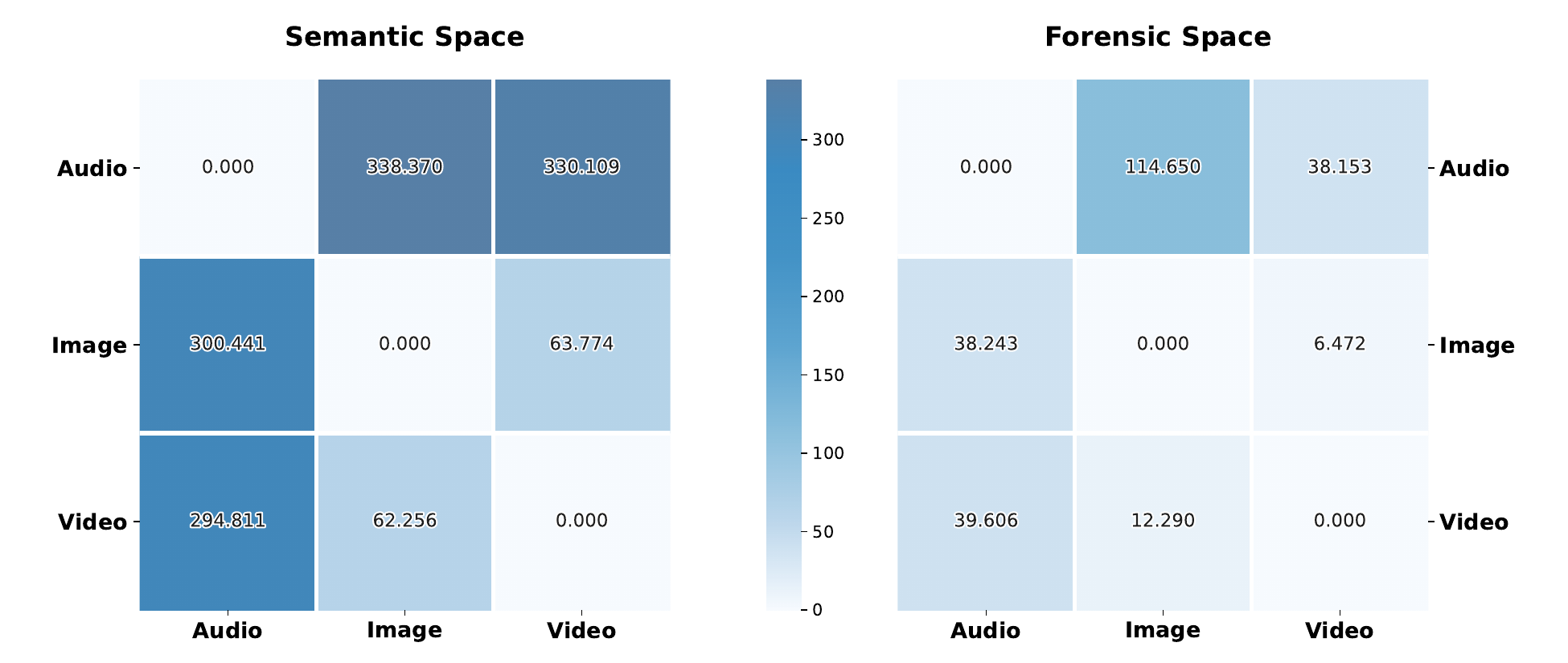}
\caption{Cross-modal distribution consistency analysis. The heatmaps illustrate the Kullback-Leibler (KL) divergence between different modalities in the semantic space (left) and the proposed forensic space (right).}
    \label{fig:kl}
\end{figure}

\subsection{Ablation Study}

To verify that MAF captures genuine latent forgery knowledge rather than overfitting to modality-specific surface features, we conduct a controlled ablation under the Strong MAF setting. We compare our full framework against two degraded baselines: (1) \textit{Single-modality linear classifier}: trained on a single aligned modality and evaluated on two isolated "dark" modalities encoded by self-supervised ViTs. (2) \textit{Randomly initialized perceptors}: abandons pretrained semantics entirely, using random ViTs to encode two training modalities and one dark testing modality. 
As shown in Table~\ref{tab:ablation}, the single-modality baseline collapses to near-random prediction on dark modalities. The randomly initialized baseline provides only marginal gains, proving that ViT's architectural inductive bias alone is vastly insufficient for cross-modal generalization. In contrast, our complete Strong MAF framework decisively outperforms both baselines across all datasets. These results suggest the importance of moving beyond single-medium memorization toward joint multimodal optimization. While the comparative advantages of our feature decoupling operator ($\mathcal{L}_{DG}$) over conventional multi-modal learning (MML) are detailed in Fig.~\ref{fig:strong_mg_loo1}, this ablation indicates that collaborative multimodal modeling serves as an effective foundation for engaging dark modalities. This strongly validates our core paradigm shift: moving beyond surface-level artifacts to capture fundamental generative logic.

\subsection{Validation of Shared Latent Knowledge}

\noindent\textbf{Cross-Modal Distribution Consistency.} To empirically verify the objective existence of shared latent forgery knowledge, we conduct a cross-modal distribution consistency analysis across the audio, image, and video modalities. We extract and compare representations from two distinct feature spaces: the original ImageBind semantic space and the proposed MAF forensic space. As illustrated in Fig.~\ref{fig:kl}, the Kullback-Leibler (KL) divergence between cross-modal forgery features substantially decreases in the forensic space, dropping by up to 90\% compared to the semantic space. This dramatic reduction confirms the emergence of a shared, modality-agnostic forgery structure. 

\noindent\textbf{Intrinsic Dimensionality Reduction.} Furthermore, as detailed in Table~\ref{tab:pca_k95}, principal component analysis (PCA) reveals that the effective intrinsic dimensionality ($k_{95}$) of the features drops sharply from over 100 in the semantic space to a highly compact range of 2--50 in the forensic space. Coupled with the substantial reduction in cross-modal KL divergence (Fig.~\ref{fig:kl}), this intrinsic dimensionality reduction provides compelling empirical evidence consistent with the hypothesis that after disentanglement, forgery features converge into a highly compact subspace stripped of modal-specific physical redundancies. Ultimately, this subspace constitutes the cross-modal invariant forgery fingerprint successfully captured by the MAF framework.

\begin{table}[t]
\caption{Intrinsic dimensionality analysis. We report the PCA effective dimensionality ($k_{95}$) of real and fake features in both the semantic and forensic spaces across three modalities. The sharp drop in $k_{95}$ empirically validates that forgery features converge into a highly compact, low-dimensional subspace.}
    \label{tab:pca_k95}
    \centering
    \small
    \setlength{\tabcolsep}{4pt}
    \begin{tabular}{ll|ccc}
        \toprule
        \textbf{Feature Space} & \textbf{Label} & \textbf{Audio} & \textbf{Image} & \textbf{Video} \\
        \midrule
        \multirow{2}{*}{Semantic Space (IB)} 
            & Real & 105 & 192 & 199 \\
            & Fake & 102 & 199 & 199 \\
        \midrule
        \rowcolor[HTML]{E5F0F7}
            & Real 
            & \textbf{3} \scriptsize{\textcolor{red}{($\downarrow$97\%)}}
            & \textbf{50} \scriptsize{\textcolor{red}{($\downarrow$74\%)}}
            & \textbf{40} \scriptsize{\textcolor{red}{($\downarrow$80\%)}} \\
        \rowcolor[HTML]{E5F0F7}
        \multirow{-2}{*}{\textbf{Forensic Space (MAF)}}
            & Fake 
            & \textbf{2} \scriptsize{\textcolor{red}{($\downarrow$98\%)}}
            & \textbf{44} \scriptsize{\textcolor{red}{($\downarrow$78\%)}}
            & \textbf{32} \scriptsize{\textcolor{red}{($\downarrow$84\%)}} \\
        \bottomrule
    \end{tabular}
\end{table}

\begin{figure}[t]
    \centering
\includegraphics[width=0.9\linewidth]{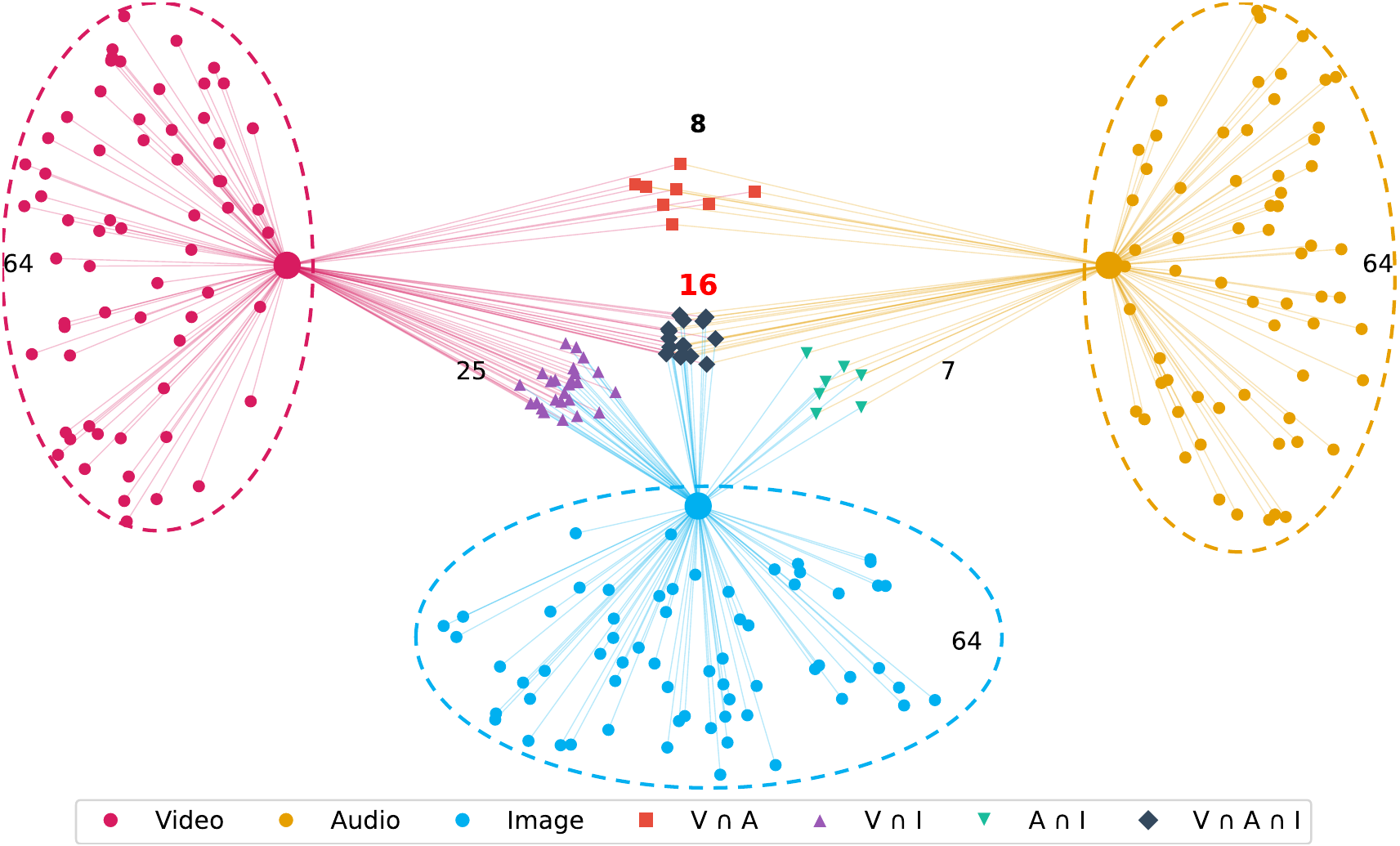}
\caption{{Cross-modal neuron co-activation.} Visualizing the top-64 activated MLP neurons per modality reveals peripheral modality-specific clusters and a robust central core of 16 neurons co-activated across all media, mechanistically confirming the extraction of universal forgery fingerprints.}
    \label{fig:sankey}
\end{figure}

\noindent\textbf{Shared Latent Forgery Knowledge.} To mechanistically validate this concept, we investigate cross-modal neuron co-activation patterns within the MAF framework. As illustrated by the network topology in Fig.~\ref{fig:sankey}, analyzing the top-64 activated neurons per modality reveals a profound structural decoupling. Exactly 16 core neurons consistently intersect across all three modalities to constitute a unified activation hub (representing the forgery essence $\mathcal{E}$), while the peripheral 48 neurons isolate modality-specific pathways (physical styles $\mathcal{S}$). This explicit neural disentanglement proves that MAF spontaneously forms a universal forensic subspace, fundamentally explaining its robust source-free domain generalization when confronting entirely unseen modalities.

\section{Discussion and Outlook}

\noindent\textbf{From Semantic Alignment to Forensic Alignment.} 
While multimodal foundation models excel at semantic binding, the pursuit of semantic invariance paradoxically diminishes forensic variance. When confronted with unseen "dark modalities," conventional detectors collapse due to the {semantic masking effect}. Specifically, pre-trained perceptors heavily entangle low-level statistical anomalies, the exact loci of forgery artifacts, with macro-level concepts. To resolve this fundamental conflict, the MAF framework orchestrates a paradigm shift from semantic to forensic alignment. By employing the universal forgery detector $\mathcal{F}(\cdot)$ as a feature disentanglement operator, MAF explicitly strips modality-specific physical style $\mathcal{S}$ from semantic representations to isolate the invariant cross-modal forgery essence $\mathcal{E}$. This orthogonalization liberates forensic fingerprints from their imaging constraints, establishing a universal forensics space independent of surface-level semantics.

\noindent\textbf{The Nature of Shared Latent Forgery Knowledge.} 
Our empirical evaluations establish a fundamental property of generative AI: the objective existence of Shared Latent Forgery Knowledge. Despite the profound physical divergence between one-dimensional audio waveforms and three-dimensional spatiotemporal videos, generative algorithms imprint inherently consistent statistical biases onto their outputs. This cross-modal consistency dictates that deepfake artifacts are not transient surface anomalies bound to specific sensors, but immutable logical flaws inherited directly from the generative mechanism. This assertion is powerfully corroborated by our Strong MAF evaluations. Even within a strictly defined "perception island", where test modalities remain completely isolated in both physical and semantic spaces, MAF successfully extrapolates these underlying algorithmic biases. By achieving robust source-free domain generalization using only unlabeled target signals for isolated self-supervision, the framework proves it transcends the mechanical memorization of artifacts to capture the universal mathematical essence of generative models.

\noindent\textbf{Practical Defense Implications.} 
Current forensic paradigms are trapped in an unsustainable and reactive arms race. The emergence of every novel forged medium dictates costly data collection and model retraining. This modality-bound approach is exceptionally vulnerable in restricted domains like medical imaging and infrared sensors, where data acquisition faces strict bottlenecks. By distilling a universal forgery fingerprint, the MAF framework establishes a highly efficient defense architecture trained entirely on known data to generalize across all unseen modalities. This mechanism effectively dismantles conventional data dependencies. Furthermore, adversaries increasingly deploy undisclosed "dark modalities" and heterogeneous sensors to bypass traditional security. Consequently, detectors tethered to specific imaging processes will rapidly become obsolete. Grounded in the DeepModal-Bench protocols, MAF introduces a proactive defense blueprint. Shifting the forensic objective from chasing transient surface artifacts to identifying immutable algorithmic logic allows MAF to future-proof multimedia security. This approach ultimately ensures robust detection capabilities against the asymmetric influx of entirely novel AI threats.

\section{Conclusion}
In this paper, we redefine the objective of multimodal deepfake detection by shifting the paradigm from modality-specific feature fusion to modality generalization. To overcome the critical "modality-binding" bottleneck, we propose the modality-agnostic forgery (MAF) framework, which explicitly decouples superficial physical styles to capture the shared latent forgery knowledge inherent to generative algorithms. Evaluated under the rigorous Weak and Strong MAF scenarios within our novel DeepModal-Bench, our approach demonstrates exceptional source-free robustness against unseen and isolated "dark modalities". Ultimately, this work empirically validates the existence of universal forgery traces, establishing a foundational, proactive defense strategy that targets fundamental generative logic rather than transient surface artifacts.

\bibliographystyle{plain}
\bibliography{ref}

\clearpage
\newpage

\begin{appendices}

\section*{Appendix}
The appendices provide additional details that support and extend the main paper. 
Appendix ~\ref{implementation} provides a detailed record of the shared MLP architecture, LoRA fine-tuning configurations, and training hyperparameter settings, ensuring high reproducibility of the experimental environment. Appendix ~\ref{dataset} elaborates on the audio-visual frame extraction workflow, the textual modality, and the rigorous subject-independent data partitioning protocols. Appendix ~\ref{algrithim} outlines the technical details of the integrated multi-modal balancing and domain generalization algorithms within the framework. Appendix ~\ref{modalselection} provides a comprehensive introduction to the three model selection protocols used for evaluating generalization performance. Appendix ~\ref{detaileddata} presents the complete set of detailed experimental results across various benchmarks and settings.
Appendix~\ref{app:theory} provides a formal theoretical justification for the efficacy of cross-modal feature disentanglement and the robust generalization to unseen "dark modalities", grounded in the information bottleneck principle and domain adaptation theory.


\section{Implementation Details}
\label{implementation}

\noindent\textbf{Feature Processor.}
In the feature extraction stage, we design a shared four-layer MLP with a symmetric bottleneck structure as the feature processor. The input dimension $D$ is set to 1024 to match the output of encoders (e.g., ImageBind~\cite{girdhar2023imagebind}). The network consists of alternating linear layers and ReLU activation functions, following a dimensionality transformation path of $1024 \to 512 \to 256 \to 512 \to 1024$.

\noindent\textbf{Hyperparameter Optimization.}
To ensure fairness and optimal performance across all evaluated algorithms, we adopt the large-scale random search protocol from ModalBed~\cite{liu2025towards} to determine the best hyperparameter configuration for each model. Detailed configurations are provided in the following Table~\ref{tab:hyperparameters}.
Specifically, the search space for hyperparameters is defined as either continuous distributions or discrete sets, primarily utilizing the uniform distribution $\mathcal{U}(a, b)$ and the log-uniform distribution $10^{\mathcal{U}(a, b)}$. The search scope covers both general training parameters and algorithm-specific parameters: \textit{General Training Parameters:} Based on the distinct characteristics of multi-modal learning (MML) and domain generalization (DG) algorithms, we define different search intervals and default values for learning rates. Additionally, the batch size and weight decay are systematically sampled within predefined ranges. \textit{Algorithm-specific Parameters:} For specific algorithms such as IRM~\cite{arjovsky2020invariantriskminimization}, Mixup~\cite{yan2020improveunsuperviseddomainadaptation}, CDANN~\cite{Li_2018_ECCV}, and EQRM~\cite{eastwood2023probabledomaingeneralizationquantile}, we provide targeted spatial definitions for their core hyperparameters, including penalty coefficients ($\lambda$), annealing steps, and scaling factors ($\alpha$).
Through this random search approach, we are able to capture the performance upper bound of each model within an extensive parameter space, thereby providing a reliable benchmark for evaluating multi-modal generalization capabilities.

\begin{table*}[t]
\centering
\caption{Hyperparameter Search Space and Default Values}
\label{tab:hyperparameters}
\begin{tabular}{l|l|lll}
\toprule
\textbf{Type} & \textbf{Algorithm} & \textbf{Hyperparameter} & \textbf{Default Value} & \textbf{Distribution} \\ \midrule
 & & batchsize & 32 & $2^{\mathcal{U}(3, 5.5)}$ \\ \midrule
\multirow{5}{*}{MML} & \multirow{4}{*}{All} & lr & 0.001 & $10^{\mathcal{U}(-4, -2)}$ \\
 & & momentum & 0.9 & $\mathcal{U}(0.85, 0.95)$ \\
 & & weight decay & 0.0001 & $10^{\mathcal{U}(-6, -2)}$ \\
 & & patience & 70 & $\mathcal{U}(60, 80)$ \\ \cmidrule{2-5}
 & OGM~\cite{peng2022balancedmultimodallearningonthefly} & alpha & 0.1 & $\mathcal{U}(0.1, 0.3)$ \\ \midrule
\multirow{23}{*}{DG} & \multirow{2}{*}{All} & lr & 0.00005 & $10^{\mathcal{U}(-5, -3.5)}$ \\
 & & weight decay & 0 & $10^{\mathcal{U}(-6, -2)}$ \\ \cmidrule{2-5}
 & \multirow{2}{*}{IRM~\cite{arjovsky2020invariantriskminimization}} & lambda & 100 & $10^{\mathcal{U}(-1, 5)}$ \\
 & & iterations of penalty annealing & 500 & $10^{\mathcal{U}(0, 4)}$ \\ \cmidrule{2-5}
 & Mixup~\cite{yan2020improveunsuperviseddomainadaptation} & alpha & 0.2 & $10^{\mathcal{U}(-1, 1)}$ \\ \cmidrule{2-5}
 & \multirow{5}{*}{CDANN~\cite{Li_2018_ECCV}} & lambda & 1.0 & $10^{\mathcal{U}(-2, -2)}$ \\
 & & discriminator weight decay & 0 & $10^{\mathcal{U}(-6, -2)}$ \\
 & & discriminator steps & 1 & $2^{\mathcal{U}(0, 3)}$ \\
 & & gradient penalty & 0 & $10^{\mathcal{U}(-2, 1)}$ \\
 & & adam beta1 & 0.5 & $\{0, 0.5\}$ \\ \cmidrule{2-5}
 & SagNet~\cite{nam2021reducingdomaingapreducing} & weight of adversarial loss & 0.1 & $10^{\mathcal{U}(-2, 1)}$ \\ \cmidrule{2-5}
 & \multirow{2}{*}{IB\_ERM~\cite{ahuja2022invarianceprinciplemeetsinformation}} & lambda & 100 & $10^{\mathcal{U}(-1, 5)}$ \\
 & & iterations of penalty annealing & 500 & $10^{\mathcal{U}(0, 4)}$ \\ \cmidrule{2-5}
 & \multirow{2}{*}{CondCAD~\cite{ruan2022optimalrepresentationscovariateshift}} & lambda & 0.1 & $\{0.0001, \dots, 100\}$ \\
 & & temperature & 0.1 & $\{0.05, 0.1\}$ \\ \cmidrule{2-5}
 & \multirow{3}{*}{EQRM~\cite{eastwood2023probabledomaingeneralizationquantile}} & lr & 0.000001 & $10^{\mathcal{U}(-7, -5)}$ \\
 & & quantile & 0.75 & $\mathcal{U}(0.5, 0.99)$ \\
 & & iterations of burn-in & 2500 & $10^{\mathcal{U}(2.5, 3.5)}$ \\ \cmidrule{2-5}
 & ERM++~\cite{teterwak2024ermimprovedbaselinedomain} & lr & 0.00005 & $10^{\mathcal{U}(-5, -3.5)}$ \\ \cmidrule{2-5}
 & URM~\cite{krishnamachari2024uniformly} & lambda & 0.1 & $\mathcal{U}(0, 0.2)$ \\ \bottomrule
\end{tabular}
\end{table*}

\noindent\textbf{Strong MAF Configuration.}
Under the Strong MAF setting, we incorporate LoRA modules for parameter-efficient fine-tuning. Notably, LoRA hyperparameters are excluded from the random search. We fix the rank $r$ to 8, the scaling factor $\alpha$ to 32, and apply a dropout rate of 0.05. These values are selected based on standard practices for lightweight adaptation to align forgery features while preserving the backbone's general representation capabilities.

\begin{table*}[t]
\centering
\caption{Classification performance (mean $\pm$ std) under the Weak MAF setting, using test-modality validation set (oracle) for model selection.}
\label{tab:weak_mg_test_modality_oracle}
\resizebox{\textwidth}{!}{ 
\begin{tabular}{l | ll | cccc | cccc | cccc}
\toprule
\multirow{2}{*}{\textbf{Perceptor}} & \multicolumn{2}{l|}{\multirow{2}{*}{\textbf{Method}}} & \multicolumn{4}{c|}{\textbf{LAV-DF~\cite{cai2023reallymeanthatcontent}}} & \multicolumn{4}{c|}{\textbf{Fakeavcele~\cite{khalid2022fakeavcelebnovelaudiovideomultimodal}}} & \multicolumn{4}{c}{\textbf{Cele+Asv~\cite{wang2024asvspoof5crowdsourcedspeech, li2025celeb}}} \\
\cmidrule(lr){4-7} \cmidrule(lr){8-11} \cmidrule(lr){12-15}
& & & \textbf{Vid} & \textbf{Aud} & \textbf{Img} & \textbf{Avg} & \textbf{Vid} & \textbf{Aud} & \textbf{Img} & \textbf{Avg} & \textbf{Vid} & \textbf{Aud} & \textbf{Img} & \textbf{Avg} \\
\midrule
\midrule

\multirow{13}{*}{ImageBind~\cite{girdhar2023imagebind}} & \multirow{3}{*}{MML} & Concat & $63.12_{\pm0.41}$ & $62.32_{\pm0.22}$ & $63.42_{\pm0.88}$ & 62.95 & $64.59_{\pm0.15}$ & $67.05_{\pm1.34}$ & $69.42_{\pm0.55}$ & 67.02 & $58.14_{\pm0.72}$ & $59.10_{\pm0.09}$ & $59.27_{\pm0.43}$ & 58.84 \\
 &  & OGM~\cite{peng2022balancedmultimodallearningonthefly} & $60.66_{\pm0.12}$ & $61.48_{\pm0.64}$ & $59.92_{\pm0.37}$ & 60.69 & $62.08_{\pm0.91}$ & $64.47_{\pm0.56}$ & $63.42_{\pm0.82}$ & 63.32 & $56.34_{\pm0.28}$ & $57.17_{\pm0.11}$ & $57.94_{\pm0.74}$ & 57.15 \\
 &  & DLMG~\cite{NEURIPS2024_71b17f00} & $61.64_{\pm0.53}$ & $61.25_{\pm0.49}$ & $59.83_{\pm1.12}$ & 60.91 & $64.75_{\pm0.04}$ & $64.87_{\pm0.32}$ & $63.24_{\pm0.71}$ & 64.29 & $58.56_{\pm0.44}$ & $57.34_{\pm0.85}$ & $58.67_{\pm0.21}$ & 58.19 \\
\cmidrule{2-15}
 & \multirow{10}{*}{DG} & ERM & $70.02_{\pm0.66}$ & $66.86_{\pm0.19}$ & $67.61_{\pm0.33}$ & 68.16 & $62.00_{\pm0.52}$ & $64.60_{\pm0.78}$ & $77.12_{\pm0.25}$ & 67.91 & $60.12_{\pm0.91}$ & $62.02_{\pm0.44}$ & $64.22_{\pm0.12}$ & 62.12 \\
 &  & IRM~\cite{arjovsky2020invariantriskminimization} & $68.15_{\pm0.42}$ & $66.45_{\pm0.08}$ & $70.05_{\pm1.67}$ & 68.22 & $63.40_{\pm0.62}$ & $68.35_{\pm0.55}$ & $74.15_{\pm0.33}$ & 68.63 & $60.15_{\pm0.27}$ & $61.05_{\pm0.81}$ & $62.75_{\pm0.59}$ & 61.32 \\
 &  & Mixup~\cite{yan2020improveunsuperviseddomainadaptation} & $71.96_{\pm0.31}$ & $67.14_{\pm0.14}$ & $67.99_{\pm0.92}$ & 69.03 & $63.48_{\pm0.03}$ & $68.82_{\pm0.44}$ & $76.03_{\pm0.71}$ & 69.44 & $61.96_{\pm0.62}$ & $62.76_{\pm0.18}$ & $64.97_{\pm0.88}$ & 63.23 \\
 &  & CDANN~\cite{Li_2018_ECCV} & $70.88_{\pm0.55}$ & $66.18_{\pm0.72}$ & $69.79_{\pm0.22}$ & 68.95 & $62.79_{\pm0.41}$ & $68.15_{\pm0.11}$ & $78.00_{\pm1.41}$ & 69.65 & $62.28_{\pm0.34}$ & $61.48_{\pm0.67}$ & $63.58_{\pm0.09}$ & 62.45 \\
 &  & SagNet~\cite{nam2021reducingdomaingapreducing} & $70.84_{\pm0.49}$ & $66.94_{\pm0.63}$ & $68.08_{\pm0.18}$ & 68.62 & $62.42_{\pm0.55}$ & $65.95_{\pm0.29}$ & $75.55_{\pm0.77}$ & 67.97 & $62.13_{\pm0.02}$ & $62.55_{\pm1.02}$ & $63.64_{\pm0.34}$ & 62.77 \\
 &  & IB\_ERM~\cite{ahuja2022invarianceprinciplemeetsinformation} & $69.76_{\pm0.81}$ & $67.52_{\pm0.54}$ & $69.58_{\pm0.31}$ & 68.95 & $64.02_{\pm0.19}$ & $64.65_{\pm0.44}$ & $75.22_{\pm1.82}$ & 67.96 & $60.67_{\pm0.66}$ & $64.28_{\pm0.25}$ & $67.21_{\pm0.55}$ & 64.05 \\
 &  & CondCAD~\cite{ruan2022optimalrepresentationscovariateshift} & $70.24_{\pm0.24}$ & $66.46_{\pm0.89}$ & $67.93_{\pm0.11}$ & 68.21 & $63.21_{\pm0.47}$ & $64.94_{\pm0.53}$ & $75.39_{\pm0.61}$ & 67.85 & $63.89_{\pm0.38}$ & $63.35_{\pm0.42}$ & $61.98_{\pm0.15}$ & 63.07 \\
 &  & EQRM~\cite{eastwood2023probabledomaingeneralizationquantile} & $71.28_{\pm0.34}$ & $66.06_{\pm0.05}$ & $68.84_{\pm0.72}$ & 68.73 & $62.86_{\pm0.66}$ & $66.89_{\pm0.59}$ & $79.05_{\pm0.24}$ & 69.60 & $61.97_{\pm0.18}$ & $62.00_{\pm0.31}$ & $62.50_{\pm1.08}$ & 62.16 \\
 &  & ERM++~\cite{teterwak2024ermimprovedbaselinedomain} & $71.76_{\pm0.44}$ & $66.74_{\pm0.22}$ & $69.41_{\pm0.61}$ & 69.30 & $62.70_{\pm0.81}$ & $65.17_{\pm1.22}$ & $77.64_{\pm0.41}$ & 68.50 & $60.51_{\pm0.25}$ & $61.34_{\pm0.04}$ & $63.04_{\pm0.55}$ & 61.63 \\
 &  & URM~\cite{krishnamachari2024uniformly} & $72.19_{\pm0.15}$ & $67.31_{\pm0.77}$ & $68.63_{\pm0.39}$ & 69.38 & $63.75_{\pm0.52}$ & $67.48_{\pm0.18}$ & $77.83_{\pm0.66}$ & 69.69 & $61.43_{\pm0.43}$ & $61.75_{\pm1.32}$ & $66.05_{\pm0.21}$ & 63.08 \\

\midrule

\multirow{13}{*}{LanguageBind~\cite{zhu2023languagebind}} & \multirow{3}{*}{MML} & Concat & $62.88_{\pm0.11}$ & $61.96_{\pm0.44}$ & $62.72_{\pm0.28}$ & 62.52 & $63.92_{\pm0.67}$ & $66.32_{\pm0.31}$ & $69.62_{\pm0.09}$ & 66.62 & $58.34_{\pm0.55}$ & $58.30_{\pm0.15}$ & $58.82_{\pm0.71}$ & 58.49 \\
 &  & OGM~\cite{peng2022balancedmultimodallearningonthefly} & $60.03_{\pm0.32}$ & $60.82_{\pm0.56}$ & $61.21_{\pm0.81}$ & 60.69 & $64.11_{\pm0.24}$ & $62.73_{\pm0.18}$ & $63.42_{\pm1.51}$ & 63.42 & $57.58_{\pm0.43}$ & $57.42_{\pm0.66}$ & $57.54_{\pm0.25}$ & 57.51 \\
 &  & DLMG~\cite{NEURIPS2024_71b17f00} & $61.72_{\pm0.04}$ & $60.56_{\pm0.88}$ & $60.38_{\pm0.39}$ & 60.89 & $64.22_{\pm0.12}$ & $63.48_{\pm0.47}$ & $64.17_{\pm0.72}$ & 63.96 & $58.60_{\pm0.53}$ & $57.51_{\pm0.21}$ & $58.05_{\pm0.05}$ & 58.05 \\
\cmidrule{2-15}
 & \multirow{10}{*}{DG} & ERM & $67.77_{\pm0.41}$ & $61.94_{\pm0.62}$ & $67.04_{\pm0.34}$ & 65.58 & $62.33_{\pm0.18}$ & $65.32_{\pm0.49}$ & $71.82_{\pm1.15}$ & 66.49 & $59.65_{\pm0.81}$ & $60.99_{\pm0.22}$ & $60.92_{\pm0.55}$ & 60.52 \\
 &  & IRM~\cite{arjovsky2020invariantriskminimization} & $67.46_{\pm0.53}$ & $64.75_{\pm0.25}$ & $67.14_{\pm0.08}$ & 66.45 & $61.42_{\pm0.43}$ & $67.85_{\pm0.77}$ & $70.05_{\pm0.31}$ & 66.44 & $60.72_{\pm0.15}$ & $58.55_{\pm0.39}$ & $60.55_{\pm0.64}$ & 59.94 \\
 &  & Mixup~\cite{yan2020improveunsuperviseddomainadaptation} & $67.85_{\pm0.38}$ & $61.83_{\pm0.71}$ & $67.96_{\pm0.12}$ & 65.88 & $60.23_{\pm0.66}$ & $67.19_{\pm0.52}$ & $78.16_{\pm0.28}$ & 68.53 & $60.62_{\pm1.02}$ & $60.56_{\pm0.44}$ & $61.76_{\pm0.11}$ & 60.98 \\
 &  & CDANN~\cite{Li_2018_ECCV} & $68.14_{\pm0.82}$ & $63.31_{\pm0.03}$ & $66.92_{\pm0.59}$ & 66.12 & $60.88_{\pm0.19}$ & $68.05_{\pm0.88}$ & $75.05_{\pm0.24}$ & 67.99 & $60.30_{\pm0.53}$ & $61.66_{\pm0.61}$ & $60.70_{\pm0.47}$ & 60.89 \\
 &  & SagNet~\cite{nam2021reducingdomaingapreducing} & $68.98_{\pm0.22}$ & $62.91_{\pm0.41}$ & $68.41_{\pm1.44}$ & 66.77 & $61.18_{\pm0.34}$ & $66.48_{\pm0.29}$ & $75.68_{\pm0.81}$ & 67.78 & $61.08_{\pm0.55}$ & $59.78_{\pm0.04}$ & $62.29_{\pm0.66}$ & 61.05 \\
 &  & IB\_ERM~\cite{ahuja2022invarianceprinciplemeetsinformation}~\cite{ahuja2022invarianceprinciplemeetsinformation} & $69.75_{\pm0.64}$ & $63.20_{\pm0.72}$ & $66.96_{\pm0.15}$ & 66.64 & $62.28_{\pm0.49}$ & $66.88_{\pm0.54}$ & $78.24_{\pm0.22}$ & 69.13 & $61.33_{\pm0.32}$ & $62.56_{\pm1.11}$ & $62.91_{\pm0.31}$ & 62.27 \\
 &  & CondCAD~\cite{ruan2022optimalrepresentationscovariateshift} & $68.69_{\pm0.11}$ & $61.66_{\pm0.37}$ & $65.80_{\pm0.53}$ & 65.38 & $61.37_{\pm0.08}$ & $66.94_{\pm0.82}$ & $78.39_{\pm0.41}$ & 68.90 & $60.98_{\pm0.18}$ & $62.52_{\pm0.55}$ & $61.39_{\pm0.25}$ & 61.63 \\
 &  & EQRM~\cite{eastwood2023probabledomaingeneralizationquantile} & $68.75_{\pm0.71}$ & $63.51_{\pm0.05}$ & $68.38_{\pm0.61}$ & 66.88 & $60.90_{\pm1.29}$ & $68.82_{\pm0.38}$ & $74.22_{\pm0.44}$ & 67.98 & $61.04_{\pm0.21}$ & $58.63_{\pm0.67}$ & $60.92_{\pm0.09}$ & 60.20 \\
 &  & ERM++~\cite{teterwak2024ermimprovedbaselinedomain} & $68.01_{\pm0.42}$ & $63.52_{\pm1.02}$ & $67.99_{\pm0.33}$ & 66.51 & $61.27_{\pm0.62}$ & $67.88_{\pm0.15}$ & $73.14_{\pm0.49}$ & 67.43 & $59.96_{\pm0.88}$ & $59.43_{\pm0.27}$ & $61.99_{\pm0.52}$ & 60.46 \\
 &  & URM~\cite{krishnamachari2024uniformly} & $68.23_{\pm0.29}$ & $62.53_{\pm0.55}$ & $68.26_{\pm0.18}$ & 66.34 & $61.35_{\pm0.03}$ & $66.83_{\pm0.71}$ & $78.43_{\pm1.08}$ & 68.87 & $60.08_{\pm0.44}$ & $61.65_{\pm0.31}$ & $61.41_{\pm0.59}$ & 61.05 \\

\midrule

\multirow{13}{*}{UniBind~\cite{lyu2024unibind}} & \multirow{3}{*}{MML} & Concat & $62.84_{\pm0.55}$ & $63.80_{\pm0.11}$ & $62.82_{\pm0.44}$ & 63.15 & $63.92_{\pm0.67}$ & $65.82_{\pm0.21}$ & $66.62_{\pm0.39}$ & 65.45 & $59.21_{\pm0.72}$ & $59.25_{\pm1.15}$ & $58.09_{\pm0.34}$ & 58.85 \\
 &  & OGM~\cite{peng2022balancedmultimodallearningonthefly} & $60.00_{\pm0.22}$ & $60.82_{\pm0.59}$ & $61.24_{\pm0.04}$ & 60.69 & $64.13_{\pm0.82}$ & $63.08_{\pm1.22}$ & $64.11_{\pm0.47}$ & 63.77 & $56.85_{\pm0.12}$ & $57.88_{\pm0.31}$ & $57.95_{\pm0.66}$ & 57.56 \\
 &  & DLMG~\cite{NEURIPS2024_71b17f00} & $61.71_{\pm0.61}$ & $61.70_{\pm0.09}$ & $60.39_{\pm0.88}$ & 61.27 & $64.16_{\pm0.18}$ & $63.13_{\pm0.33}$ & $63.85_{\pm0.52}$ & 63.71 & $58.87_{\pm0.41}$ & $58.23_{\pm0.05}$ & $57.39_{\pm0.28}$ & 58.16 \\
\cmidrule{2-15}
 & \multirow{10}{*}{DG} & ERM & $71.07_{\pm0.15}$ & $66.03_{\pm0.34}$ & $68.86_{\pm0.71}$ & 68.65 & $63.26_{\pm1.41}$ & $62.92_{\pm0.55}$ & $77.32_{\pm0.22}$ & 67.83 & $60.22_{\pm0.64}$ & $61.63_{\pm0.81}$ & $65.12_{\pm0.49}$ & 62.32 \\
 &  & IRM~\cite{arjovsky2020invariantriskminimization} & $67.95_{\pm0.49}$ & $66.95_{\pm0.21}$ & $69.85_{\pm0.03}$ & 68.25 & $62.65_{\pm0.12}$ & $66.85_{\pm0.72}$ & $74.35_{\pm0.53}$ & 67.95 & $60.02_{\pm0.88}$ & $61.15_{\pm0.18}$ & $62.75_{\pm0.32}$ & 61.31 \\
 &  & Mixup~\cite{yan2020improveunsuperviseddomainadaptation} & $71.95_{\pm0.39}$ & $65.94_{\pm1.62}$ & $69.30_{\pm0.41}$ & 69.06 & $61.65_{\pm0.55}$ & $64.38_{\pm0.08}$ & $76.81_{\pm0.34}$ & 67.61 & $61.86_{\pm0.11}$ & $63.06_{\pm0.67}$ & $64.81_{\pm0.52}$ & 63.24 \\
 &  & CDANN~\cite{Li_2018_ECCV} & $71.33_{\pm0.24}$ & $65.87_{\pm0.81}$ & $67.94_{\pm0.53}$ & 68.38 & $61.98_{\pm0.66}$ & $66.45_{\pm1.02}$ & $78.38_{\pm0.15}$ & 68.94 & $61.88_{\pm0.43}$ & $61.24_{\pm0.55}$ & $62.18_{\pm0.31}$ & 61.77 \\
 &  & SagNet~\cite{nam2021reducingdomaingapreducing} & $71.77_{\pm0.05}$ & $66.53_{\pm0.55}$ & $68.18_{\pm0.28}$ & 68.83 & $61.32_{\pm0.44}$ & $64.93_{\pm0.11}$ & $76.90_{\pm0.71}$ & 67.72 & $62.33_{\pm0.09}$ & $61.44_{\pm0.41}$ & $65.57_{\pm0.82}$ & 63.11 \\
 &  & IB\_ERM~\cite{ahuja2022invarianceprinciplemeetsinformation} & $71.05_{\pm1.12}$ & $66.47_{\pm0.38}$ & $68.19_{\pm0.19}$ & 68.57 & $62.53_{\pm0.53}$ & $64.39_{\pm0.61}$ & $77.11_{\pm0.25}$ & 68.01 & $63.01_{\pm0.55}$ & $64.48_{\pm0.04}$ & $66.58_{\pm0.49}$ & 64.69 \\
 &  & CondCAD~\cite{ruan2022optimalrepresentationscovariateshift} & $70.09_{\pm0.31}$ & $66.26_{\pm0.72}$ & $68.36_{\pm0.14}$ & 68.24 & $63.10_{\pm0.42}$ & $65.94_{\pm0.39}$ & $76.99_{\pm0.55}$ & 68.68 & $63.33_{\pm0.15}$ & $61.56_{\pm0.88}$ & $63.39_{\pm0.64}$ & 62.76 \\
 &  & EQRM~\cite{eastwood2023probabledomaingeneralizationquantile} & $71.52_{\pm0.27}$ & $65.76_{\pm0.54}$ & $69.54_{\pm0.08}$ & 68.94 & $62.11_{\pm0.21}$ & $65.48_{\pm1.32}$ & $78.28_{\pm0.44}$ & 68.62 & $61.52_{\pm0.53}$ & $61.45_{\pm0.25}$ & $63.52_{\pm1.08}$ & 62.16 \\
 &  & ERM++~\cite{teterwak2024ermimprovedbaselinedomain} & $70.84_{\pm0.61}$ & $66.67_{\pm0.33}$ & $68.49_{\pm0.55}$ & 68.67 & $63.52_{\pm0.18}$ & $63.65_{\pm0.71}$ & $76.23_{\pm0.12}$ & 67.80 & $61.19_{\pm0.42}$ & $60.92_{\pm0.66}$ & $63.14_{\pm0.34}$ & 61.75 \\
 &  & URM~\cite{krishnamachari2024uniformly} & $71.11_{\pm0.81}$ & $67.05_{\pm0.11}$ & $68.63_{\pm0.49}$ & 68.93 & $62.63_{\pm0.55}$ & $64.50_{\pm0.22}$ & $76.69_{\pm1.41}$ & 67.94 & $60.40_{\pm0.28}$ & $62.63_{\pm0.52}$ & $64.53_{\pm0.39}$ & 62.52 \\

\bottomrule
\end{tabular}
}
\end{table*}

\section{Dataset Preprocessing}
\label{dataset}
\noindent\textbf{Visual Modality and Video Processing.}
The visual modality is generated by discretizing raw video streams into static frames. Using the ffmpeg tool, we extract 4 frames from each video clip, enforcing a minimum temporal interval of 0.4 seconds to eliminate information redundancy. For partially forged datasets such as LAV-DF~\cite{cai2023reallymeanthatcontent}, we implement a precise slicing logic based on the \textit{Fake Periods} defined in the metadata. Specifically, for authentic samples, signals are extracted from the entire duration; for forged samples, we lock the extraction to the specific manipulated segments to ensure the model captures core forensic traces rather than irrelevant background information.

\noindent\textbf{Textual Modality and Preprocessing Standards.}
The inclusion of the textual modality and the T5 encoder represents an architectural extension designed to verify the generalization potential of MAF across semantically correlated modalities. While textual input remained inactive during the primary benchmarking of this study, the framework is designed for seamless integration with human-annotated transcripts or text automatically generated via Whisper ASR. 
For cross-modal standardization, we resample all audio signals to 44.1 kHz and convert them into Mel-spectrograms with 224 frequency bins. This provides a unified representation that aligns with the visual frame extraction interval of 0.4 seconds.

\noindent\textbf{Data Partitioning and Leakage Prevention.}
We follow a rigorous isolation protocol to prevent identity (ID) leakage. For datasets with official splits, we extract samples directly from the designated partitions. For others, we first perform a random split at the video or identity level with an 8:2 ratio. 
Crucially, the frame extraction is executed independently within each subset; for instance, test frames are derived exclusively from the 20\% test partition. By partitioning videos by ID before feature extraction, we ensure that all derivative data from a single source (audio-visual clips and frames) are strictly confined to the same split. This physical and logical separation prevents frame-level data leakage and ensures the model captures universal forgery knowledge rather than specific identity-related artifacts.

\section{Details of Selected Algorithms}
\label{algrithim}

\noindent\textbf{Origin.} Origin serves as the control group, utilizing a basic linear classifier to evaluate the raw generalization ability of extracted features.

\noindent\textbf{Multimodal Learning (MML).}

\noindent\textbf{(1) Concat} adopts a straightforward feature concatenation approach to preserve complete information but is prone to imbalanced convergence across modalities. 

\noindent\textbf{(2) OGM~\cite{peng2022balancedmultimodallearningonthefly}} introduces an on-the-fly gradient modulation mechanism that monitors each branch's contribution in real-time, effectively preventing dominant modalities from suppressing weaker ones and achieving a dynamic balance in learning rates.

\noindent\textbf{(3) DLMG~\cite{NEURIPS2024_71b17f00}} is an algorithm that extracts the cross-modal shared forgery essence by separating modality-specific surface style biases, thereby enhancing the model's detection capability for unknown ``dark modalities''.

\noindent\textbf{Domain Generalization (DG).}

 \noindent\textbf{(1) ERM~\cite{ahuja2022invarianceprinciplemeetsinformation}} establishes a baseline by minimizing the average loss across source domains. 
 
 \noindent\textbf{(2) IRM~\cite{arjovsky2020invariantriskminimization}} utilizes a gradient penalty term to extract invariant causal features that remain consistent across environments. 
 
 \noindent\textbf{(3) CDANN~\cite{Li_2018_ECCV}} employs conditional adversarial networks to discard domain-specific features while preserving class-relevant information. 
 
 \noindent\textbf{(4) Mixup~\cite{yan2020improveunsuperviseddomainadaptation}} extends the model's out-of-distribution capability via linear interpolation of cross-domain samples.
 
 \noindent\textbf{(5) ERM++~\cite{teterwak2024ermimprovedbaselinedomain}} incorporates weight moving average techniques to improve training robustness through optimization smoothing.

 \noindent\textbf{(6) SagNet~\cite{nam2021reducingdomaingapreducing}} focuses on the essence of forensics by decoupling style and content to eliminate environmental style bias. 
 
 \noindent\textbf{(7) IB\_ERM~\cite{ahuja2022invarianceprinciplemeetsinformation}} compresses redundant representations using the information bottleneck principle.
 
 \noindent\textbf{(8) CondCAD~\cite{ruan2022optimalrepresentationscovariateshift}} strengthens domain bottlenecks at the conditional distribution level by combining contrastive learning with adversarial mechanisms. Regarding risk management.
 
\noindent\textbf{(9) EQRM~\cite{eastwood2023probabledomaingeneralizationquantile}} focuses on optimizing the upper quantiles of the loss distribution to mitigate "worst-case" risks.
 
 \noindent\textbf{(10) URM~\cite{krishnamachari2024uniformly}} pursues a uniform distribution of risks across domains through adversarial constraints, ensuring robust defense performance even in complex and extreme forgery scenarios.

\section{Details of Modal Selection}
\label{modalselection}

To evaluate the performance of the MAF framework in generalization tasks, this study adopts three categories of model selection protocols to identify optimal hyperparameters:

 \noindent\textbf{Training-Modality (TM) Validation Set.} Based on the Independent and Identically Distributed (IID) assumption, a validation set is constructed directly from the training modalities. This approach aims to evaluate the model's basic fitting capability under known distributions by maximizing accuracy on this set.

 \noindent\textbf{Leave-One-Modality-Out (LOO) Cross-Validation.} This method simulates an unknown "dark modality" environment by cyclically removing a single modality as the validation set. It aims to select configurations that effectively capture cross-modal "meta-distribution" features. The model is then retrained on all training modalities to achieve the best generalization gains.

 \noindent\textbf{Test-Modality Validation Set (Oracle).} Serving as an "Oracle" baseline to measure the upper bound of generalization performance, this protocol directly uses the test distribution for model selection. All models are trained for a fixed number of steps without early stopping, using only the final checkpoint for evaluation. This provides a reference coordinate for assessing generalization limits.

\begin{table*}[t]
\centering
\caption{Classification performance (mean $\pm$ std) under the Weak MAF setting, using leave-one-modality-out cross-validation for model selection.}
\label{tab:weak_leave_one_out_custom_random}
\resizebox{\textwidth}{!}{ 
\begin{tabular}{l | ll | cccc | cccc | cccc}
\toprule
\multirow{2}{*}{\textbf{Perceptor}} & \multicolumn{2}{l|}{\multirow{2}{*}{\textbf{Method}}} & \multicolumn{4}{c|}{\textbf{LAV-DF~\cite{cai2023reallymeanthatcontent}}} & \multicolumn{4}{c|}{\textbf{Fakeavcele~\cite{khalid2022fakeavcelebnovelaudiovideomultimodal}}} & \multicolumn{4}{c}{\textbf{Cele+Asv~\cite{wang2024asvspoof5crowdsourcedspeech, li2025celeb}}} \\
\cmidrule(lr){4-7} \cmidrule(lr){8-11} \cmidrule(lr){12-15}
& & & \textbf{Vid} & \textbf{Aud} & \textbf{Img} & \textbf{Avg} & \textbf{Vid} & \textbf{Aud} & \textbf{Img} & \textbf{Avg} & \textbf{Vid} & \textbf{Aud} & \textbf{Img} & \textbf{Avg} \\
\midrule
\midrule

\multirow{13}{*}{ImageBind~\cite{girdhar2023imagebind}} & \multirow{3}{*}{MML} & Concat & $61.20_{\pm1.93}$ & $60.65_{\pm1.22}$ & $60.93_{\pm1.46}$ & 60.93 & $63.38_{\pm0.36}$ & $64.78_{\pm2.19}$ & $64.46_{\pm1.33}$ & 64.21 & $58.44_{\pm1.89}$ & $57.82_{\pm1.42}$ & $57.40_{\pm1.94}$ & 57.89 \\
 &  & OGM~\cite{peng2022balancedmultimodallearningonthefly} & $60.00_{\pm0.46}$ & $60.82_{\pm0.34}$ & $61.24_{\pm0.35}$ & 60.69 & $61.69_{\pm0.43}$ & $65.59_{\pm0.39}$ & $63.37_{\pm0.26}$ & 63.55 & $56.91_{\pm1.02}$ & $56.12_{\pm1.41}$ & $56.87_{\pm0.89}$ & 56.63 \\
 &  & DLMG~\cite{NEURIPS2024_71b17f00} & $61.66_{\pm1.24}$ & $59.72_{\pm0.99}$ & $60.49_{\pm1.00}$ & 60.62 & $64.44_{\pm1.23}$ & $63.45_{\pm0.16}$ & $63.06_{\pm0.82}$ & 63.65 & $57.30_{\pm0.88}$ & $58.01_{\pm1.65}$ & $57.39_{\pm1.41}$ & 57.57 \\
\cmidrule{2-15}
 & \multirow{10}{*}{DG} & ERM & $71.03_{\pm0.28}$ & $66.13_{\pm0.95}$ & $68.01_{\pm0.89}$ & 68.39 & $62.07_{\pm0.14}$ & $63.60_{\pm0.27}$ & $72.97_{\pm1.02}$ & 66.21 & $58.40_{\pm1.11}$ & $61.81_{\pm1.01}$ & $62.36_{\pm1.27}$ & 60.86 \\
 &  & IRM~\cite{arjovsky2020invariantriskminimization} & $67.50_{\pm0.88}$ & $66.55_{\pm1.25}$ & $68.79_{\pm1.25}$ & 67.61 & $62.30_{\pm0.03}$ & $64.95_{\pm0.03}$ & $71.31_{\pm0.83}$ & 66.19 & $60.03_{\pm1.18}$ & $61.90_{\pm1.74}$ & $61.76_{\pm1.92}$ & 61.23 \\
 &  & Mixup~\cite{yan2020improveunsuperviseddomainadaptation} & $71.08_{\pm0.37}$ & $66.86_{\pm1.01}$ & $69.31_{\pm0.22}$ & 69.08 & $63.13_{\pm1.16}$ & $65.86_{\pm1.23}$ & $73.33_{\pm0.03}$ & 67.44 & $62.10_{\pm1.29}$ & $61.39_{\pm0.59}$ & $63.72_{\pm1.89}$ & 62.40 \\
 &  & CDANN~\cite{Li_2018_ECCV} & $68.33_{\pm0.72}$ & $66.13_{\pm1.14}$ & $68.35_{\pm1.27}$ & 67.60 & $61.46_{\pm1.03}$ & $66.68_{\pm0.41}$ & $69.95_{\pm0.06}$ & 66.03 & $62.53_{\pm1.72}$ & $61.88_{\pm1.40}$ & $63.63_{\pm2.20}$ & 62.68 \\
 &  & SagNet~\cite{nam2021reducingdomaingapreducing} & $71.10_{\pm1.03}$ & $65.79_{\pm0.59}$ & $68.23_{\pm1.79}$ & 68.37 & $62.39_{\pm0.40}$ & $65.69_{\pm0.97}$ & $76.53_{\pm0.76}$ & 68.20 & $61.03_{\pm2.65}$ & $61.48_{\pm1.12}$ & $63.43_{\pm1.79}$ & 61.98 \\
 &  & IB\_ERM~\cite{ahuja2022invarianceprinciplemeetsinformation} & $70.73_{\pm0.89}$ & $66.14_{\pm0.70}$ & $65.60_{\pm0.83}$ & 67.49 & $63.45_{\pm2.22}$ & $64.78_{\pm0.01}$ & $71.34_{\pm1.25}$ & 66.52 & $60.93_{\pm1.52}$ & $62.04_{\pm1.33}$ & $65.83_{\pm1.66}$ & 62.93 \\
 &  & CondCAD~\cite{ruan2022optimalrepresentationscovariateshift} & $67.61_{\pm0.03}$ & $67.08_{\pm1.26}$ & $66.77_{\pm0.80}$ & 67.15 & $61.97_{\pm0.11}$ & $64.37_{\pm0.76}$ & $69.64_{\pm0.65}$ & 65.33 & $60.52_{\pm1.63}$ & $62.84_{\pm2.55}$ & $61.34_{\pm1.85}$ & 61.57 \\
 &  & EQRM~\cite{eastwood2023probabledomaingeneralizationquantile} & $70.91_{\pm0.75}$ & $67.11_{\pm0.58}$ & $69.07_{\pm1.12}$ & 69.03 & $63.55_{\pm1.53}$ & $65.09_{\pm0.38}$ & $75.77_{\pm0.41}$ & 68.14 & $59.77_{\pm1.43}$ & $60.22_{\pm1.48}$ & $62.05_{\pm1.51}$ & 60.68 \\
 &  & ERM++~\cite{teterwak2024ermimprovedbaselinedomain} & $70.29_{\pm2.22}$ & $66.84_{\pm0.66}$ & $67.99_{\pm1.09}$ & 68.37 & $62.88_{\pm1.23}$ & $63.04_{\pm0.68}$ & $70.69_{\pm1.07}$ & 65.54 & $59.21_{\pm1.33}$ & $60.70_{\pm2.79}$ & $63.12_{\pm2.09}$ & 61.01 \\
 &  & URM~\cite{krishnamachari2024uniformly} & $71.71_{\pm0.28}$ & $65.45_{\pm0.59}$ & $69.05_{\pm1.17}$ & 68.74 & $61.76_{\pm0.65}$ & $66.07_{\pm1.15}$ & $74.25_{\pm1.41}$ & 67.36 & $60.16_{\pm1.79}$ & $62.57_{\pm2.31}$ & $63.98_{\pm1.38}$ & 62.24 \\
\midrule
\multirow{13}{*}{LanguageBind~\cite{zhu2023languagebind}} & \multirow{3}{*}{MML} & Concat & $60.72_{\pm0.22}$ & $61.32_{\pm1.20}$ & $62.69_{\pm0.80}$ & 61.58 & $64.07_{\pm0.71}$ & $63.07_{\pm1.17}$ & $62.47_{\pm2.11}$ & 63.20 & $57.72_{\pm1.07}$ & $58.18_{\pm1.36}$ & $57.97_{\pm2.23}$ & 57.96 \\
 &  & OGM~\cite{peng2022balancedmultimodallearningonthefly} & $60.04_{\pm0.69}$ & $60.52_{\pm2.06}$ & $61.20_{\pm0.99}$ & 60.59 & $63.52_{\pm1.31}$ & $63.32_{\pm1.32}$ & $61.22_{\pm1.01}$ & 62.69 & $59.06_{\pm1.75}$ & $56.21_{\pm1.91}$ & $56.42_{\pm1.18}$ & 57.23 \\
 &  & DLMG~\cite{NEURIPS2024_71b17f00} & $61.00_{\pm0.38}$ & $61.64_{\pm0.75}$ & $60.21_{\pm1.07}$ & 60.95 & $63.40_{\pm0.09}$ & $62.47_{\pm1.04}$ & $64.46_{\pm0.86}$ & 63.44 & $56.65_{\pm1.99}$ & $57.47_{\pm1.92}$ & $58.25_{\pm1.89}$ & 57.46 \\
\cmidrule{2-15}
 & \multirow{10}{*}{DG} & ERM& $68.78_{\pm1.10}$ & $62.90_{\pm2.19}$ & $68.24_{\pm1.30}$ & 66.64 & $59.93_{\pm0.71}$ & $67.16_{\pm0.95}$ & $62.95_{\pm0.33}$ & 63.35 & $59.27_{\pm1.60}$ & $59.32_{\pm1.28}$ & $60.37_{\pm1.14}$ & 59.65 \\
 &  & IRM~\cite{arjovsky2020invariantriskminimization} & $67.70_{\pm0.34}$ & $63.43_{\pm0.26}$ & $68.20_{\pm0.63}$ & 66.44 & $61.00_{\pm1.83}$ & $66.15_{\pm0.59}$ & $63.40_{\pm0.23}$ & 63.52 & $58.80_{\pm1.30}$ & $60.35_{\pm1.38}$ & $61.00_{\pm1.72}$ & 60.05 \\
 &  & Mixup~\cite{yan2020improveunsuperviseddomainadaptation} & $67.11_{\pm1.37}$ & $62.96_{\pm0.07}$ & $68.49_{\pm1.37}$ & 66.19 & $60.13_{\pm1.20}$ & $68.80_{\pm0.41}$ & $62.94_{\pm1.18}$ & 63.96 & $60.21_{\pm1.85}$ & $59.76_{\pm1.32}$ & $61.41_{\pm1.03}$ & 60.46 \\
 &  & CDANN~\cite{Li_2018_ECCV} & $67.41_{\pm0.41}$ & $62.96_{\pm0.57}$ & $67.97_{\pm1.31}$ & 66.11 & $61.04_{\pm1.17}$ & $64.18_{\pm0.97}$ & $63.77_{\pm0.24}$ & 63.00 & $62.10_{\pm1.89}$ & $60.39_{\pm1.99}$ & $62.27_{\pm1.61}$ & 61.59 \\
 &  & SagNet~\cite{nam2021reducingdomaingapreducing} & $68.93_{\pm1.04}$ & $61.70_{\pm1.40}$ & $68.31_{\pm0.58}$ & 66.31 & $59.03_{\pm0.47}$ & $65.88_{\pm0.15}$ & $62.80_{\pm0.84}$ & 62.57 & $58.04_{\pm1.29}$ & $60.87_{\pm1.33}$ & $61.23_{\pm1.28}$ & 60.05 \\
 &  & IB\_ERM~\cite{ahuja2022invarianceprinciplemeetsinformation} & $68.57_{\pm2.07}$ & $63.30_{\pm1.07}$ & $67.36_{\pm0.47}$ & 66.41 & $62.13_{\pm2.39}$ & $68.17_{\pm0.55}$ & $65.23_{\pm0.42}$ & 65.18 & $60.00_{\pm2.19}$ & $60.81_{\pm1.07}$ & $62.33_{\pm1.50}$ & 61.05 \\
 &  & CondCAD~\cite{ruan2022optimalrepresentationscovariateshift} & $65.74_{\pm1.51}$ & $62.04_{\pm0.95}$ & $64.84_{\pm0.50}$ & 64.21 & $60.60_{\pm1.04}$ & $65.43_{\pm0.93}$ & $64.19_{\pm1.31}$ & 63.41 & $59.94_{\pm1.92}$ & $60.80_{\pm1.32}$ & $61.10_{\pm1.04}$ & 60.61 \\
 &  & EQRM~\cite{eastwood2023probabledomaingeneralizationquantile} & $69.08_{\pm1.04}$ & $61.37_{\pm1.45}$ & $68.17_{\pm0.07}$ & 66.21 & $59.63_{\pm1.23}$ & $65.27_{\pm1.10}$ & $63.43_{\pm1.75}$ & 62.78 & $59.59_{\pm2.54}$ & $58.61_{\pm1.63}$ & $59.47_{\pm1.73}$ & 59.22 \\
 &  & ERM++~\cite{teterwak2024ermimprovedbaselinedomain} & $67.65_{\pm1.05}$ & $62.64_{\pm1.45}$ & $67.62_{\pm1.40}$ & 65.97 & $61.78_{\pm0.32}$ & $67.96_{\pm2.40}$ & $63.79_{\pm0.31}$ & 64.51 & $58.89_{\pm2.28}$ & $60.32_{\pm1.44}$ & $60.65_{\pm1.65}$ & 59.95 \\
 &  & URM~\cite{krishnamachari2024uniformly} & $69.26_{\pm0.40}$ & $63.39_{\pm1.39}$ & $67.18_{\pm0.43}$ & 66.61 & $59.95_{\pm0.91}$ & $68.30_{\pm0.04}$ & $63.95_{\pm0.24}$ & 64.07 & $59.31_{\pm1.54}$ & $59.85_{\pm1.79}$ & $60.97_{\pm1.66}$ & 60.04 \\
\midrule
\multirow{13}{*}{UniBind~\cite{lyu2024unibind}} & \multirow{3}{*}{MML} & Concat & $60.43_{\pm1.00}$ & $61.76_{\pm0.21}$ & $61.17_{\pm0.38}$ & 61.12 & $64.31_{\pm1.24}$ & $66.02_{\pm0.89}$ & $63.33_{\pm1.49}$ & 64.55 & $57.25_{\pm1.79}$ & $58.54_{\pm1.67}$ & $58.67_{\pm1.44}$ & 58.15 \\
 &  & OGM~\cite{peng2022balancedmultimodallearningonthefly} & $61.22_{\pm0.84}$ & $59.97_{\pm1.27}$ & $59.93_{\pm0.02}$ & 60.37 & $62.95_{\pm0.31}$ & $62.30_{\pm1.30}$ & $64.07_{\pm0.20}$ & 63.11 & $57.93_{\pm1.30}$ & $56.17_{\pm1.24}$ & $56.95_{\pm1.35}$ & 57.02 \\
 &  & DLMG~\cite{NEURIPS2024_71b17f00} & $60.71_{\pm1.56}$ & $61.35_{\pm1.26}$ & $60.79_{\pm1.20}$ & 60.95 & $63.91_{\pm1.17}$ & $62.95_{\pm0.05}$ & $63.80_{\pm1.12}$ & 63.55 & $58.01_{\pm1.83}$ & $57.73_{\pm2.61}$ & $57.25_{\pm1.98}$ & 57.66 \\
\cmidrule{2-15}
 & \multirow{10}{*}{DG} & ERM & $71.86_{\pm0.63}$ & $66.07_{\pm0.84}$ & $69.03_{\pm0.17}$ & 68.99 & $61.27_{\pm1.28}$ & $62.52_{\pm1.32}$ & $77.07_{\pm1.33}$ & 66.95 & $58.98_{\pm1.40}$ & $61.32_{\pm1.29}$ & $63.57_{\pm1.48}$ & 61.29 \\
 &  & IRM~\cite{arjovsky2020invariantriskminimization} & $68.50_{\pm2.28}$ & $66.85_{\pm0.23}$ & $70.20_{\pm0.94}$ & 68.52 & $62.98_{\pm1.44}$ & $63.57_{\pm1.32}$ & $69.80_{\pm1.29}$ & 65.45 & $60.97_{\pm1.83}$ & $60.93_{\pm1.59}$ & $63.20_{\pm1.73}$ & 61.70 \\
 &  & Mixup~\cite{yan2020improveunsuperviseddomainadaptation} & $69.51_{\pm0.57}$ & $67.28_{\pm0.67}$ & $69.34_{\pm0.84}$ & 68.71 & $62.50_{\pm0.36}$ & $63.40_{\pm0.75}$ & $76.11_{\pm0.90}$ & 67.34 & $61.46_{\pm2.12}$ & $62.26_{\pm1.63}$ & $64.51_{\pm1.86}$ & 62.74 \\
 &  & CDANN~\cite{Li_2018_ECCV} & $70.49_{\pm1.00}$ & $66.41_{\pm0.41}$ & $69.25_{\pm1.47}$ & 68.72 & $61.50_{\pm1.43}$ & $65.48_{\pm0.28}$ & $73.43_{\pm0.48}$ & 66.80 & $63.02_{\pm2.99}$ & $61.79_{\pm1.42}$ & $62.63_{\pm1.68}$ & 62.48 \\
 &  & SagNet~\cite{nam2021reducingdomaingapreducing} & $71.53_{\pm0.73}$ & $67.39_{\pm0.21}$ & $69.23_{\pm1.06}$ & 69.38 & $61.31_{\pm0.60}$ & $64.77_{\pm1.22}$ & $77.59_{\pm1.00}$ & 67.89 & $62.28_{\pm1.15}$ & $60.86_{\pm1.34}$ & $62.43_{\pm1.86}$ & 61.86 \\
 &  & IB\_ERM~\cite{ahuja2022invarianceprinciplemeetsinformation} & $70.19_{\pm0.30}$ & $67.53_{\pm1.39}$ & $67.91_{\pm0.87}$ & 68.54 & $63.76_{\pm0.18}$ & $64.10_{\pm0.85}$ & $71.03_{\pm2.16}$ & 66.30 & $60.53_{\pm0.95}$ & $63.38_{\pm1.39}$ & $65.43_{\pm1.86}$ & 63.11 \\
 &  & CondCAD~\cite{ruan2022optimalrepresentationscovariateshift} & $67.77_{\pm0.08}$ & $66.40_{\pm1.23}$ & $66.61_{\pm0.63}$ & 66.93 & $61.39_{\pm0.41}$ & $63.99_{\pm1.43}$ & $69.94_{\pm0.93}$ & 65.11 & $61.42_{\pm1.13}$ & $60.80_{\pm1.82}$ & $60.96_{\pm1.87}$ & 61.06 \\
 &  & EQRM~\cite{eastwood2023probabledomaingeneralizationquantile} & $71.68_{\pm0.56}$ & $66.26_{\pm0.66}$ & $68.48_{\pm0.18}$ & 68.81 & $61.78_{\pm0.23}$ & $65.21_{\pm1.49}$ & $78.37_{\pm1.29}$ & 68.45 & $60.79_{\pm1.52}$ & $59.81_{\pm1.82}$ & $61.47_{\pm2.78}$ & 60.69 \\
 &  & ERM++~\cite{teterwak2024ermimprovedbaselinedomain} & $70.99_{\pm0.11}$ & $66.47_{\pm0.14}$ & $69.59_{\pm0.12}$ & 69.02 & $62.27_{\pm1.82}$ & $63.24_{\pm1.21}$ & $69.92_{\pm0.63}$ & 65.14 & $60.19_{\pm2.70}$ & $60.64_{\pm2.00}$ & $62.69_{\pm1.90}$ & 61.17 \\
 &  & URM~\cite{krishnamachari2024uniformly} & $64.47_{\pm0.22}$ & $65.15_{\pm0.51}$ & $68.68_{\pm0.75}$ & 66.10 & $60.68_{\pm1.45}$ & $63.92_{\pm0.20}$ & $75.48_{\pm0.71}$ & 66.69 & $60.88_{\pm1.12}$ & $62.66_{\pm1.02}$ & $63.09_{\pm1.96}$ & 62.21 \\

\bottomrule
\end{tabular}
}
\end{table*}

\section{Detailed Experimental Results}
\label{detaileddata}
This section presents the experimental evaluation results under various modality combinations. We systematically tested 3 multi-modal learning (MML) methods and 10 domain generalization (DG) algorithms, supported by three different modality-binding perceptors. All evaluations strictly follow the three model selection protocols defined in Appendix~\ref{algrithim} and are conducted on diverse datasets, including LAV-DF~\cite{cai2023reallymeanthatcontent}, FakeAVCeleb~\cite{khalid2022fakeavcelebnovelaudiovideomultimodal}, and Celeb+Asv~\cite{wang2024asvspoof5crowdsourcedspeech, li2025celeb}, to ensure a rigorous and comprehensive benchmark.
The experimental data is organized into six core tables to systematically present our findings: Table~\ref{tab:weak_mg_test_modality_oracle}, Table~\ref{tab:weak_leave_one_out_custom_random} and Table~\ref{tab:weak_training_val_random_dist}, report performance under the Weak MAF setting, while Table~\ref{tab:strong_oracle_varied_final}, Table~\ref{tab:strong_leave_one_out_random} and Table~\ref{tab:strong_training_val_random} provide an in-depth comparison under the Strong MAF setting. These quantitative results remain highly consistent with the visualization trends observed in Figures ~\ref{fig:weakmaf}, ~\ref{fig:weak_radar_charts}, ~\ref{fig:weak_model_selection}, ~\ref{fig:strong_mg_loo1}, ~\ref{fig:strong_model_selection}, ~\ref{fig:strong_radar_charts}. Overall, the experimental results demonstrate that the MAF framework has significant advantages in capturing shared latent forgery knowledge. Especially when dealing with distribution shifts and unknown "dark modalities", MAF shows better robustness and generalization potential than traditional fusion methods.

\subsection{Results of Weak MAF}
Under the Weak MAF setting, the results in Table ~\ref{tab:weak_mg_test_modality_oracle} exhibit clear performance trends in test-modality validation set. Regarding perceptor comparisons, LanguageBind~\cite{zhu2023languagebind} generally underperforms compared to ImageBind~\cite{girdhar2023imagebind} and UniBind~\cite{lyu2024unibind}, reflecting the superiority of the latter two in representing forgery features. From the dataset perspective, the well-aligned LAV-DF~\cite{cai2023reallymeanthatcontent} and FakeAVCeleb\cite{khalid2022fakeavcelebnovelaudiovideomultimodal} datasets show higher detection accuracy than the unaligned Celeb+Asv~\cite{wang2024asvspoof5crowdsourcedspeech,li2025celeb}, validating that cross-modal correlation enhances forensic effectiveness. Furthermore, the Oracle model selection protocol demonstrates lower variance and high numerical stability, effectively filtering out training fluctuations to identify the performance upper bound. Overall, the framework maintains robust discriminative power across various combinations of perceptors and datasets, proving that its decoupling strategy successfully strips away style noise while preserving universal forgery imprints.

Table ~\ref{tab:weak_leave_one_out_custom_random} presents the generalization performance of models using the leave-one-modality-out cross-validation protocol. In perceptor comparisons, ImageBind~\cite{girdhar2023imagebind} and UniBind~\cite{lyu2024unibind} consistently outperform LanguageBind~\cite{zhu2023languagebind} due to their more robust feature representation capabilities. This provides a more reliable foundation for simulating unknown modality environments.  From the dataset perspective, the average accuracy of the well-aligned LAV-DF~\cite{cai2023reallymeanthatcontent} and FakeAVCeleb~\cite{khalid2022fakeavcelebnovelaudiovideomultimodal} datasets is significantly higher than that of the unaligned Celeb+Asv~\cite{li2025celeb,wang2024asvspoof5crowdsourcedspeech} when handling modality removal tasks. This further validates that cross-modal semantic correlation enhances the capture of ``meta-distribution'' features. Compared to the Oracle protocol, the numerical variance under LOO is slightly higher and the performance is more conservative, reflecting the inherent challenges of model selection without target modality priors. Nevertheless, the DG algorithms integrated into the MAF framework maintain stable discriminative power even under these strict constraints. This demonstrates the framework's empirical potential to transfer universal forgery knowledge from known domains to unknown ``dark modalities''.

Table ~\ref{tab:weak_training_val_random_dist} presents the classification performance trends using the training-modality validation set protocol. In perceptor comparisons, ImageBind~\cite{girdhar2023imagebind} and UniBind~\cite{lyu2024unibind} outperform LanguageBind~\cite{zhu2023languagebind} due to their stronger lower-level representation capabilities, demonstrating higher consistency in feature extraction. From the dataset perspective, the semantically aligned LAV-DF~\cite{cai2023reallymeanthatcontent} and FakeAVCeleb~\cite{khalid2022fakeavcelebnovelaudiovideomultimodal} datasets achieve significantly higher accuracy than the unaligned Celeb+Asv~\cite{wang2024asvspoof5crowdsourcedspeech,li2025celeb}, confirming that cross-modal correlation enhances the capture of generative biases. Since the training and validation distributions are consistent under this protocol, the results exhibit smaller numerical variance and more stable performance, effectively evaluating the model's basic fitting capability within the known distribution. Overall, this table demonstrates that the DG algorithms integrated into the MAF framework possess strong discriminative power under standard validation workflows, robustly stripping away modality style noise to lock onto core forensic features.

\subsection{Results of Strong MAF}
Under the Strong MAF setting, Table ~\ref{tab:strong_oracle_varied_final} exhibits a trend consistent with Weak MAF: ImageBind~\cite{girdhar2023imagebind} and UniBind~\cite{lyu2024unibind} outperform LanguageBind~\cite{zhu2023languagebind}. Similarly, the semantically aligned LAV-DF and FakeAVCeleb~\cite{khalid2022fakeavcelebnovelaudiovideomultimodal} datasets significantly outperform the unaligned Celeb+Asv\cite{li2025celeb,wang2024asvspoof5crowdsourcedspeech}. Due to more thorough perceptor isolation (e.g., incorporating LoRA fine-tuning and excluding it from the random search), the performance upper bounds of all algorithms have narrowed compared to Weak MAF. However, under the guidance of the Oracle protocol, model selection still demonstrates very low variance and high stability. Overall, this table proves that the MAF framework can effectively strip away style noise through its decoupling strategy and consistently lock onto discriminative forensic features under extreme generalization challenges.

Table ~\ref{tab:strong_leave_one_out_random} utilizes the leave-one-modality-out protocol  and reveals patterns consistent with other experimental results. In perceptor comparisons, ImageBind~\cite{girdhar2023imagebind} and UniBind~\cite{lyu2024unibind} outperform LanguageBind~\cite{zhu2023languagebind}, demonstrating their superior representation capabilities. From the dataset perspective, the semantically aligned LAV-DF\cite{cai2023reallymeanthatcontent} and FakeAVCeleb~\cite{khalid2022fakeavcelebnovelaudiovideomultimodal} exhibit significantly higher performance than the unaligned Celeb+Asv\cite{li2025celeb,wang2024asvspoof5crowdsourcedspeech}. Compared to the Weak MAF setting, the performance upper bounds for all algorithms have narrowed. However, despite the challenge of simulating unknown modality environments, the LOO protocol still maintains good numerical stability for model selection.

Table ~\ref{tab:strong_training_val_random} utilizes the training-modality protocol and reveals performance trends consistent with the Oracle and LOO protocols. In perceptor comparisons, ImageBind~\cite{girdhar2023imagebind} and UniBind~\cite{lyu2024unibind} outperform LanguageBind~\cite{zhu2023languagebind}. From the dataset perspective, the semantically aligned LAV-DF~\cite{cai2023reallymeanthatcontent} and FakeAVCeleb~\cite{khalid2022fakeavcelebnovelaudiovideomultimodal} datasets significantly outperform the unaligned Celeb+Asv\cite{li2025celeb,wang2024asvspoof5crowdsourcedspeech}. Compared to the Weak MAF setting, the performance upper bounds for all algorithms have narrowed. Compared to the other two protocols, the TM protocol exhibits the lowest numerical variance and highest stability because the training and validation sets share the same distribution. Overall, this table demonstrates that under a standard training workflow, the MAF framework effectively strips away style noise through its decoupling strategy and consistently locks onto discriminative forensic features.

\begin{table*}[t]
\centering
\caption{Classification performance (mean $\pm$ std) under the Weak MAF setting, using training-modality validation set for model selection.}
\label{tab:weak_training_val_random_dist}
\resizebox{\textwidth}{!}{ 
\begin{tabular}{l | ll | cccc | cccc | cccc}
\toprule
\multirow{2}{*}{\textbf{Perceptor}} & \multicolumn{2}{l|}{\multirow{2}{*}{\textbf{Method}}} & \multicolumn{4}{c|}{\textbf{LAV-DF~\cite{cai2023reallymeanthatcontent}}} & \multicolumn{4}{c|}{\textbf{Fakeavcele~\cite{khalid2022fakeavcelebnovelaudiovideomultimodal}}} & \multicolumn{4}{c}{\textbf{Cele+Asv~\cite{wang2024asvspoof5crowdsourcedspeech, li2025celeb}}} \\
\cmidrule(lr){4-7} \cmidrule(lr){8-11} \cmidrule(lr){12-15}
& & & \textbf{Vid} & \textbf{Aud} & \textbf{Img} & \textbf{Avg} & \textbf{Vid} & \textbf{Aud} & \textbf{Img} & \textbf{Avg} & \textbf{Vid} & \textbf{Aud} & \textbf{Img} & \textbf{Avg} \\
\midrule
\midrule

\multirow{13}{*}{ImageBind~\cite{girdhar2023imagebind}} & \multirow{3}{*}{MML} & Concat & $64.21_{\pm0.78}$ & $62.66_{\pm0.72}$ & $63.51_{\pm0.73}$ & 63.46 & $64.50_{\pm0.99}$ & $66.57_{\pm0.35}$ & $65.60_{\pm0.47}$ & 65.56 & $58.71_{\pm0.03}$ & $58.96_{\pm1.39}$ & $58.86_{\pm0.79}$ & 58.84 \\
 &  & OGM~\cite{peng2022balancedmultimodallearningonthefly} & $59.88_{\pm1.16}$ & $61.23_{\pm0.72}$ & $62.02_{\pm0.24}$ & 61.04 & $64.00_{\pm0.14}$ & $61.59_{\pm1.47}$ & $63.70_{\pm0.04}$ & 63.10 & $58.13_{\pm0.95}$ & $56.91_{\pm0.81}$ & $56.32_{\pm0.41}$ & 57.12 \\
 &  & DLMG~\cite{NEURIPS2024_71b17f00} & $60.03_{\pm0.73}$ & $61.29_{\pm0.20}$ & $61.23_{\pm0.39}$ & 60.85 & $63.85_{\pm0.40}$ & $62.88_{\pm0.38}$ & $63.44_{\pm0.37}$ & 63.39 & $58.07_{\pm0.66}$ & $58.09_{\pm0.68}$ & $57.65_{\pm0.54}$ & 57.94 \\
\cmidrule{2-15}
 & \multirow{10}{*}{DG} & ERM~\cite{vapnik1991principles} & $71.35_{\pm0.66}$ & $65.85_{\pm1.44}$ & $67.81_{\pm0.30}$ & 68.34 & $62.83_{\pm0.02}$ & $61.08_{\pm0.84}$ & $76.01_{\pm0.50}$ & 66.64 & $59.91_{\pm0.65}$ & $61.96_{\pm0.73}$ & $63.71_{\pm1.39}$ & 61.86 \\
 &  & IRM~\cite{arjovsky2020invariantriskminimization} & $68.64_{\pm0.15}$ & $66.89_{\pm0.90}$ & $70.84_{\pm0.45}$ & 68.79 & $61.72_{\pm0.35}$ & $64.63_{\pm0.58}$ & $72.44_{\pm0.65}$ & 66.26 & $60.49_{\pm0.50}$ & $61.60_{\pm0.06}$ & $63.04_{\pm0.92}$ & 61.71 \\
 &  & Mixup~\cite{yan2020improveunsuperviseddomainadaptation} & $71.15_{\pm1.04}$ & $66.94_{\pm0.42}$ & $69.11_{\pm0.06}$ & 69.07 & $62.65_{\pm1.00}$ & $63.94_{\pm0.16}$ & $71.15_{\pm0.43}$ & 65.91 & $61.95_{\pm0.75}$ & $62.90_{\pm0.75}$ & $65.65_{\pm0.26}$ & 63.50 \\
 &  & CDANN~\cite{Li_2018_ECCV} & $70.41_{\pm0.85}$ & $67.70_{\pm0.19}$ & $69.48_{\pm0.63}$ & 69.20 & $61.26_{\pm0.87}$ & $65.92_{\pm0.65}$ & $76.62_{\pm0.54}$ & 67.93 & $62.08_{\pm1.00}$ & $61.52_{\pm0.86}$ & $63.67_{\pm0.21}$ & 62.42 \\
 &  & SagNet~\cite{nam2021reducingdomaingapreducing} & $70.15_{\pm0.78}$ & $65.50_{\pm0.34}$ & $68.08_{\pm0.33}$ & 67.91 & $62.75_{\pm1.05}$ & $64.07_{\pm0.97}$ & $69.87_{\pm0.29}$ & 65.56 & $61.37_{\pm0.37}$ & $64.22_{\pm0.89}$ & $66.57_{\pm0.77}$ & 64.05 \\
 &  & IB\_ERM~\cite{ahuja2022invarianceprinciplemeetsinformation} & $70.58_{\pm0.04}$ & $67.09_{\pm0.54}$ & $69.46_{\pm0.88}$ & 69.04 & $61.82_{\pm0.72}$ & $64.17_{\pm0.51}$ & $69.39_{\pm0.66}$ & 65.13 & $61.94_{\pm0.66}$ & $61.93_{\pm0.22}$ & $64.67_{\pm0.70}$ & 62.85 \\
 &  & CondCAD~\cite{ruan2022optimalrepresentationscovariateshift} & $70.61_{\pm0.85}$ & $65.67_{\pm0.74}$ & $69.02_{\pm0.43}$ & 68.43 & $61.48_{\pm0.45}$ & $62.49_{\pm0.58}$ & $74.81_{\pm0.89}$ & 66.26 & $63.07_{\pm0.27}$ & $62.15_{\pm0.95}$ & $62.16_{\pm0.65}$ & 62.46 \\
 &  & EQRM~\cite{eastwood2023probabledomaingeneralizationquantile} & $70.56_{\pm0.77}$ & $67.57_{\pm0.18}$ & $69.11_{\pm0.03}$ & 69.08 & $61.13_{\pm0.07}$ & $68.14_{\pm0.77}$ & $75.98_{\pm0.16}$ & 68.42 & $62.34_{\pm0.68}$ & $60.96_{\pm1.00}$ & $63.08_{\pm0.28}$ & 62.13 \\
 &  & ERM++~\cite{teterwak2024ermimprovedbaselinedomain} & $71.67_{\pm0.87}$ & $67.18_{\pm1.09}$ & $69.64_{\pm1.27}$ & 69.50 & $62.73_{\pm0.32}$ & $63.68_{\pm0.94}$ & $77.23_{\pm0.96}$ & 67.88 & $60.93_{\pm0.43}$ & $60.44_{\pm0.84}$ & $62.65_{\pm0.20}$ & 61.34 \\
 &  & URM~\cite{krishnamachari2024uniformly} & $71.09_{\pm0.18}$ & $66.37_{\pm0.23}$ & $68.62_{\pm1.13}$ & 68.69 & $62.02_{\pm1.14}$ & $63.07_{\pm0.49}$ & $74.41_{\pm0.82}$ & 66.50 & $61.52_{\pm0.28}$ & $62.67_{\pm0.74}$ & $65.32_{\pm1.23}$ & 63.17 \\
\midrule
\multirow{13}{*}{LanguageBind~\cite{zhu2023languagebind}} & \multirow{3}{*}{MML} & Concat & $63.82_{\pm0.28}$ & $63.39_{\pm0.48}$ & $62.80_{\pm0.94}$ & 63.34 & $62.91_{\pm0.97}$ & $68.06_{\pm0.04}$ & $70.09_{\pm0.07}$ & 67.02 & $57.91_{\pm0.52}$ & $58.98_{\pm0.37}$ & $58.81_{\pm0.91}$ & 58.57 \\
 &  & OGM~\cite{peng2022balancedmultimodallearningonthefly} & $59.87_{\pm0.60}$ & $60.35_{\pm0.10}$ & $61.45_{\pm0.37}$ & 60.56 & $63.25_{\pm0.45}$ & $62.62_{\pm0.33}$ & $63.34_{\pm0.60}$ & 63.07 & $57.60_{\pm0.78}$ & $57.46_{\pm0.05}$ & $57.12_{\pm0.88}$ & 57.39 \\
 &  & DLMG~\cite{NEURIPS2024_71b17f00} & $60.68_{\pm0.93}$ & $61.29_{\pm0.97}$ & $61.40_{\pm0.63}$ & 61.12 & $64.34_{\pm0.77}$ & $64.08_{\pm0.97}$ & $63.44_{\pm0.82}$ & 63.95 & $57.01_{\pm0.41}$ & $57.76_{\pm0.44}$ & $58.84_{\pm0.35}$ & 57.87 \\
\cmidrule{2-15}
 & \multirow{10}{*}{DG} & ERM~\cite{vapnik1991principles}& $69.21_{\pm0.81}$ & $61.49_{\pm1.34}$ & $68.07_{\pm0.25}$ & 66.26 & $59.14_{\pm0.53}$ & $63.06_{\pm0.14}$ & $71.01_{\pm0.28}$ & 64.40 & $59.55_{\pm0.17}$ & $61.42_{\pm0.75}$ & $61.51_{\pm0.96}$ & 60.83 \\
 &  & IRM~\cite{arjovsky2020invariantriskminimization} & $67.14_{\pm0.50}$ & $63.99_{\pm0.13}$ & $68.84_{\pm0.87}$ & 66.66 & $59.94_{\pm0.57}$ & $65.29_{\pm1.19}$ & $70.64_{\pm1.27}$ & 65.29 & $60.12_{\pm0.95}$ & $59.11_{\pm0.10}$ & $60.64_{\pm0.70}$ & 59.96 \\
 &  & Mixup~\cite{yan2020improveunsuperviseddomainadaptation} & $69.53_{\pm0.65}$ & $61.39_{\pm0.48}$ & $68.16_{\pm0.31}$ & 66.36 & $60.62_{\pm0.16}$ & $66.50_{\pm0.03}$ & $78.75_{\pm0.87}$ & 68.62 & $61.05_{\pm0.03}$ & $60.30_{\pm0.56}$ & $62.45_{\pm0.93}$ & 61.27 \\
 &  & CDANN~\cite{Li_2018_ECCV} & $66.83_{\pm0.67}$ & $63.29_{\pm0.41}$ & $68.10_{\pm1.44}$ & 66.07 & $58.17_{\pm0.62}$ & $65.10_{\pm0.71}$ & $75.17_{\pm0.69}$ & 66.15 & $59.92_{\pm0.01}$ & $60.81_{\pm0.19}$ & $61.12_{\pm0.72}$ & 60.62 \\
 &  & SagNet~\cite{nam2021reducingdomaingapreducing} & $69.27_{\pm0.54}$ & $61.77_{\pm0.98}$ & $66.65_{\pm0.02}$ & 65.90 & $61.46_{\pm0.05}$ & $64.52_{\pm1.38}$ & $76.27_{\pm0.39}$ & 67.42 & $61.98_{\pm0.80}$ & $60.75_{\pm0.54}$ & $62.39_{\pm0.16}$ & 61.71 \\
 &  & IB\_ERM~\cite{ahuja2022invarianceprinciplemeetsinformation} & $69.79_{\pm0.75}$ & $63.34_{\pm0.25}$ & $68.80_{\pm0.99}$ & 67.31 & $61.18_{\pm0.82}$ & $68.22_{\pm0.55}$ & $77.37_{\pm0.28}$ & 68.92 & $60.62_{\pm0.33}$ & $61.78_{\pm0.83}$ & $62.47_{\pm0.43}$ & 61.62 \\
 &  & CondCAD~\cite{ruan2022optimalrepresentationscovariateshift} & $67.99_{\pm0.39}$ & $62.18_{\pm0.92}$ & $66.48_{\pm0.20}$ & 65.55 & $59.99_{\pm0.57}$ & $64.93_{\pm0.07}$ & $74.48_{\pm1.48}$ & 66.47 & $61.52_{\pm0.24}$ & $62.06_{\pm0.39}$ & $61.25_{\pm0.33}$ & 61.61 \\
 &  & EQRM~\cite{eastwood2023probabledomaingeneralizationquantile} & $68.51_{\pm0.51}$ & $62.82_{\pm0.08}$ & $68.11_{\pm0.66}$ & 66.48 & $60.55_{\pm0.28}$ & $67.46_{\pm0.22}$ & $73.21_{\pm0.20}$ & 67.07 & $59.68_{\pm0.94}$ & $59.96_{\pm0.46}$ & $60.69_{\pm0.47}$ & 60.11 \\
 &  & ERM++~\cite{teterwak2024ermimprovedbaselinedomain} & $67.81_{\pm0.34}$ & $62.93_{\pm0.06}$ & $68.24_{\pm0.87}$ & 66.33 & $59.56_{\pm0.05}$ & $65.68_{\pm0.81}$ & $72.73_{\pm0.18}$ & 65.99 & $59.37_{\pm0.48}$ & $60.90_{\pm0.63}$ & $61.73_{\pm0.31}$ & 60.67 \\
 &  & URM~\cite{krishnamachari2024uniformly} & $68.72_{\pm0.27}$ & $61.88_{\pm0.94}$ & $67.22_{\pm0.02}$ & 65.94 & $59.82_{\pm0.91}$ & $66.47_{\pm0.05}$ & $78.22_{\pm0.77}$ & 68.17 & $59.96_{\pm0.50}$ & $61.20_{\pm0.06}$ & $61.02_{\pm0.73}$ & 60.73 \\
\midrule
\multirow{13}{*}{UniBind~\cite{lyu2024unibind}} & \multirow{3}{*}{MML} & Concat & $64.01_{\pm0.92}$ & $63.66_{\pm0.99}$ & $63.11_{\pm0.37}$ & 63.59 & $63.41_{\pm0.93}$ & $66.86_{\pm0.25}$ & $69.21_{\pm0.95}$ & 66.49 & $58.51_{\pm0.40}$ & $58.10_{\pm1.43}$ & $58.61_{\pm0.69}$ & 58.41 \\
 &  & OGM~\cite{peng2022balancedmultimodallearningonthefly} & $61.18_{\pm0.69}$ & $60.63_{\pm1.03}$ & $60.14_{\pm1.46}$ & 60.65 & $63.49_{\pm0.25}$ & $62.38_{\pm0.05}$ & $63.58_{\pm0.88}$ & 63.15 & $56.81_{\pm0.46}$ & $57.54_{\pm0.48}$ & $58.07_{\pm0.39}$ & 57.47 \\
 &  & DLMG~\cite{NEURIPS2024_71b17f00} & $60.80_{\pm1.32}$ & $61.39_{\pm0.84}$ & $61.53_{\pm0.53}$ & 61.24 & $64.02_{\pm0.09}$ & $61.97_{\pm0.77}$ & $63.44_{\pm0.20}$ & 63.14 & $57.94_{\pm0.66}$ & $58.76_{\pm0.79}$ & $57.97_{\pm0.44}$ & 58.22 \\
\cmidrule{2-15}
 & \multirow{10}{*}{DG} & ERM~\cite{vapnik1991principles} & $70.49_{\pm0.02}$ & $65.78_{\pm0.49}$ & $67.18_{\pm0.38}$ & 67.82 & $62.44_{\pm0.40}$ & $62.48_{\pm0.81}$ & $72.91_{\pm0.18}$ & 65.94 & $61.42_{\pm0.55}$ & $62.36_{\pm0.05}$ & $64.89_{\pm0.03}$ & 62.89 \\
 &  & IRM~\cite{arjovsky2020invariantriskminimization} & $68.34_{\pm0.51}$ & $67.19_{\pm1.47}$ & $70.64_{\pm0.06}$ & 68.72 & $61.18_{\pm0.54}$ & $61.83_{\pm0.83}$ & $74.55_{\pm0.95}$ & 65.85 & $59.70_{\pm0.35}$ & $62.00_{\pm0.70}$ & $62.84_{\pm0.03}$ & 61.51 \\
 &  & Mixup~\cite{yan2020improveunsuperviseddomainadaptation} & $71.11_{\pm0.82}$ & $66.78_{\pm0.85}$ & $67.76_{\pm0.42}$ & 68.55 & $61.17_{\pm0.36}$ & $62.37_{\pm1.45}$ & $73.95_{\pm0.96}$ & 65.83 & $61.57_{\pm0.21}$ & $63.78_{\pm0.63}$ & $65.95_{\pm0.54}$ & 63.77 \\
 &  & CDANN~\cite{Li_2018_ECCV} & $71.38_{\pm1.48}$ & $66.00_{\pm0.85}$ & $68.45_{\pm1.13}$ & 68.61 & $61.42_{\pm0.23}$ & $64.62_{\pm1.18}$ & $76.61_{\pm0.33}$ & 67.55 & $62.92_{\pm1.42}$ & $61.30_{\pm0.27}$ & $62.27_{\pm0.40}$ & 62.16 \\
 &  & SagNet~\cite{nam2021reducingdomaingapreducing} & $71.27_{\pm0.15}$ & $65.69_{\pm1.01}$ & $68.72_{\pm0.26}$ & 68.56 & $60.87_{\pm0.13}$ & $63.65_{\pm0.48}$ & $72.10_{\pm0.08}$ & 65.54 & $61.77_{\pm0.43}$ & $64.22_{\pm0.10}$ & $66.90_{\pm1.22}$ & 64.30 \\
 &  & IB\_ERM~\cite{ahuja2022invarianceprinciplemeetsinformation} & $70.77_{\pm0.05}$ & $66.64_{\pm0.16}$ & $68.37_{\pm0.17}$ & 68.59 & $62.85_{\pm0.03}$ & $61.54_{\pm0.39}$ & $72.71_{\pm0.25}$ & 65.70 & $62.78_{\pm0.97}$ & $62.32_{\pm0.63}$ & $65.27_{\pm0.19}$ & 63.46 \\
 &  & CondCAD~\cite{ruan2022optimalrepresentationscovariateshift} & $69.84_{\pm0.12}$ & $67.21_{\pm0.07}$ & $67.35_{\pm1.50}$ & 68.13 & $60.28_{\pm0.78}$ & $61.73_{\pm0.45}$ & $75.27_{\pm0.22}$ & 65.76 & $62.66_{\pm0.88}$ & $62.92_{\pm0.58}$ & $63.48_{\pm0.36}$ & 63.02 \\
 &  & EQRM~\cite{eastwood2023probabledomaingeneralizationquantile} & $72.14_{\pm0.17}$ & $66.19_{\pm0.01}$ & $69.32_{\pm0.71}$ & 69.22 & $61.28_{\pm0.29}$ & $63.66_{\pm0.02}$ & $76.21_{\pm0.22}$ & 67.05 & $61.29_{\pm0.03}$ & $60.68_{\pm0.74}$ & $63.61_{\pm1.29}$ & 61.86 \\
 &  & ERM++~\cite{teterwak2024ermimprovedbaselinedomain} & $71.35_{\pm0.94}$ & $67.09_{\pm0.33}$ & $69.31_{\pm0.56}$ & 69.25 & $62.57_{\pm0.10}$ & $63.14_{\pm0.11}$ & $76.48_{\pm0.51}$ & 67.40 & $60.73_{\pm0.94}$ & $61.55_{\pm0.21}$ & $63.23_{\pm1.01}$ & 61.84 \\
 &  & URM~\cite{krishnamachari2024uniformly} & $71.31_{\pm0.12}$ & $66.17_{\pm0.84}$ & $68.62_{\pm0.23}$ & 68.70 & $61.64_{\pm0.27}$ & $62.41_{\pm0.06}$ & $71.67_{\pm0.89}$ & 65.24 & $62.10_{\pm1.48}$ & $63.03_{\pm0.32}$ & $64.62_{\pm1.11}$ & 63.25 \\
\midrule
\bottomrule
\end{tabular}
}
\end{table*}

\begin{table*}[t]
\centering
\caption{Classification performance (mean $\pm$ std) under the Strong MAF setting, using test-modality validation set (oracle) set for model selection.}
\label{tab:strong_oracle_varied_final}
\resizebox{\textwidth}{!}{ 
\begin{tabular}{l|ll | cccc | cccc | cccc}
\toprule
\multirow{2}{*}{\textbf{Perceptor}} & \multicolumn{2}{l|}{\multirow{2}{*}{\textbf{Method}}} & \multicolumn{4}{c|}{\textbf{LAV-DF~\cite{cai2023reallymeanthatcontent}}} & \multicolumn{4}{c|}{\textbf{Fakeavcele~\cite{khalid2022fakeavcelebnovelaudiovideomultimodal}}} & \multicolumn{4}{c}{\textbf{Cele+Asv~\cite{wang2024asvspoof5crowdsourcedspeech, li2025celeb}}} \\
\cmidrule(lr){4-7} \cmidrule(lr){8-11} \cmidrule(lr){12-15}
& & & \textbf{Vid} & \textbf{Aud} & \textbf{Img} & \textbf{Avg} & \textbf{Vid} & \textbf{Aud} & \textbf{Img} & \textbf{Avg} & \textbf{Vid} & \textbf{Aud} & \textbf{Img} & \textbf{Avg} \\
\midrule
\midrule

\multirow{13}{*}{ImageBind~\cite{girdhar2023imagebind}} & \multirow{3}{*}{MML} & Concat & $61.76_{\pm0.37}$ & $61.24_{\pm1.48}$ & $62.32_{\pm0.29}$ & 61.77 & $64.27_{\pm0.63}$ & $65.24_{\pm1.14}$ & $64.47_{\pm0.85}$ & 64.66 & $59.27_{\pm1.73}$ & $59.49_{\pm0.42}$ & $59.47_{\pm1.26}$ & 59.41 \\
 &  & OGM~\cite{peng2022balancedmultimodallearningonthefly} & $60.42_{\pm0.51}$ & $60.72_{\pm0.14}$ & $60.37_{\pm0.76}$ & 60.50 & $62.97_{\pm0.92}$ & $62.81_{\pm1.68}$ & $63.47_{\pm0.39}$ & 63.08 & $58.57_{\pm1.05}$ & $58.22_{\pm0.83}$ & $58.18_{\pm1.44}$ & 58.32 \\
 &  & DLMG~\cite{NEURIPS2024_71b17f00} & $60.35_{\pm1.12}$ & $60.71_{\pm0.85}$ & $58.77_{\pm1.34}$ & 59.94 & $63.30_{\pm0.48}$ & $64.56_{\pm0.27}$ & $62.65_{\pm1.51}$ & 63.50 & $59.32_{\pm0.73}$ & $58.84_{\pm1.29}$ & $59.17_{\pm1.96}$ & 59.11 \\
\cmidrule{2-15}
 & \multirow{10}{*}{DG} & ERM & $59.52_{\pm0.19}$ & $59.14_{\pm1.62}$ & $59.11_{\pm0.33}$ & 59.26 & $61.79_{\pm1.47}$ & $64.03_{\pm0.81}$ & $59.07_{\pm1.24}$ & 61.63 & $57.83_{\pm0.68}$ & $57.10_{\pm1.85}$ & $58.31_{\pm0.96}$ & 57.75 \\
 &  & IRM~\cite{arjovsky2020invariantriskminimization} & $61.46_{\pm0.84}$ & $60.30_{\pm0.45}$ & $61.16_{\pm1.71}$ & 60.97 & $61.23_{\pm0.26}$ & $61.66_{\pm1.37}$ & $65.15_{\pm0.52}$ & 62.68 & $58.84_{\pm1.18}$ & $59.00_{\pm0.91}$ & $58.74_{\pm0.13}$ & 58.86 \\
 &  & Mixup~\cite{yan2020improveunsuperviseddomainadaptation} & $58.99_{\pm1.22}$ & $59.59_{\pm0.73}$ & $58.89_{\pm1.55}$ & 59.16 & $60.43_{\pm0.39}$ & $61.70_{\pm1.83}$ & $62.56_{\pm0.64}$ & 61.56 & $58.18_{\pm0.48}$ & $58.68_{\pm1.09}$ & $56.85_{\pm1.75}$ & 57.90 \\
 &  & CDANN~\cite{Li_2018_ECCV} & $60.63_{\pm0.56}$ & $60.85_{\pm1.14}$ & $58.66_{\pm0.92}$ & 60.05 & $61.07_{\pm1.62}$ & $61.85_{\pm0.15}$ & $61.43_{\pm1.43}$ & 61.45 & $58.74_{\pm1.29}$ & $61.51_{\pm0.37}$ & $59.88_{\pm0.82}$ & 60.04 \\
 &  & SagNet~\cite{nam2021reducingdomaingapreducing} & $60.89_{\pm1.41}$ & $60.05_{\pm0.66}$ & $61.83_{\pm1.12}$ & 60.92 & $61.26_{\pm0.75}$ & $61.99_{\pm1.24}$ & $59.93_{\pm0.48}$ & 61.06 & $58.85_{\pm0.33}$ & $56.94_{\pm1.51}$ & $58.24_{\pm0.91}$ & 58.01 \\
 &  & IB\_ERM~\cite{ahuja2022invarianceprinciplemeetsinformation} & $59.82_{\pm0.85}$ & $61.53_{\pm0.24}$ & $60.78_{\pm1.68}$ & 60.71 & $63.22_{\pm1.32}$ & $62.39_{\pm0.55}$ & $61.88_{\pm1.14}$ & 62.50 & $58.80_{\pm0.19}$ & $58.42_{\pm1.47}$ & $58.16_{\pm0.73}$ & 58.46 \\
 &  & CondCAD~\cite{ruan2022optimalrepresentationscovariateshift} & $59.52_{\pm1.55}$ & $59.01_{\pm0.37}$ & $60.94_{\pm1.18}$ & 59.82 & $63.57_{\pm0.81}$ & $63.49_{\pm1.62}$ & $62.67_{\pm0.29}$ & 63.24 & $59.68_{\pm1.04}$ & $59.99_{\pm0.52}$ & $59.05_{\pm1.37}$ & 59.57 \\
 &  & EQRM~\cite{eastwood2023probabledomaingeneralizationquantile} & $58.92_{\pm0.92}$ & $59.62_{\pm1.51}$ & $59.06_{\pm0.15}$ & 59.20 & $61.52_{\pm0.44}$ & $62.68_{\pm1.88}$ & $61.20_{\pm1.09}$ & 61.80 & $57.55_{\pm1.34}$ & $58.49_{\pm0.26}$ & $59.30_{\pm1.55}$ & 58.45 \\
 &  & ERM++~\cite{teterwak2024ermimprovedbaselinedomain} & $60.53_{\pm1.29}$ & $60.14_{\pm0.33}$ & $59.53_{\pm1.85}$ & 60.07 & $63.59_{\pm0.14}$ & $64.44_{\pm1.12}$ & $62.29_{\pm0.73}$ & 63.44 & $57.77_{\pm1.43}$ & $57.76_{\pm1.68}$ & $58.94_{\pm0.42}$ & 58.16 \\
 &  & URM~\cite{krishnamachari2024uniformly} & $59.95_{\pm0.48}$ & $58.69_{\pm1.04}$ & $59.68_{\pm1.37}$ & 59.44 & $61.66_{\pm1.55}$ & $63.79_{\pm0.91}$ & $62.40_{\pm1.11}$ & 62.62 & $58.04_{\pm1.83}$ & $58.08_{\pm0.55}$ & $57.98_{\pm1.29}$ & 58.03 \\

\midrule

\multirow{13}{*}{LanguageBind~\cite{zhu2023languagebind}} & \multirow{3}{*}{MML} & Concat & $60.91_{\pm1.12}$ & $59.79_{\pm0.37}$ & $61.11_{\pm1.68}$ & 60.60 & $63.11_{\pm1.44}$ & $63.22_{\pm0.15}$ & $63.53_{\pm0.92}$ & 63.29 & $58.44_{\pm0.73}$ & $58.17_{\pm1.55}$ & $58.56_{\pm0.29}$ & 58.39 \\
 &  & OGM~\cite{peng2022balancedmultimodallearningonthefly} & $59.61_{\pm0.83}$ & $57.62_{\pm1.24}$ & $60.20_{\pm0.19}$ & 59.14 & $61.97_{\pm1.37}$ & $61.91_{\pm0.48}$ & $60.83_{\pm1.05}$ & 61.57 & $56.07_{\pm0.96}$ & $56.55_{\pm1.88}$ & $57.57_{\pm0.33}$ & 56.73 \\
 &  & DLMG~\cite{NEURIPS2024_71b17f00} & $58.74_{\pm0.15}$ & $58.59_{\pm1.43}$ & $58.76_{\pm0.66}$ & 58.70 & $60.55_{\pm0.75}$ & $60.99_{\pm1.83}$ & $62.00_{\pm0.26}$ & 61.18 & $58.90_{\pm1.51}$ & $56.92_{\pm0.91}$ & $57.22_{\pm1.18}$ & 57.68 \\
\cmidrule{2-15}
 & \multirow{10}{*}{DG} & ERM & $57.74_{\pm1.47}$ & $60.16_{\pm0.39}$ & $60.05_{\pm1.26}$ & 59.32 & $60.81_{\pm1.18}$ & $61.03_{\pm0.92}$ & $63.33_{\pm0.52}$ & 61.72 & $57.39_{\pm1.62}$ & $58.08_{\pm0.15}$ & $56.63_{\pm1.09}$ & 57.37 \\
 &  & IRM~\cite{arjovsky2020invariantriskminimization} & $60.21_{\pm0.55}$ & $61.48_{\pm1.68}$ & $60.71_{\pm0.81}$ & 60.80 & $62.10_{\pm0.24}$ & $60.75_{\pm1.51}$ & $64.50_{\pm0.92}$ & 62.45 & $57.53_{\pm1.34}$ & $59.14_{\pm0.73}$ & $58.59_{\pm1.85}$ & 58.42 \\
 &  & Mixup~\cite{yan2020improveunsuperviseddomainadaptation} & $60.22_{\pm1.83}$ & $60.45_{\pm0.14}$ & $58.57_{\pm0.91}$ & 59.75 & $59.42_{\pm1.12}$ & $60.16_{\pm0.48}$ & $62.29_{\pm1.62}$ & 60.62 & $56.42_{\pm0.26}$ & $56.51_{\pm1.47}$ & $59.10_{\pm1.18}$ & 57.34 \\
 &  & CDANN~\cite{Li_2018_ECCV} & $59.85_{\pm0.29}$ & $59.65_{\pm1.55}$ & $59.32_{\pm1.09}$ & 59.61 & $59.13_{\pm0.73}$ & $60.78_{\pm1.24}$ & $62.40_{\pm0.33}$ & 60.77 & $59.02_{\pm1.88}$ & $60.83_{\pm0.42}$ & $58.64_{\pm1.51}$ & 59.50 \\
 &  & SagNet~\cite{nam2021reducingdomaingapreducing} & $59.22_{\pm1.62}$ & $58.75_{\pm0.84}$ & $61.51_{\pm1.18}$ & 59.83 & $60.36_{\pm0.15}$ & $60.29_{\pm1.47}$ & $63.23_{\pm0.56}$ & 61.29 & $58.33_{\pm0.92}$ & $58.13_{\pm1.73}$ & $58.14_{\pm0.39}$ & 58.20 \\
 &  & IB\_ERM~\cite{ahuja2022invarianceprinciplemeetsinformation} & $59.80_{\pm0.73}$ & $60.56_{\pm1.11}$ & $59.86_{\pm1.52}$ & 60.07 & $62.93_{\pm0.85}$ & $61.94_{\pm0.29}$ & $62.69_{\pm1.68}$ & 62.52 & $58.11_{\pm0.44}$ & $57.76_{\pm1.37}$ & $58.49_{\pm1.14}$ & 58.12 \\
 &  & CondCAD~\cite{ruan2022optimalrepresentationscovariateshift} & $60.88_{\pm1.15}$ & $60.72_{\pm0.64}$ & $60.87_{\pm1.83}$ & 60.82 & $63.04_{\pm0.33}$ & $63.04_{\pm1.12}$ & $61.86_{\pm1.55}$ & 62.65 & $59.29_{\pm0.26}$ & $60.38_{\pm1.29}$ & $60.81_{\pm0.91}$ & 60.16 \\
 &  & EQRM~\cite{eastwood2023probabledomaingeneralizationquantile} & $59.20_{\pm0.37}$ & $58.29_{\pm1.41}$ & $60.07_{\pm0.88}$ & 59.19 & $62.14_{\pm1.62}$ & $62.12_{\pm0.55}$ & $62.57_{\pm1.18}$ & 62.28 & $56.13_{\pm1.04}$ & $57.92_{\pm0.15}$ & $58.78_{\pm1.47}$ & 57.61 \\
 &  & ERM++~\cite{teterwak2024ermimprovedbaselinedomain} & $60.25_{\pm1.51}$ & $59.94_{\pm0.73}$ & $59.15_{\pm1.24}$ & 59.78 & $61.70_{\pm0.39}$ & $63.83_{\pm1.85}$ & $62.94_{\pm0.92}$ & 62.82 & $57.87_{\pm1.14}$ & $57.70_{\pm1.43}$ & $57.67_{\pm0.26}$ & 57.75 \\
 &  & URM~\cite{krishnamachari2024uniformly} & $58.17_{\pm0.92}$ & $57.85_{\pm1.34}$ & $59.45_{\pm0.11}$ & 58.49 & $61.03_{\pm1.68}$ & $62.32_{\pm0.83}$ & $63.77_{\pm1.41}$ & 62.37 & $56.86_{\pm0.55}$ & $59.52_{\pm1.12}$ & $57.10_{\pm1.88}$ & 57.83 \\

\midrule

\multirow{13}{*}{UniBind~\cite{lyu2024unibind}} & \multirow{3}{*}{MML} & Concat & $61.47_{\pm0.26}$ & $61.12_{\pm1.55}$ & $60.70_{\pm0.84}$ & 61.10 & $63.67_{\pm1.18}$ & $63.92_{\pm0.39}$ & $64.45_{\pm1.62}$ & 64.01 & $58.77_{\pm0.15}$ & $59.34_{\pm1.47}$ & $58.97_{\pm1.11}$ & 59.03 \\
 &  & OGM~\cite{peng2022balancedmultimodallearningonthefly} & $59.05_{\pm1.41}$ & $59.37_{\pm0.52}$ & $60.25_{\pm1.18}$ & 59.56 & $61.21_{\pm0.92}$ & $62.92_{\pm1.68}$ & $61.63_{\pm0.73}$ & 61.92 & $58.61_{\pm0.29}$ & $58.29_{\pm1.44}$ & $56.78_{\pm1.15}$ & 57.89 \\
 &  & DLMG~\cite{NEURIPS2024_71b17f00} & $60.96_{\pm0.85}$ & $58.79_{\pm1.62}$ & $60.46_{\pm0.14}$ & 60.07 & $61.85_{\pm1.37}$ & $61.19_{\pm0.75}$ & $62.37_{\pm1.11}$ & 61.80 & $58.31_{\pm1.51}$ & $57.56_{\pm0.48}$ & $57.45_{\pm1.09}$ & 57.77 \\
\cmidrule{2-15}
 & \multirow{10}{*}{DG} & ERM & $58.52_{\pm1.24}$ & $58.65_{\pm0.19}$ & $58.53_{\pm1.43}$ & 58.57 & $59.85_{\pm0.92}$ & $62.62_{\pm1.55}$ & $59.07_{\pm0.37}$ & 60.51 & $59.02_{\pm1.18}$ & $58.22_{\pm0.81}$ & $58.44_{\pm1.62}$ & 58.56 \\
 &  & IRM~\cite{arjovsky2020invariantriskminimization} & $59.79_{\pm0.33}$ & $61.16_{\pm1.12}$ & $61.76_{\pm1.55}$ & 60.90 & $61.73_{\pm0.73}$ & $60.85_{\pm0.26}$ & $62.92_{\pm1.14}$ & 61.83 & $56.79_{\pm1.47}$ & $58.90_{\pm0.55}$ & $58.79_{\pm1.88}$ & 58.16 \\
 &  & Mixup~\cite{yan2020improveunsuperviseddomainadaptation} & $59.41_{\pm1.15}$ & $59.59_{\pm0.39}$ & $58.55_{\pm1.62}$ & 59.18 & $59.31_{\pm1.44}$ & $61.06_{\pm0.52}$ & $60.51_{\pm1.11}$ & 60.29 & $58.48_{\pm0.73}$ & $56.80_{\pm1.18}$ & $56.58_{\pm1.51}$ & 57.29 \\
 &  & CDANN~\cite{Li_2018_ECCV} & $60.69_{\pm1.51}$ & $58.48_{\pm0.29}$ & $60.79_{\pm0.83}$ & 59.99 & $62.03_{\pm1.12}$ & $60.72_{\pm1.34}$ & $62.01_{\pm0.55}$ & 61.59 & $59.19_{\pm1.11}$ & $58.20_{\pm1.83}$ & $59.22_{\pm0.42}$ & 58.87 \\
 &  & SagNet~\cite{nam2021reducingdomaingapreducing} & $58.55_{\pm0.73}$ & $59.80_{\pm1.55}$ & $59.39_{\pm0.15}$ & 59.25 & $60.15_{\pm1.24}$ & $61.98_{\pm0.33}$ & $60.13_{\pm1.85}$ & 60.75 & $57.38_{\pm1.22}$ & $58.71_{\pm0.91}$ & $58.13_{\pm0.66}$ & 58.07 \\
 &  & IB\_ERM~\cite{ahuja2022invarianceprinciplemeetsinformation} & $59.29_{\pm1.68}$ & $60.34_{\pm0.92}$ & $58.94_{\pm0.48}$ & 59.52 & $61.21_{\pm1.15}$ & $62.56_{\pm0.73}$ & $62.67_{\pm1.47}$ & 62.15 & $57.57_{\pm0.82}$ & $60.09_{\pm1.11}$ & $58.55_{\pm1.55}$ & 58.74 \\
 &  & CondCAD~\cite{ruan2022optimalrepresentationscovariateshift} & $61.41_{\pm0.39}$ & $58.80_{\pm1.24}$ & $60.89_{\pm0.55}$ & 60.37 & $62.02_{\pm1.51}$ & $63.26_{\pm1.12}$ & $63.67_{\pm0.45}$ & 62.98 & $58.65_{\pm0.96}$ & $59.35_{\pm1.88}$ & $60.73_{\pm0.13}$ & 59.58 \\
 &  & EQRM~\cite{eastwood2023probabledomaingeneralizationquantile} & $58.84_{\pm1.55}$ & $59.97_{\pm0.64}$ & $59.63_{\pm1.37}$ & 59.48 & $60.57_{\pm0.26}$ & $61.17_{\pm1.15}$ & $63.44_{\pm0.52}$ & 61.73 & $59.73_{\pm0.88}$ & $57.01_{\pm1.62}$ & $57.52_{\pm0.39}$ & 58.09 \\
 &  & ERM++~\cite{teterwak2024ermimprovedbaselinedomain} & $58.42_{\pm0.75}$ & $59.07_{\pm1.14}$ & $59.99_{\pm0.37}$ & 59.16 & $62.75_{\pm1.43}$ & $63.20_{\pm0.91}$ & $63.05_{\pm1.29}$ & 63.00 & $57.19_{\pm0.55}$ & $58.04_{\pm1.12}$ & $59.55_{\pm1.63}$ & 58.26 \\
 &  & URM~\cite{krishnamachari2024uniformly} & $57.76_{\pm1.51}$ & $59.17_{\pm0.88}$ & $58.93_{\pm1.62}$ & 58.62 & $60.72_{\pm0.33}$ & $61.74_{\pm1.41}$ & $61.28_{\pm0.55}$ & 61.25 & $57.18_{\pm0.37}$ & $57.18_{\pm1.51}$ & $57.92_{\pm1.21}$ & 57.43 \\

\bottomrule
\end{tabular}
}
\end{table*}

\begin{table*}[t]
\centering
\caption{Classification performance (mean $\pm$ std) under the Strong MAF setting, using leave-one-modality-out cross-validation set for model selection.}
\label{tab:strong_leave_one_out_random}
\resizebox{\textwidth}{!}{ 
\begin{tabular}{l|ll | cccc | cccc | cccc}
\toprule
\multirow{2}{*}{\textbf{Perceptor}} & \multicolumn{2}{l|}{\multirow{2}{*}{\textbf{Method}}} & \multicolumn{4}{c|}{\textbf{LAV-DF~\cite{cai2023reallymeanthatcontent}}} & \multicolumn{4}{c|}{\textbf{Fakeavcele~\cite{khalid2022fakeavcelebnovelaudiovideomultimodal}}} & \multicolumn{4}{c}{\textbf{Cele+Asv~\cite{wang2024asvspoof5crowdsourcedspeech, li2025celeb}}} \\
\cmidrule(lr){4-7} \cmidrule(lr){8-11} \cmidrule(lr){12-15}
& & & \textbf{Vid} & \textbf{Aud} & \textbf{Img} & \textbf{Avg} & \textbf{Vid} & \textbf{Aud} & \textbf{Img} & \textbf{Avg} & \textbf{Vid} & \textbf{Aud} & \textbf{Img} & \textbf{Avg} \\
\midrule
\midrule

\multirow{13}{*}{ImageBind~\cite{girdhar2023imagebind}} & \multirow{3}{*}{MML} & Concat & $59.62_{\pm1.99}$ & $59.31_{\pm0.88}$ & $59.67_{\pm1.77}$ & 59.53 & $63.33_{\pm1.28}$ & $64.07_{\pm0.35}$ & $61.78_{\pm1.71}$ & 63.06 & $59.15_{\pm0.62}$ & $59.41_{\pm1.36}$ & $60.09_{\pm0.98}$ & 59.55 \\
 &  & OGM~\cite{peng2022balancedmultimodallearningonthefly} & $59.35_{\pm1.03}$ & $59.10_{\pm0.65}$ & $60.92_{\pm1.19}$ & 59.79 & $62.93_{\pm0.25}$ & $61.69_{\pm0.85}$ & $63.13_{\pm0.30}$ & 62.58 & $57.95_{\pm1.69}$ & $59.33_{\pm0.61}$ & $58.10_{\pm1.03}$ & 58.46 \\
 &  & DLMG~\cite{NEURIPS2024_71b17f00} & $60.72_{\pm1.67}$ & $62.01_{\pm1.78}$ & $60.53_{\pm1.93}$ & 61.09 & $62.88_{\pm0.14}$ & $62.09_{\pm1.94}$ & $63.48_{\pm1.28}$ & 62.82 & $57.28_{\pm1.22}$ & $58.35_{\pm0.51}$ & $57.18_{\pm1.72}$ & 57.60 \\
\cmidrule{2-15}
 & \multirow{10}{*}{DG} & ERM & $61.36_{\pm1.88}$ & $60.71_{\pm0.45}$ & $58.69_{\pm0.95}$ & 60.25 & $60.37_{\pm1.39}$ & $61.11_{\pm0.42}$ & $62.17_{\pm1.84}$ & 61.22 & $59.42_{\pm0.93}$ & $58.76_{\pm1.28}$ & $58.43_{\pm1.20}$ & 58.87 \\
 &  & IRM~\cite{arjovsky2020invariantriskminimization} & $63.83_{\pm1.13}$ & $61.86_{\pm1.42}$ & $62.22_{\pm0.12}$ & 62.64 & $61.00_{\pm0.43}$ & $61.10_{\pm1.10}$ & $63.96_{\pm0.30}$ & 62.02 & $59.00_{\pm0.29}$ & $57.41_{\pm1.47}$ & $59.28_{\pm1.30}$ & 58.56 \\
 &  & Mixup~\cite{yan2020improveunsuperviseddomainadaptation} & $60.95_{\pm1.02}$ & $60.01_{\pm1.77}$ & $60.47_{\pm1.50}$ & 60.48 & $61.61_{\pm0.79}$ & $60.81_{\pm0.66}$ & $62.70_{\pm1.05}$ & 61.71 & $59.03_{\pm1.85}$ & $57.44_{\pm1.17}$ & $59.62_{\pm1.18}$ & 58.70 \\
 &  & CDANN~\cite{Li_2018_ECCV} & $60.31_{\pm1.08}$ & $60.18_{\pm1.51}$ & $60.51_{\pm1.82}$ & 60.33 & $61.84_{\pm1.22}$ & $62.68_{\pm0.15}$ & $62.36_{\pm0.94}$ & 62.29 & $60.96_{\pm1.10}$ & $60.12_{\pm0.80}$ & $58.81_{\pm1.93}$ & 59.96 \\
 &  & SagNet~\cite{nam2021reducingdomaingapreducing} & $60.96_{\pm0.94}$ & $60.74_{\pm1.45}$ & $61.73_{\pm1.33}$ & 61.14 & $60.02_{\pm0.50}$ & $62.22_{\pm1.75}$ & $62.56_{\pm1.47}$ & 61.60 & $57.78_{\pm1.84}$ & $57.08_{\pm0.23}$ & $58.79_{\pm0.12}$ & 57.88 \\
 &  & IB\_ERM~\cite{ahuja2022invarianceprinciplemeetsinformation} & $58.72_{\pm0.53}$ & $61.24_{\pm0.54}$ & $59.63_{\pm0.41}$ & 59.86 & $62.30_{\pm0.72}$ & $61.07_{\pm1.28}$ & $63.40_{\pm0.12}$ & 62.26 & $59.61_{\pm0.45}$ & $59.73_{\pm1.04}$ & $59.78_{\pm1.88}$ & 59.71 \\
 &  & CondCAD~\cite{ruan2022optimalrepresentationscovariateshift} & $59.26_{\pm1.77}$ & $60.48_{\pm1.43}$ & $59.70_{\pm0.68}$ & 59.81 & $61.21_{\pm1.55}$ & $61.86_{\pm1.18}$ & $63.45_{\pm1.01}$ & 62.17 & $59.33_{\pm1.23}$ & $61.58_{\pm0.67}$ & $60.55_{\pm1.38}$ & 60.49 \\
 &  & EQRM~\cite{eastwood2023probabledomaingeneralizationquantile} & $60.99_{\pm1.49}$ & $61.97_{\pm1.28}$ & $59.35_{\pm0.63}$ & 60.77 & $61.22_{\pm0.19}$ & $61.97_{\pm1.86}$ & $64.15_{\pm0.99}$ & 62.45 & $58.99_{\pm0.62}$ & $58.60_{\pm0.70}$ & $58.96_{\pm0.87}$ & 58.85 \\
 &  & ERM++~\cite{teterwak2024ermimprovedbaselinedomain} & $60.77_{\pm1.31}$ & $61.35_{\pm0.82}$ & $61.36_{\pm0.78}$ & 61.16 & $61.38_{\pm1.60}$ & $63.20_{\pm1.83}$ & $60.90_{\pm1.01}$ & 61.83 & $58.58_{\pm0.45}$ & $57.74_{\pm1.26}$ & $58.35_{\pm1.81}$ & 58.22 \\
 &  & URM~\cite{krishnamachari2024uniformly} & $58.86_{\pm0.76}$ & $60.57_{\pm1.01}$ & $61.53_{\pm1.39}$ & 60.32 & $60.71_{\pm0.40}$ & $63.50_{\pm1.47}$ & $61.00_{\pm1.54}$ & 61.74 & $57.99_{\pm0.69}$ & $57.98_{\pm1.44}$ & $59.46_{\pm1.56}$ & 58.48 \\
\midrule
\multirow{13}{*}{LanguageBind~\cite{zhu2023languagebind}} & \multirow{3}{*}{MML} & Concat & $59.47_{\pm0.19}$ & $59.15_{\pm1.01}$ & $59.82_{\pm1.19}$ & 59.48 & $62.77_{\pm0.49}$ & $63.45_{\pm0.50}$ & $62.94_{\pm1.70}$ & 63.05 & $57.31_{\pm1.27}$ & $59.26_{\pm0.67}$ & $57.78_{\pm1.04}$ & 58.12 \\
 &  & OGM~\cite{peng2022balancedmultimodallearningonthefly} & $58.77_{\pm0.60}$ & $58.90_{\pm0.65}$ & $60.63_{\pm1.32}$ & 59.43 & $60.52_{\pm1.12}$ & $62.32_{\pm0.68}$ & $62.68_{\pm1.16}$ & 61.84 & $58.93_{\pm1.67}$ & $58.82_{\pm1.52}$ & $58.68_{\pm1.97}$ & 58.81 \\
 &  & DLMG~\cite{NEURIPS2024_71b17f00} & $59.71_{\pm1.43}$ & $61.28_{\pm1.03}$ & $59.94_{\pm0.73}$ & 60.31 & $61.77_{\pm0.31}$ & $60.46_{\pm0.33}$ & $61.90_{\pm1.23}$ & 61.38 & $58.21_{\pm1.61}$ & $58.03_{\pm0.75}$ & $57.52_{\pm1.97}$ & 57.92 \\
\cmidrule{2-15}
 & \multirow{10}{*}{DG} & ERM & $60.85_{\pm1.81}$ & $59.08_{\pm0.13}$ & $60.33_{\pm1.37}$ & 60.09 & $60.03_{\pm1.23}$ & $59.89_{\pm0.40}$ & $61.21_{\pm0.79}$ & 60.38 & $58.87_{\pm1.13}$ & $58.70_{\pm1.86}$ & $58.13_{\pm0.45}$ & 58.57 \\
 &  & IRM~\cite{arjovsky2020invariantriskminimization} & $63.11_{\pm0.72}$ & $61.55_{\pm0.11}$ & $60.10_{\pm0.15}$ & 61.59 & $59.80_{\pm1.49}$ & $59.73_{\pm1.03}$ & $63.40_{\pm1.40}$ & 60.98 & $57.47_{\pm1.48}$ & $58.57_{\pm1.01}$ & $58.16_{\pm1.64}$ & 58.07 \\
 &  & Mixup~\cite{yan2020improveunsuperviseddomainadaptation} & $60.60_{\pm1.08}$ & $60.65_{\pm0.97}$ & $61.98_{\pm0.82}$ & 61.08 & $61.11_{\pm1.18}$ & $60.53_{\pm1.65}$ & $62.71_{\pm1.88}$ & 61.45 & $59.52_{\pm1.95}$ & $56.87_{\pm0.34}$ & $58.99_{\pm0.39}$ & 58.46 \\
 &  & CDANN~\cite{Li_2018_ECCV} & $59.70_{\pm0.96}$ & $60.42_{\pm0.66}$ & $59.62_{\pm0.65}$ & 59.91 & $61.17_{\pm1.17}$ & $62.01_{\pm1.04}$ & $61.94_{\pm0.69}$ & 61.71 & $59.42_{\pm1.38}$ & $59.07_{\pm1.14}$ & $60.04_{\pm0.86}$ & 59.51 \\
 &  & SagNet~\cite{nam2021reducingdomaingapreducing} & $60.64_{\pm0.67}$ & $59.35_{\pm1.10}$ & $61.15_{\pm0.92}$ & 60.38 & $61.53_{\pm0.84}$ & $60.84_{\pm0.20}$ & $62.03_{\pm1.65}$ & 61.47 & $58.20_{\pm0.60}$ & $59.20_{\pm0.78}$ & $58.53_{\pm0.34}$ & 58.64 \\
 &  & IB\_ERM~\cite{ahuja2022invarianceprinciplemeetsinformation} & $61.47_{\pm0.96}$ & $60.55_{\pm0.91}$ & $61.84_{\pm0.65}$ & 61.29 & $63.29_{\pm1.38}$ & $60.60_{\pm0.47}$ & $62.74_{\pm1.75}$ & 62.21 & $58.64_{\pm1.80}$ & $59.16_{\pm1.72}$ & $58.77_{\pm0.10}$ & 58.86 \\
 &  & CondCAD~\cite{ruan2022optimalrepresentationscovariateshift} & $58.99_{\pm1.11}$ & $59.58_{\pm0.83}$ & $60.52_{\pm0.84}$ & 59.70 & $61.94_{\pm1.38}$ & $61.67_{\pm0.65}$ & $62.53_{\pm0.29}$ & 62.05 & $60.15_{\pm0.43}$ & $58.95_{\pm1.43}$ & $59.25_{\pm0.79}$ & 59.45 \\
 &  & EQRM~\cite{eastwood2023probabledomaingeneralizationquantile} & $60.26_{\pm0.56}$ & $61.46_{\pm1.48}$ & $60.10_{\pm0.15}$ & 60.61 & $60.57_{\pm1.20}$ & $61.62_{\pm0.58}$ & $63.61_{\pm0.65}$ & 61.93 & $57.83_{\pm1.04}$ & $57.59_{\pm1.88}$ & $58.60_{\pm1.93}$ & 58.01 \\
 &  & ERM++~\cite{teterwak2024ermimprovedbaselinedomain} & $60.08_{\pm0.36}$ & $60.73_{\pm1.86}$ & $59.48_{\pm0.42}$ & 60.10 & $60.13_{\pm0.85}$ & $62.56_{\pm1.15}$ & $60.40_{\pm1.82}$ & 61.03 & $58.88_{\pm1.92}$ & $59.79_{\pm0.72}$ & $57.89_{\pm1.93}$ & 58.85 \\
 &  & URM~\cite{krishnamachari2024uniformly} & $60.32_{\pm1.23}$ & $60.24_{\pm0.15}$ & $60.32_{\pm0.93}$ & 60.29 & $61.84_{\pm1.28}$ & $62.93_{\pm1.28}$ & $59.88_{\pm1.31}$ & 61.55 & $57.47_{\pm1.32}$ & $57.77_{\pm1.70}$ & $58.90_{\pm0.97}$ & 58.05 \\
\midrule
\multirow{13}{*}{UniBind~\cite{lyu2024unibind}} & \multirow{3}{*}{MML} & Concat & $60.17_{\pm0.23}$ & $60.28_{\pm0.38}$ & $59.86_{\pm0.67}$ & 60.10 & $59.99_{\pm1.77}$ & $63.13_{\pm1.46}$ & $60.37_{\pm0.75}$ & 61.16 & $57.56_{\pm1.57}$ & $59.15_{\pm1.28}$ & $58.41_{\pm1.91}$ & 58.37 \\
 &  & OGM~\cite{peng2022balancedmultimodallearningonthefly} & $60.14_{\pm1.28}$ & $59.08_{\pm0.78}$ & $58.63_{\pm1.43}$ & 59.28 & $61.78_{\pm0.90}$ & $62.98_{\pm1.32}$ & $61.30_{\pm1.54}$ & 62.02 & $57.28_{\pm0.96}$ & $59.04_{\pm1.98}$ & $57.37_{\pm1.86}$ & 57.90 \\
 &  & DLMG~\cite{NEURIPS2024_71b17f00} & $59.08_{\pm0.62}$ & $59.93_{\pm1.87}$ & $61.02_{\pm1.81}$ & 60.01 & $61.24_{\pm0.97}$ & $62.45_{\pm1.90}$ & $63.76_{\pm1.89}$ & 62.48 & $58.42_{\pm1.90}$ & $59.24_{\pm0.56}$ & $59.04_{\pm0.72}$ & 58.90 \\
\cmidrule{2-15}
 & \multirow{10}{*}{DG} & ERM & $60.69_{\pm0.30}$ & $59.12_{\pm1.22}$ & $60.20_{\pm1.02}$ & 60.00 & $61.57_{\pm1.97}$ & $59.87_{\pm1.36}$ & $61.46_{\pm0.44}$ & 60.97 & $57.68_{\pm0.57}$ & $58.02_{\pm1.47}$ & $57.41_{\pm0.52}$ & 57.70 \\
 &  & IRM~\cite{arjovsky2020invariantriskminimization} & $60.80_{\pm1.32}$ & $60.90_{\pm0.40}$ & $61.24_{\pm1.67}$ & 60.98 & $60.15_{\pm0.30}$ & $59.65_{\pm0.83}$ & $62.84_{\pm1.16}$ & 60.88 & $58.33_{\pm1.18}$ & $57.88_{\pm0.81}$ & $57.47_{\pm0.15}$ & 57.89 \\
 &  & Mixup~\cite{yan2020improveunsuperviseddomainadaptation} & $60.14_{\pm1.01}$ & $60.15_{\pm1.81}$ & $60.85_{\pm0.18}$ & 60.38 & $60.09_{\pm0.79}$ & $60.06_{\pm0.69}$ & $62.81_{\pm0.99}$ & 60.99 & $58.12_{\pm0.91}$ & $58.08_{\pm1.33}$ & $58.62_{\pm1.96}$ & 58.27 \\
 &  & CDANN~\cite{Li_2018_ECCV} & $61.24_{\pm1.28}$ & $60.90_{\pm1.65}$ & $58.87_{\pm1.62}$ & 60.34 & $62.39_{\pm0.39}$ & $62.06_{\pm0.97}$ & $63.38_{\pm0.42}$ & 62.61 & $59.10_{\pm1.51}$ & $60.32_{\pm1.52}$ & $59.41_{\pm1.82}$ & 59.61 \\
 &  & SagNet~\cite{nam2021reducingdomaingapreducing} & $59.68_{\pm0.91}$ & $59.54_{\pm1.52}$ & $59.83_{\pm1.38}$ & 59.68 & $59.45_{\pm0.61}$ & $59.41_{\pm1.32}$ & $61.73_{\pm1.53}$ & 60.20 & $58.63_{\pm1.76}$ & $57.09_{\pm0.74}$ & $58.33_{\pm1.92}$ & 58.02 \\
 &  & IB\_ERM~\cite{ahuja2022invarianceprinciplemeetsinformation} & $60.38_{\pm0.20}$ & $58.89_{\pm0.77}$ & $59.86_{\pm0.48}$ & 59.71 & $63.21_{\pm1.50}$ & $61.03_{\pm0.40}$ & $62.83_{\pm1.36}$ & 62.36 & $57.76_{\pm1.55}$ & $59.01_{\pm0.51}$ & $58.55_{\pm1.40}$ & 58.44 \\
 &  & CondCAD~\cite{ruan2022optimalrepresentationscovariateshift} & $59.53_{\pm1.30}$ & $60.98_{\pm1.26}$ & $58.70_{\pm0.47}$ & 59.74 & $61.76_{\pm1.00}$ & $60.72_{\pm1.52}$ & $61.80_{\pm1.51}$ & 61.43 & $61.30_{\pm0.47}$ & $59.40_{\pm1.74}$ & $59.42_{\pm0.60}$ & 60.04 \\
 &  & EQRM~\cite{eastwood2023probabledomaingeneralizationquantile} & $59.43_{\pm0.67}$ & $58.99_{\pm0.15}$ & $61.67_{\pm0.46}$ & 60.03 & $60.99_{\pm1.54}$ & $60.18_{\pm1.80}$ & $62.47_{\pm0.73}$ & 61.21 & $56.82_{\pm1.91}$ & $59.65_{\pm0.57}$ & $57.02_{\pm0.93}$ & 57.83 \\
 &  & ERM++~\cite{teterwak2024ermimprovedbaselinedomain} & $60.73_{\pm1.63}$ & $59.27_{\pm1.90}$ & $59.43_{\pm0.68}$ & 59.81 & $62.69_{\pm1.25}$ & $61.50_{\pm0.19}$ & $62.98_{\pm0.70}$ & 62.39 & $59.35_{\pm1.83}$ & $58.16_{\pm0.14}$ & $58.79_{\pm0.35}$ & 58.77 \\
 &  & URM~\cite{krishnamachari2024uniformly} & $59.44_{\pm0.50}$ & $60.07_{\pm0.63}$ & $58.25_{\pm0.98}$ & 59.25 & $61.48_{\pm1.35}$ & $60.41_{\pm0.21}$ & $60.98_{\pm1.19}$ & 60.96 & $55.94_{\pm1.64}$ & $57.60_{\pm0.57}$ & $59.08_{\pm0.21}$ & 57.54 \\

\bottomrule
\end{tabular}
}
\end{table*}

\begin{table*}[t]
\centering
\caption{Classification performance (mean $\pm$ std) under the Strong MAF setting, using training-modality validation set for model selection.}
\label{tab:strong_training_val_random}
\resizebox{\textwidth}{!}{ 
\begin{tabular}{l|ll | cccc | cccc | cccc}
\toprule
\multirow{2}{*}{\textbf{Perceptor}} & \multicolumn{2}{l|}{\multirow{2}{*}{\textbf{Method}}} & \multicolumn{4}{c|}{\textbf{LAV-DF~\cite{cai2023reallymeanthatcontent}}} & \multicolumn{4}{c|}{\textbf{Fakeavcele~\cite{khalid2022fakeavcelebnovelaudiovideomultimodal}}} & \multicolumn{4}{c}{\textbf{Cele+Asv~\cite{wang2024asvspoof5crowdsourcedspeech, li2025celeb}}} \\
\cmidrule(lr){4-7} \cmidrule(lr){8-11} \cmidrule(lr){12-15}
& & & \textbf{Vid} & \textbf{Aud} & \textbf{Img} & \textbf{Avg} & \textbf{Vid} & \textbf{Aud} & \textbf{Img} & \textbf{Avg} & \textbf{Vid} & \textbf{Aud} & \textbf{Img} & \textbf{Avg} \\
\midrule
\midrule

\multirow{13}{*}{ImageBind~\cite{girdhar2023imagebind}} & \multirow{3}{*}{MML} & Concat & $61.05_{\pm0.72}$ & $59.47_{\pm0.54}$ & $60.08_{\pm1.02}$ & 60.20 & $63.88_{\pm0.71}$ & $62.63_{\pm0.66}$ & $64.50_{\pm1.25}$ & 63.67 & $59.75_{\pm1.91}$ & $58.49_{\pm1.65}$ & $58.70_{\pm1.84}$ & 58.98 \\
 &  & OGM~\cite{peng2022balancedmultimodallearningonthefly} & $61.34_{\pm0.26}$ & $61.53_{\pm1.02}$ & $58.80_{\pm0.36}$ & 60.56 & $63.55_{\pm1.55}$ & $64.56_{\pm1.82}$ & $63.88_{\pm0.16}$ & 64.00 & $56.72_{\pm0.84}$ & $58.38_{\pm0.56}$ & $58.76_{\pm1.01}$ & 57.95 \\
 &  & DLMG~\cite{NEURIPS2024_71b17f00} & $61.53_{\pm1.09}$ & $62.05_{\pm1.35}$ & $60.60_{\pm0.29}$ & 61.39 & $64.58_{\pm1.31}$ & $64.56_{\pm0.19}$ & $63.52_{\pm0.22}$ & 64.22 & $57.52_{\pm0.69}$ & $58.22_{\pm1.25}$ & $58.13_{\pm0.84}$ & 57.96 \\
\cmidrule{2-15}
 & \multirow{10}{*}{DG} & ERM & $60.51_{\pm1.12}$ & $62.05_{\pm1.61}$ & $61.86_{\pm1.00}$ & 61.47 & $62.58_{\pm0.97}$ & $62.29_{\pm0.34}$ & $61.97_{\pm1.69}$ & 62.28 & $59.18_{\pm1.15}$ & $56.62_{\pm1.42}$ & $56.42_{\pm1.52}$ & 57.41 \\
 &  & IRM~\cite{arjovsky2020invariantriskminimization} & $61.45_{\pm1.94}$ & $62.03_{\pm1.90}$ & $60.56_{\pm0.99}$ & 61.35 & $63.79_{\pm1.72}$ & $62.96_{\pm1.09}$ & $64.44_{\pm0.22}$ & 63.73 & $60.20_{\pm1.34}$ & $60.21_{\pm0.53}$ & $57.59_{\pm1.99}$ & 59.33 \\
 &  & Mixup~\cite{yan2020improveunsuperviseddomainadaptation} & $60.86_{\pm0.59}$ & $61.25_{\pm1.14}$ & $59.69_{\pm0.15}$ & 60.60 & $62.34_{\pm1.27}$ & $62.53_{\pm1.42}$ & $62.46_{\pm1.32}$ & 62.44 & $59.28_{\pm0.67}$ & $58.21_{\pm0.92}$ & $58.50_{\pm0.62}$ & 58.66 \\
 &  & CDANN~\cite{Li_2018_ECCV} & $60.31_{\pm1.55}$ & $59.24_{\pm0.31}$ & $59.45_{\pm0.21}$ & 59.67 & $62.14_{\pm1.96}$ & $64.09_{\pm0.39}$ & $63.60_{\pm1.11}$ & 63.28 & $60.13_{\pm0.90}$ & $60.89_{\pm0.87}$ & $59.04_{\pm0.40}$ & 60.02 \\
 &  & SagNet~\cite{nam2021reducingdomaingapreducing} & $59.80_{\pm1.63}$ & $61.47_{\pm1.82}$ & $61.24_{\pm0.82}$ & 60.84 & $64.19_{\pm1.72}$ & $63.15_{\pm0.30}$ & $63.74_{\pm1.31}$ & 63.69 & $58.57_{\pm1.38}$ & $58.96_{\pm1.01}$ & $60.32_{\pm0.81}$ & 59.28 \\
 &  & IB\_ERM~\cite{ahuja2022invarianceprinciplemeetsinformation} & $60.99_{\pm0.63}$ & $61.21_{\pm0.17}$ & $61.83_{\pm0.42}$ & 61.34 & $62.59_{\pm1.65}$ & $61.38_{\pm1.15}$ & $64.94_{\pm1.07}$ & 62.97 & $58.48_{\pm0.25}$ & $59.00_{\pm1.60}$ & $59.33_{\pm0.95}$ & 58.94 \\
 &  & CondCAD~\cite{ruan2022optimalrepresentationscovariateshift} & $59.84_{\pm0.72}$ & $59.99_{\pm1.51}$ & $60.13_{\pm0.25}$ & 59.99 & $64.46_{\pm0.18}$ & $62.31_{\pm1.44}$ & $61.74_{\pm0.51}$ & 62.84 & $60.10_{\pm0.84}$ & $59.08_{\pm1.67}$ & $59.48_{\pm1.25}$ & 59.55 \\
 &  & EQRM~\cite{eastwood2023probabledomaingeneralizationquantile} & $60.57_{\pm1.05}$ & $61.55_{\pm0.71}$ & $59.00_{\pm1.08}$ & 60.37 & $64.25_{\pm0.67}$ & $64.27_{\pm1.80}$ & $62.51_{\pm0.28}$ & 63.68 & $59.12_{\pm1.92}$ & $58.23_{\pm0.78}$ & $57.13_{\pm0.50}$ & 58.16 \\
 &  & ERM++~\cite{teterwak2024ermimprovedbaselinedomain} & $59.64_{\pm1.57}$ & $61.21_{\pm1.17}$ & $61.63_{\pm1.91}$ & 60.83 & $62.19_{\pm1.12}$ & $63.74_{\pm0.20}$ & $64.66_{\pm1.92}$ & 63.53 & $59.13_{\pm0.77}$ & $59.03_{\pm0.64}$ & $58.18_{\pm1.68}$ & 58.78 \\
 &  & URM~\cite{krishnamachari2024uniformly} & $60.51_{\pm1.47}$ & $59.56_{\pm0.35}$ & $59.48_{\pm1.98}$ & 59.85 & $62.78_{\pm0.82}$ & $62.83_{\pm0.67}$ & $63.01_{\pm0.60}$ & 62.87 & $58.73_{\pm0.72}$ & $58.50_{\pm1.28}$ & $57.24_{\pm0.29}$ & 58.16 \\

\midrule

\multirow{13}{*}{LanguageBind~\cite{zhu2023languagebind}} & \multirow{3}{*}{MML} & Concat & $61.32_{\pm1.84}$ & $61.10_{\pm0.47}$ & $61.03_{\pm0.63}$ & 61.15 & $62.67_{\pm1.71}$ & $64.78_{\pm0.88}$ & $63.15_{\pm1.47}$ & 63.53 & $59.05_{\pm1.19}$ & $57.50_{\pm1.85}$ & $57.65_{\pm0.55}$ & 58.07 \\
 &  & OGM~\cite{peng2022balancedmultimodallearningonthefly} & $60.30_{\pm1.08}$ & $61.17_{\pm1.10}$ & $60.89_{\pm1.88}$ & 60.79 & $62.81_{\pm0.14}$ & $64.09_{\pm1.95}$ & $63.52_{\pm1.81}$ & 63.47 & $57.49_{\pm0.92}$ & $58.32_{\pm1.30}$ & $57.91_{\pm0.29}$ & 57.91 \\
 &  & DLMG~\cite{NEURIPS2024_71b17f00} & $59.55_{\pm1.99}$ & $61.44_{\pm1.91}$ & $60.11_{\pm0.59}$ & 60.37 & $64.20_{\pm1.12}$ & $63.61_{\pm0.33}$ & $63.53_{\pm1.63}$ & 63.78 & $57.34_{\pm1.73}$ & $57.65_{\pm1.10}$ & $57.59_{\pm1.66}$ & 57.53 \\
\cmidrule{2-15}
 & \multirow{10}{*}{DG} & ERM & $59.43_{\pm1.67}$ & $60.27_{\pm0.70}$ & $61.20_{\pm0.35}$ & 60.30 & $62.43_{\pm1.70}$ & $61.57_{\pm0.12}$ & $64.30_{\pm1.42}$ & 62.77 & $58.85_{\pm1.19}$ & $57.28_{\pm1.73}$ & $57.61_{\pm1.31}$ & 57.91 \\
 &  & IRM~\cite{arjovsky2020invariantriskminimization} & $61.01_{\pm0.57}$ & $61.47_{\pm1.34}$ & $59.97_{\pm0.16}$ & 60.82 & $63.50_{\pm1.22}$ & $62.23_{\pm1.80}$ & $63.94_{\pm1.13}$ & 63.22 & $59.58_{\pm1.93}$ & $59.87_{\pm1.36}$ & $57.29_{\pm1.07}$ & 58.91 \\
 &  & Mixup~\cite{yan2020improveunsuperviseddomainadaptation} & $58.80_{\pm0.26}$ & $60.80_{\pm1.52}$ & $58.99_{\pm1.88}$ & 59.53 & $61.03_{\pm1.26}$ & $62.60_{\pm0.82}$ & $63.32_{\pm1.52}$ & 62.32 & $58.81_{\pm1.50}$ & $59.15_{\pm0.54}$ & $58.65_{\pm1.31}$ & 58.87 \\
 &  & CDANN~\cite{Li_2018_ECCV} & $59.85_{\pm1.02}$ & $60.26_{\pm1.15}$ & $61.87_{\pm1.48}$ & 60.66 & $61.83_{\pm1.92}$ & $63.77_{\pm1.07}$ & $63.99_{\pm0.92}$ & 63.20 & $58.50_{\pm0.94}$ & $60.61_{\pm1.30}$ & $58.88_{\pm0.31}$ & 59.33 \\
 &  & SagNet~\cite{nam2021reducingdomaingapreducing} & $59.68_{\pm0.89}$ & $60.23_{\pm0.29}$ & $59.67_{\pm0.29}$ & 59.86 & $63.55_{\pm1.08}$ & $62.25_{\pm0.25}$ & $62.07_{\pm0.36}$ & 62.62 & $57.76_{\pm1.79}$ & $58.83_{\pm0.48}$ & $59.65_{\pm1.87}$ & 58.75 \\
 &  & IB\_ERM~\cite{ahuja2022invarianceprinciplemeetsinformation} & $59.77_{\pm1.88}$ & $60.36_{\pm0.63}$ & $60.78_{\pm1.95}$ & 60.30 & $62.43_{\pm0.64}$ & $62.76_{\pm0.90}$ & $62.50_{\pm0.15}$ & 62.56 & $58.24_{\pm0.42}$ & $58.32_{\pm0.63}$ & $59.02_{\pm1.44}$ & 58.53 \\
 &  & CondCAD~\cite{ruan2022optimalrepresentationscovariateshift} & $59.60_{\pm1.98}$ & $59.56_{\pm0.69}$ & $59.74_{\pm0.29}$ & 59.63 & $64.16_{\pm1.74}$ & $61.85_{\pm1.52}$ & $62.32_{\pm1.57}$ & 62.78 & $59.84_{\pm0.60}$ & $58.86_{\pm0.38}$ & $60.69_{\pm0.57}$ & 59.80 \\
 &  & EQRM~\cite{eastwood2023probabledomaingeneralizationquantile} & $59.43_{\pm0.28}$ & $59.68_{\pm0.43}$ & $59.63_{\pm0.61}$ & 59.58 & $62.61_{\pm0.85}$ & $64.02_{\pm1.01}$ & $64.31_{\pm1.78}$ & 63.65 & $58.65_{\pm0.61}$ & $57.60_{\pm1.43}$ & $58.89_{\pm0.71}$ & 58.38 \\
 &  & ERM++~\cite{teterwak2024ermimprovedbaselinedomain} & $58.98_{\pm0.94}$ & $60.85_{\pm1.42}$ & $61.13_{\pm1.00}$ & 60.32 & $63.04_{\pm0.98}$ & $63.37_{\pm0.92}$ & $63.34_{\pm1.39}$ & 63.25 & $57.60_{\pm0.10}$ & $58.76_{\pm0.86}$ & $57.36_{\pm0.53}$ & 57.91 \\
 &  & URM~\cite{krishnamachari2024uniformly} & $60.03_{\pm1.90}$ & $60.39_{\pm1.60}$ & $60.16_{\pm0.76}$ & 60.19 & $63.35_{\pm0.22}$ & $62.80_{\pm1.04}$ & $63.72_{\pm1.36}$ & 63.29 & $58.25_{\pm1.08}$ & $58.09_{\pm1.55}$ & $59.31_{\pm0.41}$ & 58.55 \\

\midrule

\multirow{13}{*}{UniBind~\cite{lyu2024unibind}} & \multirow{3}{*}{MML} & Concat & $60.53_{\pm1.76}$ & $60.42_{\pm1.33}$ & $60.09_{\pm1.65}$ & 60.35 & $64.39_{\pm1.34}$ & $62.50_{\pm1.87}$ & $63.56_{\pm1.28}$ & 63.48 & $58.75_{\pm1.51}$ & $58.06_{\pm0.99}$ & $59.07_{\pm1.12}$ & 58.63 \\
 &  & OGM~\cite{peng2022balancedmultimodallearningonthefly} & $60.15_{\pm1.28}$ & $60.76_{\pm1.19}$ & $61.08_{\pm0.81}$ & 60.66 & $64.10_{\pm1.68}$ & $62.94_{\pm1.39}$ & $62.49_{\pm1.81}$ & 63.18 & $59.02_{\pm0.17}$ & $57.14_{\pm1.47}$ & $58.17_{\pm0.69}$ & 58.11 \\
 &  & DLMG~\cite{NEURIPS2024_71b17f00} & $60.71_{\pm0.59}$ & $61.43_{\pm0.78}$ & $60.56_{\pm0.96}$ & 60.90 & $63.61_{\pm1.30}$ & $63.64_{\pm1.39}$ & $62.86_{\pm1.38}$ & 63.37 & $59.36_{\pm0.46}$ & $59.54_{\pm1.86}$ & $57.48_{\pm0.94}$ & 58.79 \\
\cmidrule{2-15}
 & \multirow{10}{*}{DG} & ERM & $59.97_{\pm1.00}$ & $59.73_{\pm1.31}$ & $59.93_{\pm1.53}$ & 59.88 & $61.69_{\pm1.23}$ & $61.33_{\pm1.78}$ & $64.21_{\pm0.75}$ & 62.41 & $58.14_{\pm0.69}$ & $57.44_{\pm1.18}$ & $57.26_{\pm0.21}$ & 57.61 \\
 &  & IRM~\cite{arjovsky2020invariantriskminimization} & $61.86_{\pm0.13}$ & $62.43_{\pm1.81}$ & $60.80_{\pm0.36}$ & 61.70 & $62.38_{\pm0.71}$ & $60.45_{\pm0.46}$ & $64.49_{\pm0.80}$ & 62.44 & $59.96_{\pm0.68}$ & $59.12_{\pm0.20}$ & $60.26_{\pm1.47}$ & 59.78 \\
 &  & Mixup~\cite{yan2020improveunsuperviseddomainadaptation} & $61.34_{\pm1.25}$ & $59.10_{\pm1.94}$ & $60.63_{\pm1.84}$ & 60.36 & $62.10_{\pm0.99}$ & $62.43_{\pm1.52}$ & $62.63_{\pm1.26}$ & 62.39 & $57.40_{\pm0.31}$ & $57.95_{\pm0.58}$ & $57.22_{\pm1.60}$ & 57.52 \\
 &  & CDANN~\cite{Li_2018_ECCV} & $61.86_{\pm0.92}$ & $61.56_{\pm1.43}$ & $60.73_{\pm1.14}$ & 61.38 & $62.59_{\pm1.63}$ & $61.80_{\pm0.41}$ & $63.67_{\pm1.59}$ & 62.69 & $58.77_{\pm0.91}$ & $59.70_{\pm0.42}$ & $61.15_{\pm0.16}$ & 59.87 \\
 &  & SagNet~\cite{nam2021reducingdomaingapreducing} & $60.75_{\pm1.85}$ & $60.18_{\pm1.90}$ & $59.36_{\pm0.36}$ & 60.10 & $61.43_{\pm1.57}$ & $62.62_{\pm0.96}$ & $63.14_{\pm0.34}$ & 62.40 & $57.37_{\pm1.11}$ & $57.91_{\pm1.26}$ & $59.21_{\pm1.68}$ & 58.16 \\
 &  & IB\_ERM~\cite{ahuja2022invarianceprinciplemeetsinformation} & $59.40_{\pm1.98}$ & $60.39_{\pm0.95}$ & $61.02_{\pm0.92}$ & 60.27 & $63.99_{\pm1.54}$ & $63.23_{\pm1.00}$ & $62.70_{\pm0.97}$ & 63.31 & $59.22_{\pm1.46}$ & $60.18_{\pm0.86}$ & $59.45_{\pm0.28}$ & 59.62 \\
 &  & CondCAD~\cite{ruan2022optimalrepresentationscovariateshift} & $61.00_{\pm1.71}$ & $60.61_{\pm1.24}$ & $60.54_{\pm0.23}$ & 60.72 & $62.66_{\pm1.44}$ & $60.68_{\pm1.89}$ & $63.34_{\pm1.88}$ & 62.23 & $59.66_{\pm1.92}$ & $58.99_{\pm0.10}$ & $59.10_{\pm1.31}$ & 59.25 \\
 &  & EQRM~\cite{eastwood2023probabledomaingeneralizationquantile} & $59.43_{\pm0.67}$ & $58.99_{\pm0.15}$ & $61.67_{\pm0.46}$ & 60.03 & $60.99_{\pm1.54}$ & $60.18_{\pm1.80}$ & $62.47_{\pm0.73}$ & 61.21 & $56.82_{\pm1.91}$ & $59.65_{\pm0.57}$ & $57.02_{\pm0.93}$ & 57.83 \\
 &  & ERM++~\cite{teterwak2024ermimprovedbaselinedomain} & $60.73_{\pm1.63}$ & $59.27_{\pm1.90}$ & $59.43_{\pm0.68}$ & 59.81 & $62.69_{\pm1.25}$ & $61.50_{\pm0.19}$ & $62.98_{\pm0.70}$ & 62.39 & $59.35_{\pm1.83}$ & $58.16_{\pm0.14}$ & $58.79_{\pm0.35}$ & 58.77 \\
 &  & URM~\cite{krishnamachari2024uniformly} & $62.05_{\pm1.73}$ & $60.43_{\pm0.71}$ & $60.15_{\pm0.41}$ & 60.88 & $61.76_{\pm0.39}$ & $63.79_{\pm0.33}$ & $62.04_{\pm0.74}$ & 62.53 & $56.62_{\pm0.33}$ & $57.96_{\pm0.74}$ & $57.27_{\pm1.52}$ & 57.28 \\

\bottomrule
\end{tabular}
}
\end{table*}

\section{Theoretical Analysis}
\label{app:theory}
\subsection{Disentanglement Guarantees}
This section proves from an information-theoretic perspective how the MAF framework effectively strips away the modality-specific style $S$ and isolates the invariant cross-modal forgery essence $\varepsilon$.

\noindent\textbf{Definition (Latent Variable Hypothesis).} 
Assume any input signal $X$ is generated by two latent and mutually independent random variables: the forgery essence $\varepsilon \sim P(\varepsilon)$ and the modality style $S \sim P(S)$. The goal of the universal detector $\mathcal{F}$ is to predict the binary label $Y \in \{0,1\}$ based on the extracted feature representation $Z = \Phi(X)$.
According to the Information Bottleneck (IB)~\cite{ahuja2021invariance} principle, an ideal feature representation $Z$ should maximize its predictive power for $Y$ while minimizing the retention of redundant information (i.e., style $S$) from the source input $X$. We formulate the ideal information-theoretic objective of MAF as:
\begin{equation}
    \min_{Z} \mathcal{I}(Z; S) - \beta \mathcal{I}(Z; Y),
\end{equation}
where $\mathcal{I}(\cdot;\cdot)$ denotes mutual information, and $\beta$ is the Lagrange multiplier.

\noindent\textbf{Theorem (Disentanglement Upper Bound).} 
In the MAF framework, minimizing the domain generalization loss $\mathcal{L}_{DG}$ is equivalent to minimizing a variational upper bound on the mutual information $\mathcal{I}(Z; S)$.

\noindent\textbf{Proof.} 
By the definition of mutual information, we have:
\begin{equation}
    \mathcal{I}(Z; S) = \mathbb{E}_{P(Z,S)} \left[ \log \frac{P(Z|S)}{P(Z)} \right].
\end{equation}
Since the true marginal distribution $P(Z)$ is intractable, we introduce a variational approximation distribution $Q(Z)$. Given the non-negativity of the Kullback-Leibler (KL) divergence ($D_{KL}(P(Z) || Q(Z)) \geq 0$), we can derive the variational upper bound:
\begin{equation}
    \mathcal{I}(Z; S) \leq \mathbb{E}_{P(Z,S)} \left[ \log \frac{P(Z|S)}{Q(Z)} \right].
\end{equation}
In the MAF algorithm, we treat different physical modalities as distinct "domains". When we apply constraints like IRM~\cite{arjovsky2019invariant} or CDANN~\cite{li2018deep} (denoted as $\mathcal{L}_{DG}$), we are essentially forcing the conditional distribution $P(Z|S=k)$ for a given modality $S=k$ to approach a marginal prior $Q(Z)$ that is independent of $S$ (i.e., the "forensic space"). Therefore, by optimizing $\mathcal{L}_{DG}$, we strictly suppress the upper bound of $\mathcal{I}(Z; S)$, mathematically guaranteeing the asymptotic independence of the feature $Z$ from the physical representation $S$. Simultaneously, the cross-entropy loss $\mathcal{L}_{cls}$ ensures the maximization of $\mathcal{I}(Z; Y)$.


\subsection{Dark Modality Bounds}
To formalize the framework's capacity for zero-shot generalization to an isolated perceptual target (Strong MAF), we establish an error bound based on domain adaptation theory~\cite{ben2010theory}. Let $\mathcal{D}_{seen}$ denote the joint source domain distribution of the training modalities, and $\mathcal{D}_{unseen}$ represent the target domain distribution of the unseen "dark modality".

\noindent\textbf{Theorem ($\mathcal{H}\Delta\mathcal{H}$-Divergence Generalization Bound).}
For a universal hypothesis $h \in \mathcal{H}$, the expected target risk $\epsilon_{unseen}(h)$ is bounded by:
\begin{equation}
    \epsilon_{unseen}(h) \leq \epsilon_{seen}(h) + \frac{1}{2} d_{\mathcal{H}\Delta\mathcal{H}}(\mathcal{D}_{seen}, \mathcal{D}_{unseen}) + \lambda^*
\end{equation}

\noindent\textbf{Proof and Analytical Leap.}
The source risk $\epsilon_{seen}(h)$ is empirically minimized during training via $\mathcal{L}_{cls}$, and the ideal joint hypothesis risk $\lambda^*$ is assumed to be a negligibly small constant under our core theoretical premise that a universally shared generative logic $\varepsilon$ exists across all modalities.
The viability of the bound depends entirely on the distribution divergence term $d_{\mathcal{H}\Delta\mathcal{H}}$. In a standard semantic feature space, the macroscopic physical heterogeneity of different sensors (e.g., audio vs. visual) renders $d_{\mathcal{H}\Delta\mathcal{H}}$ intractably large, leading to baseline model collapse.
The MAF framework resolves this by mapping inputs into a decoupled forensic space. By systematically stripping away the modality-specific style component $S$, the framework isolates the latent generative trace. Because the "dark modality" is fundamentally generated by the same underlying algorithmic biases, its extracted feature residuals are mathematically forced into the joint support set of the source distribution $\mathcal{D}_{seen}$. This mechanism inherently minimizes $d_{\mathcal{H}\Delta\mathcal{H}}$, justifying the empirical collapse of the cross-modal KL divergence and proving that the expected error on entirely isolated modalities is strictly bounded.


\subsection{Dimensionality Reduction Theory}
As empirically observed in Section 6.4 (Table 3) of the main manuscript, the effective intrinsic dimensionality ($k_{95}$) experiences a dramatic collapse when features transition from the semantic space to the proposed forensic space. This section provides a rigorous mathematical explanation for this phenomenon using covariance matrix spectral theory.

Formally, let the feature matrix of a given space be $\mathbf{Z} \in \mathbb{R}^{N \times d}$, where $N$ is the number of samples and $d$ is the feature dimension. Its covariance matrix is defined as:
\begin{equation}
    \mathbf{\Sigma}_{Z} = \frac{1}{N} \sum_{i=1}^N (z_i - \bar{z})(z_i - \bar{z})^T.
\end{equation}
In the original semantic space, the representation $Z_{sem}$ is highly entangled, simultaneously capturing the modality-specific style $S$ and the latent forgery essence $\varepsilon$. Assuming independence between the generative trace and the physical medium, its covariance matrix can be approximately decomposed into two orthogonal subspaces:
\begin{equation}
    \mathbf{\Sigma}_{sem} \approx \mathbf{\Sigma}_{\varepsilon} + \mathbf{\Sigma}_{S}.
\end{equation}
Because the physical modality $S$ encapsulates highly heterogeneous macroscopic semantic variables (e.g., audio frequency spectra, spatial visual textures, and temporal video dynamics), the number of non-zero eigenvalues in $\mathbf{\Sigma}_{S}$ is massive. This inherently high-rank nature of $\mathbf{\Sigma}_{S}$ dictates that a large number of principal components are required to explain the variance, leading to an effective dimensionality of $k_{95} > 100$ in the semantic space.

During the MAF optimization process, the framework essentially learns an implicit projection matrix $\mathbf{P} \in \mathbb{R}^{d \times r}$ (where $r \ll d$) such that the mapped forensic feature is $Z_{for} = Z_{sem}\mathbf{P}$. The cross-modal invariance constraint imposed by the domain generalization loss ($\mathcal{L}_{DG}$) is mathematically equivalent to solving the following trace minimization problem:
\begin{equation}
    \min_{\mathbf{P}} \operatorname{Tr}\left(\mathbf{P}^T \mathbf{\Sigma}_{S} \mathbf{P}\right).
\end{equation}
This minimization occurs subject to the condition that $\operatorname{Tr}\left(\mathbf{P}^T \mathbf{\Sigma}_{\varepsilon} \mathbf{P}\right)$ is sufficiently preserved to maintain accurate classification performance (enforced by $\mathcal{L}_{cls}$).
To satisfy this objective, the optimal projection matrix $\mathbf{P}$ is forced to align its column vectors with the null space (or the eigenvectors corresponding to the infinitesimally small eigenvalues) of $\mathbf{\Sigma}_{S}$. Consequently, in the mapped forensic space, the variance contributed by the physical style is eradicated:
\begin{equation}
    \mathbf{\Sigma}_{for} \approx \mathbf{P}^T \mathbf{\Sigma}_{\varepsilon} \mathbf{P}.
\end{equation}
Crucially, the "shared latent generative biases" (such as localized convolutional artifacts, spectral frequency discrepancies, or specific mathematical flaws in generative algorithms) inherently possess very limited degrees of freedom. Therefore, the covariance matrix of the forgery essence, $\mathbf{\Sigma}_{\varepsilon}$, is fundamentally low-rank.
When $\mathbf{P}$ projects the data onto this low-rank subspace, it mathematically necessitates a "steep decay" in the eigenvalue spectrum. This perfectly elucidates why only 2 to 50 principal components are required to explain 95\% of the variance in the forensic space. This dimensionality collapse does not indicate a loss of discriminative information; rather, it represents the mathematical eradication of redundant style noise, successfully isolating the compact manifold of universal forgery knowledge.

\section{More Discussions}

\noindent$\triangleright$ \textbf{\textit{Q1. What is the main paradigm shift proposed by the MAF framework?}}

The MAF framework pioneers a fundamental paradigm shift from conventional "modality-specific feature fusion" to "modality generalization." Instead of merely aggregating superficial artifacts bound to specific data formats, MAF explicitly disentangles modality-specific styles from the intrinsic data semantics. By doing so, it captures the universal, latent forgery knowledge shared across diverse media, transforming the forensic objective from passively binding observed modalities to proactively generalizing against unseen "dark modalities."



\noindent$\triangleright$ \textbf{\textit{Q2. What is the main difference between semantic alignment and forensic alignment?}}

The fundamental difference lies in their objectives: aligning the content versus aligning the generative trace. Semantic alignment focuses on unifying macro-level concepts (aligning what is depicted), which inadvertently overwhelms and masks microscopic generative biases. In stark contrast, forensic alignment actively strips away these dominant content features and modality-specific physical styles. It focuses on aligning how the media was synthesized, thereby isolating the invariant cross-modal forgery essence.





\noindent$\triangleright$ \textbf{\textit{Q3. Why does the intrinsic dimensionality of features drop so sharply in the forensic space?}}

The sharp dimensionality drop reveals a profound structural truth: the shared latent generative biases possess fundamentally limited degrees of freedom. While diverse physical modalities (e.g., RGB pixels, audio waveforms) artificially inflate the feature space with high-rank stylistic noise, the core generative mechanism remains mathematically low-rank. By actively eradicating this redundant, high-dimensional modality variance, our framework mathematically collapses the feature space to its true, compact essence, the underlying covariance matrix of the forgery itself.



\noindent$\triangleright$ \textbf{\textit{Q4. Why does DeepModal-Bench include both modality-aligned and modality-unaligned dataset groups?}}

This dual-group design serves as a rigorous stress test to eliminate semantic shortcuts. In aligned data, conventional models often "cheat" by exploiting semantic inconsistencies (e.g., audio-visual mismatch) as forensic clues. The unaligned group deliberately removes this crutch. It forces models to prove they can identify universal forgery traces purely through intrinsic generative biases, verifying that the true forensic capability functions strictly independently of content correspondence.

\noindent$\triangleright$ \textbf{\textit{Q5. Why do MML methods sometimes match or surpass DG methods under the Strong MAF setting?}}

This phenomenon exposes a critical limitation of current generalization techniques. Under the Strong MAF setting, strict encoder isolation completely severs the architectural pathways that DG methods rely on to extract modality-invariant features, leading to a catastrophic degradation of their generalization advantage. Stripped of this edge, models are forced to rely on pure representational power. MML methods, having densely absorbed rich, complementary priors from the seen modalities during training, use this "brute-force" representational robustness as a safety net. Consequently, this inherent robustness effectively neutralizes the theoretical superiority of DG under extreme isolation.

\section{Limitation and Future Work}
Serving as a brave new idea (BNI), this work takes the crucial first step in shifting the forensic paradigm from modality-binding to modality generalization. Our current evaluation on standard synchronous media is merely a starting point; handling highly irregular or asynchronous heterogeneous data formats (e.g., raw event-camera streams) remains a challenging open problem. To completely close the performance gap when confronting entirely isolated "dark modalities," future work calls for the exploration of advanced self-supervised encoders and meta-learning paradigms to collectively advance this new frontier.

\end{appendices}
\end{document}